\documentclass[onecolumn]{article}         
\usepackage[top=3.2cm, bottom=3.cm, left=2.5cm, right=2.5cm]{geometry}

\renewcommand{\baselinestretch}{1.15}
\usepackage{enumitem}
\usepackage{color}
\usepackage{cite}
\usepackage[cmex10]{amsmath}
\usepackage{makecell}
\usepackage[colorlinks=true,linkcolor=red,citecolor=blue]{hyperref}
\usepackage{epsfig}
\usepackage{graphicx}
\usepackage{amsthm}
\usepackage{amssymb}
\usepackage{booktabs}
\usepackage{mathrsfs}
\usepackage{algorithm}
\usepackage{algorithmic}
\usepackage{bm}
\usepackage{subfigure}
\usepackage{sidecap}
\usepackage{multirow}
\usepackage{url}
\usepackage{diagbox}

\newtheorem{prop}{Proposition}
\theoremstyle{definition}
\newtheorem{defn}{Definition}[section]

\begin{document}
%
\title{A Functional Representation for Graph Matching\thanks{This research is funded by NSFC-projects under the contracts No.61771350 and No.41820104006.}}

\author{Fu-Dong~Wang$^1$, Gui-Song~Xia$^1$, Nan~Xue$^1$, \\ Yipeng~Zhang$^2$, Marcello Pelillo$^3$,
\vspace{3mm}\\
{\em $^1$LIESMARS-CAPTAIN, Wuhan University, Wuhan, China}\\
{\em $^2$School of Computer Science, Wuhan University, China} \\
{\em $^3$Computer Vision Lab., University of Venice, Italy} 
\vspace{2mm}\\
\{{\em fu-dong.wang, guisong.xia, xuenan, zhangyp}\}@whu.edu.cn, \\ pelillo@unive.it.
}


\maketitle

\begin{abstract}
Graph matching is an important and persistent problem in computer vision and pattern recognition for finding node-to-node correspondence between graph-structured data. However, as widely used, graph matching that incorporates pairwise constraints can be formulated as a quadratic assignment problem (QAP), which is NP-complete and results in intrinsic computational difficulties. In this paper, we present a functional representation for graph matching (FRGM) that aims to provide more geometric insights on the problem and reduce the space and time complexities of corresponding algorithms.
To achieve these goals, we represent a graph endowed with edge attributes by a linear function space equipped with a functional such as inner product or metric, that has an explicit geometric meaning.
Consequently, the correspondence between graphs can be represented as a linear representation map of that functional.
Specifically, we reformulate the linear functional representation map as a new parameterization for Euclidean graph matching, which is associative with geometric parameters for graphs under rigid or nonrigid deformations. This allows us to estimate the correspondence and geometric deformations simultaneously. The use of the representation of edge attributes rather than the affinity matrix enables us to reduce the space complexity by two orders of magnitudes. Furthermore, we propose an efficient optimization strategy with low time complexity to optimize the objective function. The experimental results on both synthetic and real-world datasets demonstrate that the proposed FRGM can achieve state-of-the-art performance.
\end{abstract}

\maketitle

\section{Introduction}\label{sec:introduction}

Graph matching (GM) is widely used to find node-to-node correspondence~\cite{[2016-Yan-ICMR],[2004-Conte-IJPRAI]} between graph-structured data in many computer vision and pattern recognition tasks, such as shape matching and retrieval~\cite{[2002-Belongie-pami],[2016-Garro-pami]}, object categorization~\cite{[2011-Duchenne]}, action recognition~\cite{[2012-Yao-eccv]}, and structure from motion~\cite{[2016-Shen-eccv]}, to name a few. In these applications, real-world data are generally represented as abstract graphs equipped with node attributes ({\em e.g., SIFT descriptor, shape context}) and edge attributes ({\em e.g., relationships between nodes}). In this way, many GM methods have been proposed based on the assumption that nodes or edges with more similar attributes are more likely to be matched. Generally, GM methods construct objective functions w.r.t. the varying correspondence to measure similarities (or dissimilarities) between nodes and edges. Then, they maximize (or minimize) the objective functions to pursue an optimal correspondence that achieves maximal (or minimal) total similarities (or dissimilarities) between two graphs. In the literature, an objective function is generally composed of unary~\cite{[2002-Belongie-pami]}, pairwise~\cite{[2005-Leordeanu],[2010-Cho-eccv]} or higher-order~\cite{[2011-Lee-cvpr],[2015-Yan-cvpr]} potentials. In practice, matching graphs using only unary potential (node attributes) might lead to undesirable results due to the insufficient discriminability of node attributes. Therefore, pairwise or higher-order potentials are often integrated to better preserve the structural alignments between graphs.

Although the past decades have witnessed remarkable progresses in GM~\cite{[2016-Yan-ICMR]}, there are still many challenges with respect to both computational difficulty and formulation expression. Specifically, as widely used, GM that incorporates pairwise constraints can be formulated as a quadratic assignment problem (QAP)~\cite{[2007-Loiola-ejor]}, among which Lawler's QAP~\cite{[1963-Lawler]} and Koopmans-Beckmann's QAP~\cite{[1957-Koopmans]} are two common formulations. However, due to the NP-complete~\cite{[1979-Garey]} nature of QAP, only approximate solutions are available in polynomial time.
In practice, solving GM problems with pairwise constraints often encounters intrinsic difficulties due to the high computational complexity in space or time. For GM methods that apply Lawler's QAP, the affinity matrix results in high space complexity $\mathbf{O}(m^2n^2)$ w.r.t. the graph sizes $(m,n)$. For GM methods that aim to solve the objective functions with discrete binary solutions through a gradually convex-concave continuous optimization strategy, the verbose iterations result in high time complexity. Restricted by these limitations, only graphs with dozens of nodes can be handled by these methods in practice. 

In addition to the computational difficulties, how to formulate the GM model for real applications is also important. Representing real-world data in the conventional graph model can provide some generalities for the general GM methods mentioned above. However, their formulations can neither reflect the geometric nature of real-world data nor handle graphs with geometric deformations (rigid or nonrigid). For example, when the edge attributes of graphs are computed as distances~\cite{[2016-Zhou-pami],[2017-Jiang-cvpr],[2017-Huu-cvpr]} on some explicit or implicit spaces that contain the real-world data, the formulations of the original GM methods that define objective functions in the form of Lawler's or Koompmans-Beckmann's QAP ignore the geometric properties behind these data. They can only achieve generality and ignore the geometric nature of real-world data. For graphs with rigid or nonrigid geometric deformations~\cite{[2016-Zhou-pami]}, the original GM methods cannot compatibly handle the two tasks that estimate both correspondence and deformation parameters because they can hardly provide the correspondence a geometric interpretation that is naturally contained in the deformation parameters.  

\begin{figure}[t!]
	\centering
	\subfigure
	{\includegraphics[width=0.6\linewidth]{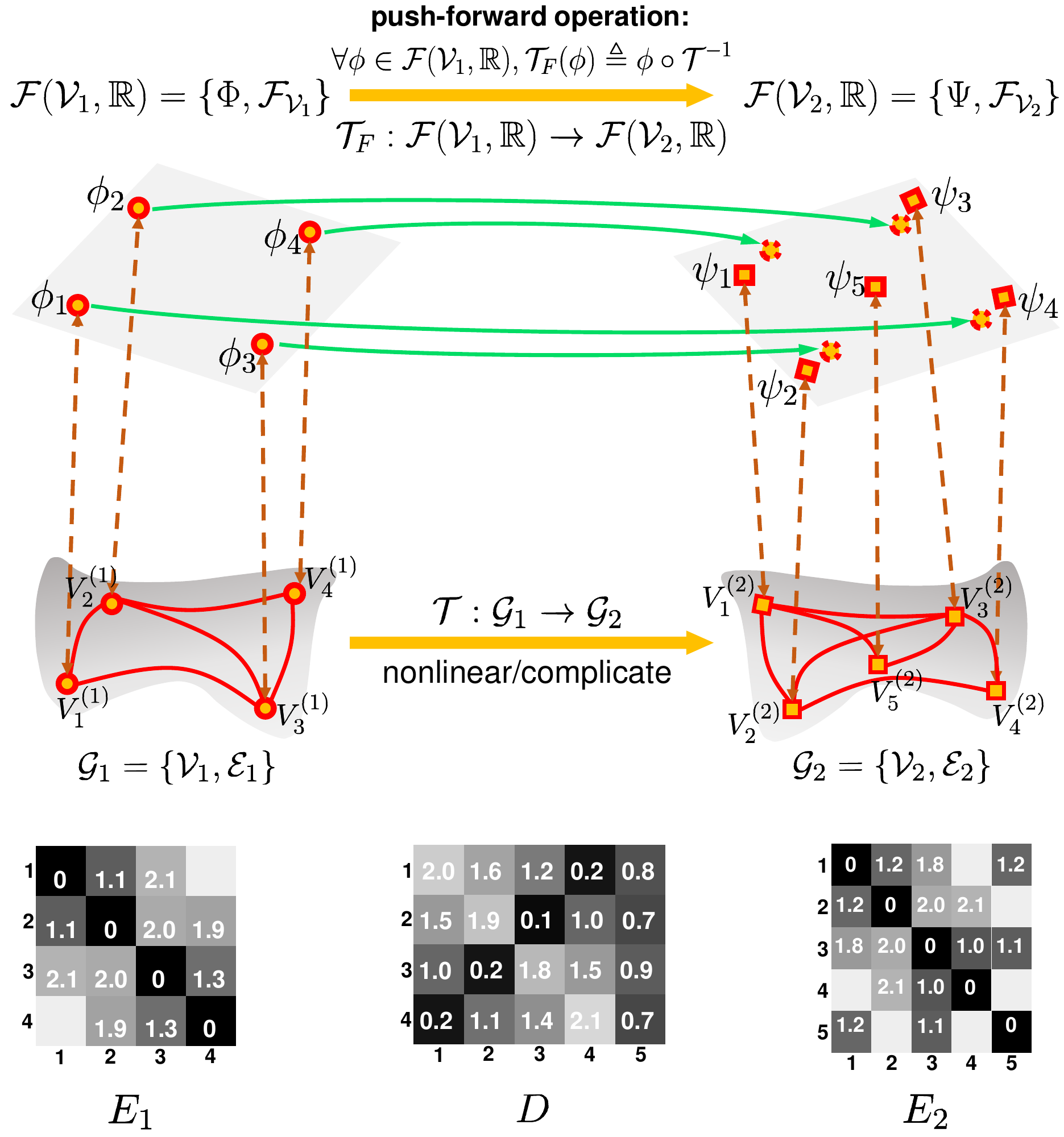}}
	\vspace{-5mm}
	\caption{{\bf FRGM}: given two graphs $\mathcal{G}_1$ and $\mathcal{G}_2$, we construct two function spaces $\mathcal{F}(\mathcal{V}_1,\mathbb{R})$ and $\mathcal{F}(\mathcal{V}_2,\mathbb{R})$ as representations, where $\Phi$ and $\Psi$ are two sets of basis functions that represent the nodes $\mathcal{V}_1$ and $\mathcal{V}_2$, and $\mathcal{F}_{\mathcal{V}_1}$ and $\mathcal{F}_{\mathcal{V}_2}$ are the inner product or metric that represent the edge attributes $\mathbf{E}_1$ and $\mathbf{E}_2$.  The matching between two graphs can be viewed as a transformation $\mathcal{T}:\mathcal{G}_1 \to \mathcal{G}_2$, which may be nonlinear and complicated. Fortunately, $\mathcal{T}$ can be recovered from a linear functional: $\mathcal{T}_F:\mathcal{F}(\mathcal{V}_1,\mathbb{R}) \to \mathcal{F}(\mathcal{V}_2,\mathbb{R})$, which is induced from $\mathcal{T}$ by the push-forward operation and represented by a linear functional representation map $\mathbf{P}\in \mathbb{R}^{m\times n}$. $\mathbf{P}$ is exactly a correspondence between graphs. Based on the inner product or metric defined as $\mathcal{F}_{\mathcal{V}_2}$, each transformed node will lie closer to its correct match, as shown in matrix $\mathbf{D}$. This property is helpful for improving the matching performance.}
	\label{fig:frgm}
\end{figure}

Facing these issues, this paper introduces a new functional representation for graph matching (FRGM). The main idea is to represent the graphs and the node-to-node correspondence in linear functional representations for both general and Euclidean GM models. Specifically, for general GM, as shown in Fig.~\ref{fig:frgm}, given two undirected graphs, we can identically represent the node sets as linear function spaces, on which some specified functionals $\mathcal{F}_{\mathcal{V}}$ ({\em e.g.}, inner product or metric) can be compatibly constructed to represent the edge attributes. Then, between the two function spaces, a functional $\mathcal{T}_F$ induced by the push-forward operation is represented by a linear representation map $\mathbf{P}$, which is exactly the correspondence between graphs. With these concepts, our general GM algorithm is proposed by minimizing the objective function w.r.t. $\mathbf{P}$ that measures the difference of graph attributes between graph $\mathcal{G}_1$ and its transformed graph $\mathcal{T}(\mathcal{G}_1)$. Namely, we want an optimal functional $\mathcal{T}_F$ in the sense of preserving the inner product or metric. For the Euclidean GM in which the graphs are embedded in Euclidean space $\mathbb{R}^d$, the functional $\mathcal{T}_F$ that plays the role of correspondence between graphs can be directly deduced on the background space $\mathbb{R}^d$. Due to the natural linearity of $\mathbb{R}^d$, $\mathcal{T}_F$ can also be represented by a linear representation map $\mathbf{P}$, which is not only a parameterization for GM but also associative with geometric parameters for graphs under geometric deformations. A preliminary version of this work was presented in~\cite{[2018-Wang]}.

FRGM only needs to compute and store the edge attributes of graphs; thus its space complexity is $\mathbf{O}(n^2)$ (with $m\leq n$). To reduce the time complexity, we first propose an optimization algorithm with time complexity $\mathbf{O}(n^3)$ based on the Frank-Wolfe method. Then, by taking advantage of the specified property of the relaxed feasible field, we improve the Frank-Wolfe method by an approximation that has a lower time complexity of $\mathbf{O}(mn)$.

The contributions of this paper can be distinguished in the following aspects:
 
\begin{itemize}	
	\item[-] We introduce a new functional representation perspective that can bridge the gap between the formulation of general GM and the geometric nature behind the real-world data. This guides us in constructing more efficient objective functions and algorithms for general GM problem.
	\item[-] For graphs embedded in Euclidean space, we extend the linear functional representation map as a new geometric parameterization that achieves compatibility with the geometric parameters of graphs. This helps to globally handle graphs with or without geometric deformations.
	\item[-] We propose GM algorithms with low space complexity and time complexity by avoiding the use of an affinity matrix and by improving the optimization strategy. The proposed algorithms outperform the state-of-the-art methods in terms of both efficiency and accuracy.
\end{itemize}  

The remainder of this paper is organized as follows. Sec.~\ref{sec:bg} presents the mathematical formulation and related work of GM. In Sec.~\ref{sec:FRGM-general} we demonstrate the functional representation for GM in general settings and the resulting algorithm. In Sec.~\ref{section:FRGM-Euclidean} and Sec.~\ref{section:deformal GM}, we discuss FRGM for matching graphs in Euclidean space with and without geometric deformations, respectively. In Sec.~\ref{section:numerical}, we present a numerical analysis of our optimization strategy. Finally, we report the experimental results and analysis in Sec.~\ref{sec:experiment} and conclude this paper in Sec.~\ref{conclusion}. 

\section{Background and Related Work}\label{sec:bg}
This section  first introduces the preliminaries and basic notations of GM and then it discusses some related works on GM.

\subsection{Definition of GM Problem}
An undirected graph $\mathcal{G}=\left\{\mathcal{V},\mathcal{E}\right\}$
of size $m$ is defined by a discrete set of nodes $\mathcal{V}=\{V_i\}_{i=1}^{m}$ and a set of undirected edges $\mathcal{E}\subseteq \mathcal{V}\times \mathcal{V}$ such that $(V_{i_1},V_{i_2})=(V_{i_2},V_{i_1})$. Generally, the edge of graph $\mathcal{G}$ is written as a symmetric edge indicator matrix (also denoted as) $\mathcal{E}\in \mathbb{R}^{m\times m}$, where $\mathcal{E}_{i_1i_2}=1$ if there is an edge between $V_{i_1}$ and $V_{i_2}$, and $\mathcal{E}_{i_1i_2}=0$ otherwise. An important generalization is the weighted graph defined by the association of
non-negative real values $\mathcal{E}_{i_1i_2}$ to graph edges, and $\mathcal{E}$ is called adjacency weight matrix. We assume graphs with no self-loop in this paper, {\em i.e.}, $\mathcal{E}_{ii}=0$. 
 
In many real applications, graph $\mathcal{G}$ is associated with node and edge attributes expressed as scalars or vectors. For an attribute graph $\mathcal{G}$, we denote  $\mathbf{v}_{i}\in \mathbb{R}^{d_v}$ as the node attribute of $V_i$ and $\mathbf{e}_{i_1i_2}\in \mathbb{R}^{d_e}$ as the edge attribute of $\mathcal{E}_{i_1i_2}$. Typically, an edge attribute matrix $\mathbf{E}\in \mathbb{R}^{m\times m}$ will be calculated by some user-specified functions such as $\mathbf{E}_{i_1i_2} = \phi(\mathbf{v}_{i_1},\mathbf{v}_{i_2})$ or $\mathbf{E}_{i_1i_2}=\phi(\mathbf{e}_{i_1i_2})$.

Given two graphs $\mathcal{G}_1=\{\mathcal{V}_1,\mathcal{E}_1\},\mathcal{G}_2=\{\mathcal{V}_2,\mathcal{E}_2\}$ of size $m$ and $n$ ($m\leq n$) respectively, the GM problem is to find an optimal node-to-node
correspondence $\mathbf{P}\in \left\{0,1\right\}^{m\times n}$ ,
where $\mathbf{P}_{ij}=1$ when the nodes $V^{(1)}_i \in \mathcal{V}_1$ and $V^{(2)}_j  \in \mathcal{V}_2$ are matched and $\mathbf{P}_{ij}=0$ otherwise. It is clear that any possible correspondence $\mathbf{P}$ equals a (partial) permutation matrix when GM imposes the one-to-(at most)-one constraints. Therefore, the feasible field of $\mathbf{P}$ can be defined as:
\begin{equation}
\mathcal{P}\triangleq \left\{\mathbf{P}\in \left\{0,1\right\}^{m\times n}; \mathbf{P1}_n=\mathbf{1}_m,\mathbf{P}^T\mathbf{1}_m\le \mathbf{1}_n\right\},
\end{equation}
where $\mathbf{1}_m$ is a unit vector. When $m=n$, $\mathbf{P}\in \mathcal{P}$ is orthogonal: $\mathbf{P}\mathbf{P}^T=\mathbf{I}_{m}$, where $\mathbf{I}_{m}$ is a unit matrix.

 To find the optimal correspondence, GM methods that incorporate pairwise constraints generally minimize or maximize their objective functions w.r.t. $\mathbf{P}$ upon the feasible field $\mathcal{P}$. There are two main typical objective functions: Lawler's QAP~\cite{[1963-Lawler]} and Koopmans-Beckmann's QAP~\cite{[1957-Koopmans]}. 

The main idea behind Lawler's QAP~\cite{[1963-Lawler],[2005-Leordeanu],[2009-Leordeanu-nips],[2010-Cho-eccv],[2016-Zhou-pami]} is to maximize the sum of
the node and edge similarities: 
{\small
\begin{align}
	\max_{\mathbf{P}\in\mathcal{P}}\mathbf{P}_v^T\mathbf{K}\mathbf{P}_v =
	\sum_{ij}\mathbf{P}_{ij}\mathbf{K}_{ij;ij} + \sum_{{\tiny{\substack{(i_1,i_2)\\(j_1,j_2)}}}}\mathbf{P}_{i_ij_1}\mathbf{K}_{i_1j_1;i_2j_2}\mathbf{P}_{i_2j_2},
		\label{eq:gm=lawler}
\end{align}}where $\mathbf{P}_v$ is the columnwise vectorized replica of $\mathbf{P}$. The diagonal element $\mathbf{K}_{ij;ij}$ measures the node affinity calculated with node attributes as $\Phi_v(\mathbf{v}^{(1)}_i,\mathbf{v}^{(2)}_j)$, and $\mathbf{K}_{i_1j_1;i_2j_2}$ measures the edge affinity calculated with edge attributes as $\Phi_e(\mathbf{e}^{(1)}_{i_1i_2},\mathbf{e}^{(2)}_{j_1,j_2})$. $\mathbf{K}\in \mathbb{R}^{mn\times mn}$ is called the affinity matrix of $\mathcal{G}_1$ and $\mathcal{G}_2$.

Koopmans-Beckmann's QAP~\cite{[1957-Koopmans],[2009-Zaslavskiy-pami]} formulates GM as
\begin{equation}
	\max_{\mathbf{P}\in\mathcal{P}} -\text{tr}(\mathbf{U}^T\mathbf{P})+\lambda\text{tr}(\mathcal{E}_1\mathbf{P}\mathcal{E}_2\mathbf{P}^T),
	\label{eq:gmKoom22}
\end{equation}
where $\{\mathbf{U}_{ij}\}\in \mathbb{R}^{m\times n}$ measures the dissimilarity between node $V^{(1)}_i$ and $V^{(2)}_j$, and $\mathcal{E}_1, \mathcal{E}_2$ are the adjacency weight matrices of $\mathcal{G}_1$ and $\mathcal{G}_2$. $\lambda\ge 0$ is a weight between the unary and pairwise terms.
This formulation differs from Eq.~\eqref{eq:gm=lawler} mainly in the pairwise term， which measures the edge compatibility as the linear similarity of adjacency matrices $\mathcal{E}_1$ and $\mathcal{E}_2$. In fact, Eq.~\eqref{eq:gmKoom22} can be regarded as a special case of Lawler's QAP (Eq.~\eqref{eq:gm=lawler}) if $\mathbf{K}=\mathcal{E}_1\otimes \mathcal{E}_2$, where $\otimes$ denotes the Kronecker product. With this formulation, the space complexity of GM is $O(n^2)$ and much lower than that $O(m^2n^2)$ of Eq.~\eqref{eq:gm=lawler}.

The Eq.~\eqref{eq:gmKoom22} has another approximation, which aims to minimize the node and edge dissimilarity between two graphs:
\begin{equation}
	\min_{\mathbf{P}\in \mathcal{P}} \langle \mathbf{P},\mathbf{U}\rangle_F +\frac{\lambda}{2} ||\mathcal{E}_1-\mathbf{P}\mathcal{E}_2\mathbf{P}^T||^2_F,
	\label{eq:gmKoom1}
\end{equation}
where $\langle \cdot,\cdot\rangle_F$ is the Frobenius dot-product defined as $\langle A,B\rangle_F =\sum_{ij}A_{ij}B_{ij}$ and $||\cdot||_F^2$ is the Frobenius matrix norm defined as $||A||_F^2 = \langle A,A\rangle_F$. 
The conversion from Eq.~\eqref{eq:gmKoom1} to Eq.~\eqref{eq:gmKoom22} holds equally under the fact that any $\mathbf{P}\in \mathcal{P}$ is an orthogonal matrix. 

Due to the NP-complete nature of the above formulations, GM methods generally approximate the discrete feasible field $\mathcal{P}$ by a continuous relaxation $\hat{\mathcal{P}}: \mathbf{P}_{ij}\in[0,1]$, which is known as the {\bf doubly stochastic relaxation}. Then the objective functions can be approximately solved by applying constrained optimization methods and employing a post-discretization step such the Hungarian algorithm~\cite{[2010-Kuhn]} to obtain a discrete binary solution. 

\subsection{Related Work}
Over the past decades, the GM problem of finding node-to-node correspondence between graphs has been extensively studied~\cite{[2004-Conte-IJPRAI],[2016-Yan-ICMR]}. Earlier works (exact GM)~\cite{[1976-Ullmann],[2004-Cordella]} tended to regard GM as (sub)graph isomorphism. However, this assumption is too strict and leads to less flexibility for real applications. Therefore, later works on GM (inexact/error-tolerant GM)~\cite{[2017-Huu-cvpr],[2016-Zhou-pami],[2009-Leordeanu-nips],[2010-Cho-eccv]} focused more on finding inexact matching between weighted graphs via optimizing more flexible objective functions.

Among the inexact GM methods, some of them aim to reduce the considerable space complexity caused by the affinity matrix $\mathbf{K}$ in Eq.~\eqref{eq:gm=lawler}. A typical work is the factorized graph matching (FGM)~\cite{[2016-Zhou-pami]}, which factorized $\mathbf{K}$ as a Kronecker product of several smaller matrices. An efficient sampling heuristic was proposed in~\cite{[2008-Zass-cvpr]} to avoid storing the whole $\mathbf{K}$ at once. Some works~\cite{[2009-Zaslavskiy-pami],[2016-Lu]} constructed objective functions similar to Eq.~\eqref{eq:gmKoom1} or Eq.~\eqref{eq:gmKoom22} to avoid using matrix $\mathbf{K}$. Our work use the representation of edge attributes rather than $\mathbf{K}$. 

Since exactly solving the objective functions upon discrete feasible field $\mathcal{P}$ is NP-complete, most GM methods relax $\mathcal{P}$ for approximation purpose in several ways. The first typical relaxation is spectral relaxation, as proposed in~\cite{[2005-Leordeanu],[2006-Cour-nips]}, by forcing $||\mathbf{P}||_2=1$; then, the solution is computed as the leading eigenvector of $\mathbf{K}$. The second relaxation~\cite{[2009-Zaslavskiy-pami]} is to consider $\mathcal{P}$ as a subset of an orthogonal matrices set such that $\mathbf{P}\mathbf{P}^T=\mathbf{I}_m$, which is the basis of converting Eq.~\eqref{eq:gmKoom1} into Eq.~\eqref{eq:gmKoom22}. Semidefinite-programming (SDP) was also applied to approximately solve the GM problem in~\cite{[Torr_solvingmarkov],[2005-Schellewald]} by introducing a new variable $\mathbf{X}=\mathbf{P}_v\mathbf{P}_v^T\in \mathbb{R}^{mn\times mn}$ under the convex semidefinite constraint $\mathbf{X}-\mathbf{P}_v\mathbf{P}_v^T\succeq 0$. Then, $\mathbf{P}$ is approximately recovered by $\mathbf{X}$.

The most widely used relaxation approach is the doubly stochastic relaxation $\hat{\mathcal{P}}$, which is the convex hull of $\mathcal{P}$. Since $\hat{\mathcal{P}}$ is a convex set defined in a linear form, it allows the GM objectives functions to be solved by more flexible convex or nonconvex optimization algorithms. To find more global optimal solutions with a binary property, the algorithms proposed in~\cite{[2009-Zaslavskiy-pami],[2016-Zhou-pami],[2014-Liu-pami],[2014-Liu-ijcv]} constructed objective functions in both convex and concave relaxations controlled by a continuation parameter, and then they developed a path-following-based strategy for optimization. These approaches are generally time consuming, particularly for matching graphs with more than dozens of nodes. The graduated assignment method~\cite{[1996-Gold]} iteratively solved a series of first-order approximations of the objective function. Its improvement~\cite{[2012-Tian]} provided more convergence analysis. The decomposition-based work in~\cite{[2013-Torresani-pami]} developed its optimization technique by referring to dual decomposition. Additionally, another method in~\cite{[2017-Huu-cvpr]} decomposed the matching constraints and then used an optimization strategy based on the alternating direction method of multipliers. To ensure binary solutions, some methods such as the integer-projected fixed point algorithm~\cite{[2009-Leordeanu-nips]} and iterative discrete gradient assignment~\cite{[2015-Yan-cvpr]}, have been proposed by searching in the discrete feasible domain. We also adopt the doubly stochastic relaxation, and we construct an objective function that can be solved with a nearly binary solution, which helps to reduce the effect of the post-discretization step.

In addition to approximating the objective functions, some works also intended to provide more interpretations of the GM problem. 
The probability-based works~\cite{[2008-Zass-cvpr],[2013-Egozi]} solved the GM problem from a maximum likelihood estimation perspective. Some learning-based works~\cite{[2009-Caetano-pami],[2012-Leordeanu-ijcv]} went further to explore how to improve the affinity matrix $\mathbf{K}$ by considering rotations and scales of real data. A pioneering work~\cite{[2018-Zanfir]} presented an end-to-end deep learning framework for
GM. A random walk view~\cite{[2010-Cho-eccv]} was introduced by simulating random walks with reweighting jumps. A max-pooling-based strategy was proposed in~\cite{[2014-Cho-cvpr]} to address the presence of outliers. Compared to these these works, the proposed FRGM provides more geometric insights for GM with general settings by using a functional representation to interpret the geometric nature of real-world data, and it then matches graphs embedded in Euclidean space by providing a new parameterization view to handle graphs under geometric deformations.

\section{Functional Representation for GM}\label{sec:FRGM-general}

This section presents the functional representation for general GM that incorporates pairwise constraints. In Sec.~\ref{sec:function-space}, we introduce the function space of a graph, on which functionals can be defined as the inner product or metric to compatibly represent the edge attributes. In Sec.~\ref{sec:FRGM map}, we discuss how to represent the correspondence between graphs as a linear functional representation map between function spaces. Finally, the correspondence is an optimal functional map obtained by the algorithm in Sec.~\ref{sec:general algo}.

\subsection{Function Space on Graph}\label{sec:function-space}
Given an undirected graph $\mathcal{G}=\{\mathcal{V},\mathcal{E}\}$ with edge attribute matrix $\mathbf{E}\in \mathbb{R}^{m\times m}$, we aim to establish function space $\mathcal{F}(\mathcal{V},\mathbb{R})$ of $\mathcal{G}$, on which some geometric structures, such as inner product or metric can be defined. This is especially meaningful when graphs are embedded in explicit or hidden manifolds. 

Let $\mathcal{F}(\mathcal{V},\mathbb{R})$ denote the function space of all real-valued functions on  $\mathcal{V}=\{V_i\}_{i=1}^{m}$. Since $\mathcal{V}$ is finite discrete, we can choose a finite set of basis functions $\Phi=\{\phi_i\}_{i=1}^m$ to explicitly construct $\mathcal{F}(\mathcal{V},\mathbb{R})$. 
\begin{defn} The {\bf\itshape function space} $\mathcal{F}(\mathcal{V},\mathbb{R})$ on graph $\mathcal{G}$ can be defined as:
\begin{equation}
	\mathcal{F}(\mathcal{V},\mathbb{R}) \triangleq \left\{\phi_\mathbf{a}=\sum_ia_i\phi_i,\, \mathbf{a}\triangleq(a_1,...,a_m)^T\in \mathbb{R}^m\right\}.
\end{equation} 
\end{defn}

For example, $\phi_i$ can be chosen as the indicator of $V_i$:
 \begin{equation}\label{eq:indicator}
 \phi_i:\mathcal{V}\to \mathbb{R},\quad \phi_i(V_j)=\left\{  
 \begin{array}{lr}  
 1,j=i. &  \\  
 0,j\neq i.\\    
 \end{array}  
 \right.
 \end{equation}
Considering the fact that the correspondence matrix $\mathbf{P}\in \hat{\mathcal{P}}$ is positive, {\em i.e.}, $\mathbf{P}_{ij}\in [0,1]$, a typical subset of $\mathcal{F}(\mathcal{V},\mathbb{R})$ can be defined as follows, which is the convex hull of $\{\phi_i\}_{i=1}^{m}$:
\begin{equation}
\mathcal{C}(\mathcal{V},\mathbb{R})\triangleq \left\{\phi_\mathbf{a}=\sum_ia_i\phi_i;\sum_ia_i=1,\mathbf{a}\in \mathbb{R}_+^m \right\}.
\end{equation} 

Once the function space $\mathcal{F}(\mathcal{V},\mathbb{R})$ is built, some trivial operations can be defined, {\em e.g.}, inner product $\langle \phi_\mathbf{a},\phi_\mathbf{b}\rangle=\sum_ia_ib_i$ and metric $d(\phi_\mathbf{a},\phi_\mathbf{b})=(\sum_i(a_i-b_i)^2)^{1/2}$. However, these definitions cannot express the edge attribute $\mathbf{E}_{i_1i_2}$. Therefore, we aim to define some other operations to represent $\mathbf{E}\in \mathbb{R}^{m\times m}$ based on $\mathcal{F}(\mathcal{V},\mathbb{R})$. An available approach is to define functionals on the product space $\mathcal{F}(\mathcal{V},\mathbb{R})\times \mathcal{F}(\mathcal{V},\mathbb{R})$. Moreover, the functionals should (1) be compatible with $\mathbf{E}$ and (2) have geometric structures such as inner product or metric, as demonstrated in the following:

\begin{defn}
	A functional $\mathbb{F}_{\mathcal{V}}:\mathcal{F}(\mathcal{V},\mathbb{R})\times\mathcal{F}(\mathcal{V},\mathbb{R})\to \mathbb{R}$ is {\bf\itshape{{compatible}}} with $\mathbf{E}$ if it satisfies $\mathbb{F}_{\mathcal{V}}(\phi_{i_1},\phi_{i_2})=\mathbf{E}_{i_1i_2}$. 
\end{defn}

Among all the compatible functionals, there are some specified ones that can be defined as the inner product or metric on the function space $\mathcal{F}(\mathcal{V},\mathbb{R})$ or its subset $\mathcal{C}(\mathcal{V},\mathbb{R})$, as follows.

\begin{figure}[ht!]
 	\centering
	{\includegraphics[width=0.77\linewidth]{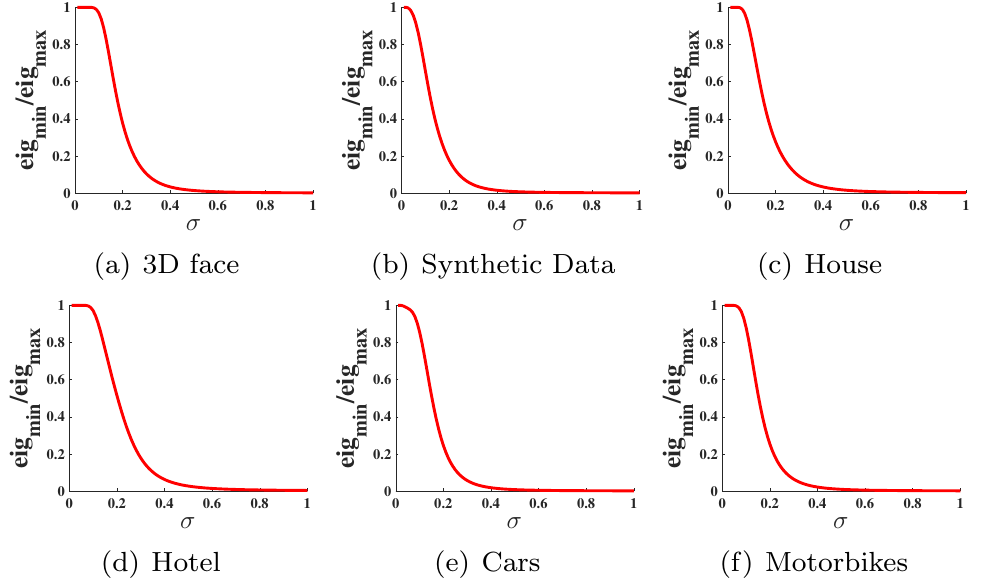}}
 	\vspace{-2mm}
 	\caption{Empirical statistics of $\text{exp}(-{{\mathbf{E}^2}/{\sigma^2}})$ extracted from the realistic and synthetic datasets used in the experimental section. For thousands of graphs in all six datasets, as $\sigma$ varies from 0 to 1, the ratio between the minimum and maximum eigenvalues of $\text{exp}(-{{\mathbf{E}^2}/{\sigma^2}})$ changes with a similar tendency.}
 	\label{fig:eig_ave}
 	\vspace{-2mm}
 \end{figure}

\begin{defn} The {\bf\itshape inner product} on the function space $\mathcal{F}(\mathcal{V},\mathbb{R})$ can be defined in an explicit form: $\forall \phi_\mathbf{a},\phi_\mathbf{b}\in \mathcal{F}(\mathcal{V},\mathbb{R})$,
	\label{def:inner}	
\begin{align}
		\mathbb{F}_{\mathcal{V}}(\phi_\mathbf{a},\phi_\mathbf{b})&\triangleq\sum_{i_1,i_2}a_{i_1}b_{i_2}\mathbb{F}_{\mathcal{V}}(\phi_{i_1},\phi_{i_2})=\sum_{i_1,i_2}a_{i_1}b_{i_2}\mathbf{E}_{i_1i_2}.
		\label{eq:inner}
\end{align}
\end{defn}
For the given edge attribute matrix $\mathbf{E}$ that is symmetric, $\mathbb{F}_{\mathcal{V}}(\cdot,\cdot)$ satisfies the first two inner product axioms: symmetry and linearity. To satisfy the third axiom, positive-definiteness, we need more knowledge about $\mathbf{E}$, {\em e.g.}, $\mathbf{E}$ is positive-definite. However, if the positive-definiteness is too strong, we can relax it to a weaker condition.
\begin{prop}\label{prop:inner}
	Assume that $\mathbf{E}$ satisfies $\mathbf{E}_{i_1i_2}= 0$ iff $i_1=i_2$. Then, the functional $\mathbb{F}_{\mathcal{V}}(\cdot,\cdot)$ in Eq.~\eqref{eq:inner} satisfies all three axioms on $\mathcal{F}(\mathcal{V},\mathbb{R})$ by replacing $\mathbf{E}$ with $\text{exp}({-{\mathbf{E}^2}/{\sigma^2}})$ with $\sigma > 0$ small enough. Here, $\mathbf{E}^2$ is a pointwise product.
\end{prop}
This proposition holds because when $\sigma > 0$ is sufficiently small, all the eigenvalues of matrix $\text{exp}({-{\mathbf{E}^2}/{\sigma^2}})$ will be positive. In particular, when $\mathbf{E}$ is computed as a metric (distance) matrix on an explicit or hidden manifold, it satisfies that $\mathbf{E}_{ii}=0$ and  $\text{exp}({-{\mathbf{E}^2}/{\sigma^2}})$ is positive-definite. Moreover, $\sigma$ can be used to adjust the eigenspace of $\text{exp}({-{\mathbf{E}^2}/{\sigma^2}})$. Fig.~\ref{fig:eig_ave} illustrates an empirical study on thousands of $\mathbf{E}$'s extracted from both realistic and synthetic datasets used in Sec.~\ref{sec:experiment}. The edge attribute $\mathbf{E}$ of each graph is computed in a metric form (either Euclidean distance or geodesic distance) and then normalized to $[0,1]$ divided by the maximum element. We can see that 
\vspace{-1mm}
\begin{itemize}
    \item[-] All the eigenvalues of $\text{exp}({-{\mathbf{E}^2}/{\sigma^2}})$ are positive.
    \item[-] The ratio between the minimum and maximum eigenvalues has a similar tendency when $\sigma$ varies from $0$ to $1$.
\end{itemize}
\vspace{-1mm}
It shows that $\text{exp}({-{\mathbf{E}^2}/{\sigma^2}})$ will become indistinguishable if $\sigma$ is too small or unbalanced if $\sigma$ is too large. We can choose a suitable $\sigma$ to adjust the eigenspace of $\text{exp}({-{\mathbf{E}^2}/{\sigma^2}})$ to achieve better matching performance.

 
The inner product $\mathbb{F}_{\mathcal{V}}(\cdot,\cdot)$ can induce a metric by definition $d(\phi_{\mathbf{a}},\phi_{\mathbf{b}})\triangleq \mathbb{F}_{\mathcal{V}}(\phi_{\mathbf{a}}-\phi_{\mathbf{b}},\phi_{\mathbf{a}}-\phi_{\mathbf{b}})^{1/2}$. Moreover, we can also define another metric on the subset $\mathcal{C}(\mathcal{V},\mathbb{R})$ based on $\mathbf{E}$ itself.

\begin{defn} The {\bf\itshape metric} on the convex hull $\mathcal{C}(\mathcal{V},\mathbb{R})$ can be defined in an implicit form: $\forall \phi_\mathbf{a},\phi_\mathbf{b}\in \mathcal{C}(\mathcal{V},\mathbb{R})$,	
	\label{def:metric}
	\begin{align}
	\mathbb{F}_{\mathcal{V}}(\phi_\mathbf{a},\phi_\mathbf{b})&=\mathop{\text{min}}\limits_{\mathbf{\pi}\in \mathcal{P}(\mathbf{a},\mathbf{b})} \pi_{i_1i_2}\mathbb{F}_{\mathcal{V}}(\phi_{i_1},\phi_{i_2})=\mathop{\text{min}}\limits_{\mathbf{\pi}\in \mathcal{P}(\mathbf{a},\mathbf{b})} \pi_{i_1i_2}\mathbf{E}_{i_1i_2},
	\label{eq:metric}
	\end{align}
	where $\mathcal{P}(\mathbf{a},\mathbf{b})=\left\{\pi\in \mathbb{R}_+^{m\times m}; \, \sum_i\pi_{ij}=b_j, \, \sum_j\pi_{ij}=a_i\right\}$. 
\end{defn}
When $\mathbf{E}$ is computed as a metric, $\mathbb{F}_{\mathcal{V}}(\cdot,\cdot)$ satisfies all three distance axioms on $\mathcal{C}(\mathcal{V},\mathbb{R})$, and it is a typical Wasserstein distance (or Sinkhorn distance)~\cite{[2013-Cuturi]}. The definition in Eq.~\eqref{eq:metric} is not differentiable w.r.t. $\mathbf{a},\mathbf{b}$; one can use the entropy-regularized Wasserstein distance~\cite{[2013-Cuturi]} to achieve differentiability.

With the function space equipped with inner product or metric, each graph is assigned with explicit geometric structures that are compatible with the edge attribute. Next, we demonstrate the idea of using the functional map representation to formulate the correspondence between graphs as a functional $\mathcal{T}_{F}$ between two function spaces $\mathcal{T}_{F}:\mathcal{F}(\mathcal{V}_1,\mathbb{R})\to\mathcal{F}(\mathcal{V}_2,\mathbb{R})$.

\subsection{Functional Map Representation for GM}\label{sec:FRGM map}

The matching between two graphs $\mathcal{G}_1=\left\{\mathcal{V}_1,\mathcal{E}_1\right\}$ and $\mathcal{G}_2=\left\{\mathcal{V}_2,\mathcal{E}_2\right\}$ can be viewed as a mapping $\mathcal{T}$ from $\mathcal{V}_1$ to $\mathcal{V}_2$, which may be nonlinear and complicated. Therefore, we use the push-forward operation to induce a functional $\mathcal{T}_F$ rather than $\mathcal{T}$ to equally represent the matching between graphs.

Assume that $\mathcal{T}: \mathcal{V}_1\to\mathcal{V}_2$ is an injective mapping; then, $\mathcal{T}|_{\mathcal{T}(\mathcal{V}_1)}:\mathcal{V}_1\to\mathcal{T}(\mathcal{V}_1)\subseteq\mathcal{V}_2$ is bijective and invertible. Without ambiguity, we can assume that $\mathcal{T}$ is bijective. Each $\mathcal{T}$ induces a natural transformation $\mathcal{T}_F:\mathcal{F}(\mathcal{V}_1,\mathbb{R})\to \mathcal{F}(\mathcal{V}_2,\mathbb{R})$ via the push-forward operation, which is widely used in functional analysis~\cite{[rudin2006]} and real applications~\cite{[2012-Ovsjanikov]}\cite{[2017-Rodola]}:

\begin{defn} The functional $\mathcal{T}_F:\mathcal{F}(\mathcal{V}_1,\mathbb{R})\to \mathcal{F}(\mathcal{V}_2,\mathbb{R})$ induced from $\mathcal{T}$ is defined as: $\forall$ $\phi\in\mathcal{F}(\mathcal{V}_1,\mathbb{R})$, the image of $\phi$ is $\mathcal{T}_F(\phi)\triangleq\phi\circ \mathcal{T}^{-1} \in \mathcal{F}(\mathcal{V}_2,\mathbb{R})$. 
\end{defn}

\begin{prop}
	The original $\mathcal{T}$ can be recovered from $\mathcal{T}_F$.
\end{prop}
For each point $V_i^{(1)}\in \mathcal{V}_1$, it can be associated with an indicator function $\phi_i$ as Eq.~\eqref{eq:indicator}. To recover the image $\mathcal{T}(V_i^{(1)})$ from $\mathcal{T}_F$, we utilize the function $\psi=\mathcal{T}_F(\phi_i):\mathcal{V}_2\to R$, which satisfies $\forall$ $V\in \mathcal{V}_2,$
\begin{equation}
\psi(V)\triangleq\phi\circ \mathcal{T}^{-1}(V)=\left\{  
\begin{array}{lr}  
1, \, \mathcal{T}^{-1}(V)=V_i^{(1)}, &  \\  
0, \, \mathcal{T}^{-1}(V)\neq V_i^{(1)}.\\    
\end{array}  
\right. 
\end{equation}  
Since $\mathcal{T}$ is bijective and invertible, a unique $ V\in \mathcal{V}_2$ exists s.t. $\mathcal{T}^{-1}(V)=V_i^{(1)}$. Then, once we find $\psi(V_j^{(2)})=1$, we have $\mathcal{T}^{-1}(V^{(2)}_j)=V_i^{(1)}$, and $V_j^{(2)}$ must equal the image $\mathcal{T}(V^{(1)}_i)$ of $V^{(1)}_i$: $\mathcal{T}(V^{(1)}_i)=V_j^{(2)}$. Thus, the functional $\mathcal{T}_F$ can be used to equally represent $\mathcal{T}$.

\begin{prop}
 $\mathcal{T}_F$ is a linear mapping from function spaces $\mathcal{F}(\mathcal{V}_1,\mathbb{R})$ to $\mathcal{F}(\mathcal{V}_2,\mathbb{R})$.
\end{prop}
It holds because $\forall f_1, f_2\in \mathcal{F}(\mathcal{V}_1,\mathbb{R}),  \alpha_1,\alpha_2\in \mathbb{R}$,  
\begin{align}
\mathcal{T}_F(\alpha_1f_1 +\alpha_2f_2) &=(\alpha_1f_1 +\alpha_2f_2)\circ\mathcal{T}^{-1}
\nonumber\\
&=\alpha_1f_1\circ\mathcal{T}^{-1}+\alpha_2f_2\circ\mathcal{T}^{-1} \nonumber \\
&=\alpha_1\mathcal{T}_F(f_1)+\alpha_2\mathcal{T}_F(f_2). \nonumber
\end{align}
Although $\mathcal{T}$ may be nonlinear and complicated, $\mathcal{T}_F$ is linear and simple. 

With function spaces $\mathcal{F}(\mathcal{V}_1,\mathbb{R})$ and $\mathcal{F}(\mathcal{V}_2,\mathbb{R})$ defined by basis functions $\Phi=\{\phi_i\}_{i=1}^m$ and $\Psi=\{\psi_j\}_{j=1}^n$ respectively, each basis function $\phi_i$ can be transformed into $\mathcal{F}(\mathcal{V}_2,\mathbb{R})$ and represented in a linear form as $\mathcal{T}_F(\phi_i)=\sum_{j=1}^n\mathbf{P}_{ij}\psi_j$. Whenever $\mathbf{P}$ reaches an extreme point of the feasible field $\hat{\mathcal{P}}$, it is a binary correspondence between graphs, and consequently, $\phi_i$ is transformed into ({\itshape i.e.}, matches) a $\psi_{j'}$, where $\mathbf{P}_{ij'} =1,\mathbf{P}_{i,j\neq j'}=0$.

To find an optimal correspondence between two graphs with edge attributes $\mathbf{E}_1\in \mathbb{R}^{m\times m}$ and $\mathbf{E}_2\in \mathbb{R}^{n\times n}$, we declare that the induced functional $\mathcal{T}_F$ should be able to preserve the geometric structures defined on function spaces. Namely, $\mathcal{T}_F$ should be the inner product or metric preserving. More precisely, for each pair $(\phi_{i_1},\phi_{i_2})$, the functional value $\mathbb{F}_{\mathcal{V}_1}(\phi_{i_1},\phi_{i_2})$ should be similar to the functional value of the transformed pair $(\mathcal{T}_F(\phi_{i_1}),\mathcal{T}_F(\phi_{i_2}))$, which is calculated as
{\small 
\begin{equation}
	\mathbb{F}_{\mathcal{V}_2}(\mathcal{T}_F(\phi_{i_1}),\mathcal{T}_F(\phi_{i_2}))=\mathbb{F}_{\mathcal{V}_2}(\sum_{j=1}^n\mathbf{P}_{i_1j}\psi_j, \, \sum_{j=1}^n\mathbf{P}_{i_2j}\psi_j).
\end{equation}}The functionals defined in { Definition~\ref{def:inner}} or { Definition~\ref{def:metric}} can be used to calculate it. Finally, to incorporate the pairwise constraints, we aim to minimize the total sum as follows:
{\small 
\begin{align}
&\sum_{(i_1,i_2)}\mathcal{E}_{1_{i_ii_2}}\left[\mathbb{F}_{\mathcal{V}_1}(\phi_{i_1}, \, \phi_{i_2})-\mathbb{F}_{\mathcal{V}_2}\big(\mathcal{T}_F(\phi_{i_1}), \, \mathcal{T}_F(\phi_{i_2})\big)\right]^2 \nonumber \\
\triangleq& \sum_{(i_1,i_2)}\mathcal{E}_{1_{i_ii_2}}\left[\mathbf{E}_{1_{i_1i_2}}-\mathbf{F}(\mathbf{P})_{i_1i_2}\right]^2
\triangleq||\mathbf{E}_1-\mathbf{F}(\mathbf{P})||^2_{F,\mathcal{E}_1},
\end{align}}
where $\mathbf{F}(\mathbf{P})_{i_1i_2}\triangleq\mathbb{F}_{\mathcal{V}_2}\left(\sum_{j=1}^n\mathbf{P}_{i_1j}\psi_j, \, \sum_{j=1}^n\mathbf{P}_{i_2j}\psi_j\right)$ is computed based on the edge attributes matrix $\mathbf{E}_2$. Note tha the affinity matrix $\mathbf{K}$ with size $\mathbf{O}(m^2n^2)$ is replaced here by the edge attributes matrix $\mathbf{E}_1$ with size $\mathbf{O}(m^2)$ and $\mathbf{E}_2$ with size $\mathbf{O}(n^2)$.

\subsection{FRGM-G: matching graphs with general settings}\label{sec:general algo}
Here, we propose our FRGM-G algorithm for matching graphs with general settings, {\em i.e.} without knowledge on the geometrical structures of the graphs. To find an optimal correspondence, {\itshape{i.e.}}, functional map $\mathbf{P}$ mentioned above, we first minimize an objective function as 
\begin{equation}
	J_{ori}(\mathbf{P})=(1-\alpha_1)\langle \mathbf{P}, \mathbf{U}\rangle_F + \alpha_1 ||\mathbf{E}_1-\mathbf{F}(\mathbf{P})||^2_{F,\mathcal{E}_1},
\end{equation}
where $\alpha_1\in [0,1]$ balances the weights of the unary term and pairwise term. In general, $J_{ori}(\mathbf{P})$ is nonconvex and minimizing $J_{ori}(\mathbf{P})$ upon the feasible field $\hat{\mathcal{P}}$ results in a local minimum. The minimizer $\mathbf{P}_1^*$ may be not binary, and the post-discretization of $\mathbf{P}_1^*$ may reduce the matching accuracy. Therefore, we next construct another objective function to find a better solution based on the obtained $\mathbf{P}_1^*$.

According to the definition $\mathcal{T}_F(\phi_i)=\sum_{j=1}^n\mathbf{P}_{ij}\psi_j$, each $\mathcal{T}_F(\phi_i)$ lies in the convex set $\mathcal{C}(\mathcal{V}_2,\mathbb{R})$, which is the convex hull of $\{\psi_j\}_{j=1}^n$. Therefore, the transformed functions $\{\mathcal{T}_F(\phi_i)\}_{i=1}^m$ lies in the same function space spanned by $\{\psi_j\}_{j=1}^n$, and the offset between $\{\mathcal{T}_F(\phi_i)\}_{i=1}^m$ and $\{\psi_j\}_{j=1}^n$ can be controlled. Moreover, since $\mathbf{P}_1^*$ indeed preserves the pairwise geometric structure between two graphs, $\mathcal{T}_F(\phi_i)$ will lie closer to the correct matching $\psi_{\delta_i}$. This means that, based on the metric defined on the function spaces, the distance $d(\mathcal{T}_F(\phi_i),\psi_{_{\delta_i}})$ make sense and will be smaller than $d(\mathcal{T}_F(\phi_i),\psi_{_{j\neq\delta_i}})$.
Therefore, we define the second objective function as:
{\small \begin{align}
	J_{int}(\mathbf{P})&=(1-\alpha_2)\langle \mathbf{P},\mathbf{D}\rangle_F
	+\alpha_2||\mathbf{F}(\mathbf{P}^*_1)-\mathbf{F}(\mathbf{P})||^2_{F,\mathcal{E}_1},
\end{align}}where $\mathbf{D}_{ij} = d(\mathcal{T}_F(\phi_i),\psi_{_{j}})$ is the distance between $\mathcal{T}_F(\phi_i)$ and $\psi_{_{j}}$ computed by the metric functional defined on $\mathcal{F}(\mathcal{V}_2,\mathbb{R})$ or $\mathcal{C}(\mathcal{V}_2,\mathbb{R})$. The minimizer $\mathbf{P}^*_2$ can be viewed as a displacement interpolation: to minimize $\langle \mathbf{P},\mathbf{D}\rangle_F$ we obtain a solution $\mathbf{P}^*_0$ that is an extreme point (thus, binary) of the feasible field $\hat{\mathcal{P}}$; to minimize $||\mathbf{F}(\mathbf{P}^*_1)-\mathbf{F}(\mathbf{P})||^2_{F,\mathcal{E}_1}$, we obtain a solution that equals $\mathbf{P}^*_1\in \hat{\mathcal{P}}$. Then, $\mathbf{P}^*_2$ is an interpolation between $\mathbf{P}^*_0$ and $\mathbf{P}^*_2$ controlled by $\alpha_2\in [0,1]$. Finally, we use the Hungarian method to discretize $\mathbf{P}^*_2$ into being binary.

\section{FRGM in Euclidean Space}\label{section:FRGM-Euclidean}
In many computer vision applications, graphs are often embedded in explicit or implicit manifolds $\mathcal{M}$, {\em e.g.}, Euclidean space $\mathbb{R}^d$ and surface $\mathcal{S}$, where graphs with nodes $\mathcal{V}\in \mathcal{M}$ are naturally associated with some specific geometric properties. For example, the node attributes can be computed as SIFT~\cite{[2004-Lowe]}, shape context~\cite{[2002-Belongie-pami]}, HKS~\cite{[2009-Sun]} and so on, and the edge attribute matrix $\mathbf{E}$ can be computed as Euclidean distance on $\mathbb{R}^d$ or geodesic distance on surface $\mathcal{S}$. 

We can use the proposed method for general GM in Sec.~\ref{sec:FRGM-general} to match graphs in these cases. Furthermore, for graphs embedded in $\mathbb{R}^d$, we can construct another method for Euclidean GM based on the fact that the functional representation of $\mathcal{T}_F$ between abstract function spaces can be deduced into the concrete Euclidean space $\mathbb{R}^d$ with explicit geometric interpretations. Since each node can be represented as a vector $V_j^{(2)}\in \mathbb{R}^d$, the expression $\mathbf{P}_{ij}V_j^{(2)}$ naturally makes sense. Consequently, we can directly define the unknown transformation $\mathcal{T}: \mathcal{V}_1\to \mathcal{V}_2$ in a linear form:
\begin{align}
\mathcal{T}: \mathcal{V}_1 &\to \mathcal{V}_2, \\
V^{(1)}_i\mapsto \mathcal{T}(V^{(1)}_i) &=\sum_{j=1}^{n}\mathbf{P}_{ij}V^{(2)}_j\label{equation4}.
\end{align}
The transformed nodes can be rewritten in a matrix notation $\mathcal{T}(\mathcal{V}_1)\triangleq\mathbf{P}\mathcal{V}_2$ and $\mathcal{T}(V^{(1)}_i)\triangleq(\mathbf{P}\mathcal{V}_2)_{i}$. Now, $\mathbf{P}\in \mathbb{R}^{m\times n}$ is a linear representation map of the unknown transformation $\mathcal{T}$. With the
constraint that $\mathbf{P}\in \hat{\mathcal{P}}$, each node $\mathcal{T}(V^{(1)}_i)$ lies in the convex hull of $\mathcal{V}_2\in \mathbb{R}^d$. Once $\mathbf{P}$ reaches a binary correspondence matrix, $V^{(1)}_i$ is transformed into $V^2_{j'}$, where $\mathbf{P}_{ij'} =1,\mathbf{P}_{i,j\neq j'}=0$.

For graphs embedded in Euclidean spaces, the edge attributes, such as edge length and edge orientation, are widely used. The edge attributes of the transformed graph $\mathcal{T}(\mathcal{V}_1)=\mathbf{P}\mathcal{V}_2$ can be computed as a function w.r.t. $\mathbf{P}$ as:
\begin{itemize}
	\item[-]{\bf\em edge length} computed as the Euclidean distance 
	$$ ||\mathcal{T}(V^{(1)}_{i_1})-\mathcal{T}(V^{(1)}_{i_2})||=||(\mathbf{P}\mathcal{V}_2)_{i_1}-(\mathbf{P}\mathcal{V}_2)_{i_2}||,$$
	\item[-]{\bf\em edge orientation} computed as the vector between nodes
	$$ \overrightarrow{\mathcal{T}(V^{(1)}_{i_1})-\mathcal{T}(V^{(1)}_{i_2})}=\overrightarrow{(\mathbf{P}\mathcal{V}_2)_{i_1}-(\mathbf{P}\mathcal{V}_2)_{i_2}},$$
\end{itemize}
where $||\cdot||$ is the Euclidean $L_2$ norm. We propose our algorithm for matching graphs in Euclidean space, {\em i.e.} {\bf FRGM-E}, in the following sections. 

\vspace{-3mm}
\subsection{Preserving edge-length}
Given two graphs with visually similar structures, a general constraint is to preserve the edge length between the original edge $V^{(1)}_{i_1i_2}\triangleq (V^{(1)}_{i_1},V^{(1)}_{i_2})$ and its corresponding edge $\mathcal{T}(V^{(1)}_{i_1i_2})\triangleq (\mathcal{T}(V^{(1)}_{i_1}),\mathcal{T}(V^{(1)}_{i_2}))\triangleq(\mathbf{P}\mathcal{V}_2)_{i_1i_2}$.
Thus, the pairwise potential of the first objective function can be defined as follows:
\begin{align}
J_{non}(\mathbf{P})
&=\sum_{(i_1,i_2)}\mathcal{E}_{1_{i_1i_2}}(||V^{(1)}_{i_1i_2}||-||\mathcal{T}(V^{(1)}_{i_1i_2})||)^2 \\
&=\sum_{(i_1,i_2)}\mathcal{E}_{1_{i_1i_2}}(||V^{(1)}_{i_1i_2}||-||(\mathbf{P}\mathcal{V}_2)_{i_1i_2}||)^2.
\end{align}
We can add a unary term $\langle \mathbf{P},\mathbf{U}\rangle_F$ computed with node attributes to this pairwise term as follows:
{\small
\begin{align}
	J_{non}(\mathbf{P})&=(1-\lambda_1)\langle \mathbf{P},\mathbf{U}\rangle_F +\lambda_1\sum_{(i_1,i_2)}\mathcal{E}_{1_{i_1i_2}}(||V^{(1)}_{i_1i_2}||-||(\mathbf{P}\mathcal{V}_2)_{i_1i_2}||)^2.
\end{align}}

\begin{figure}[t!]
	\begin{center}
		\subfigure[]
		{\includegraphics[height=0.3\linewidth]{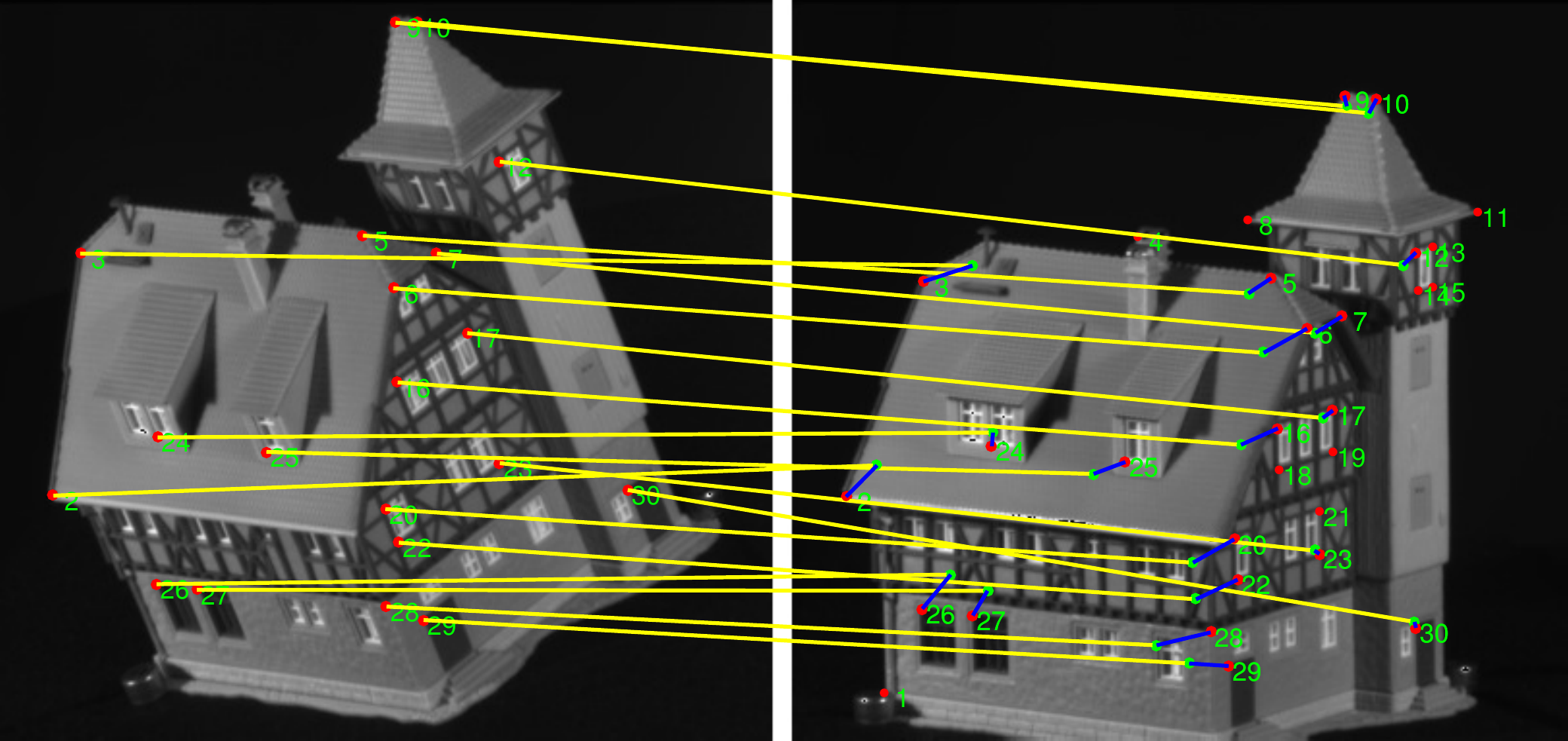}}
		\hspace{1mm}\subfigure[]
		{\includegraphics[height=0.3\linewidth]{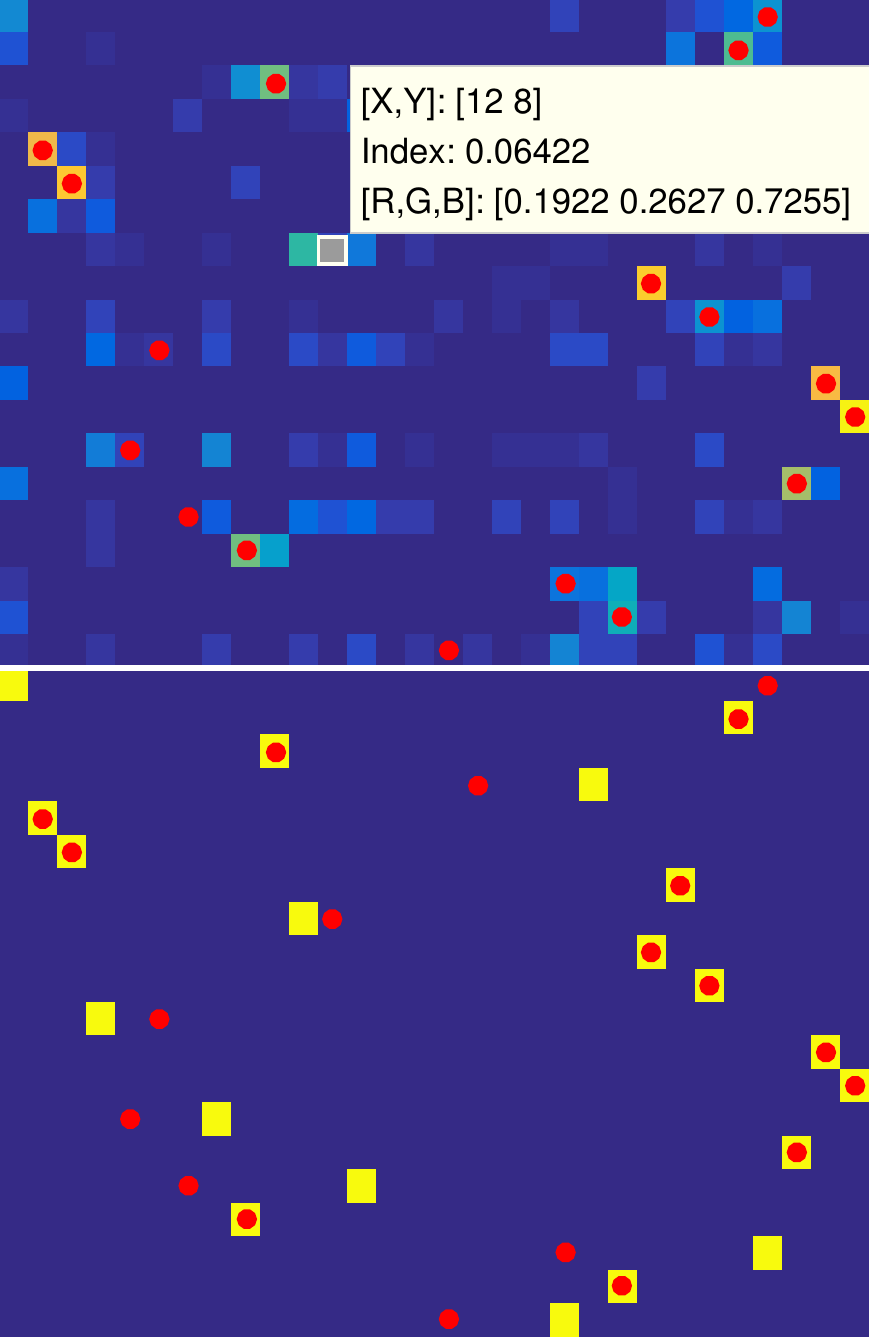}} \\
		\vspace{-2mm}
		\subfigure[]
		{\includegraphics[height=0.3\linewidth]{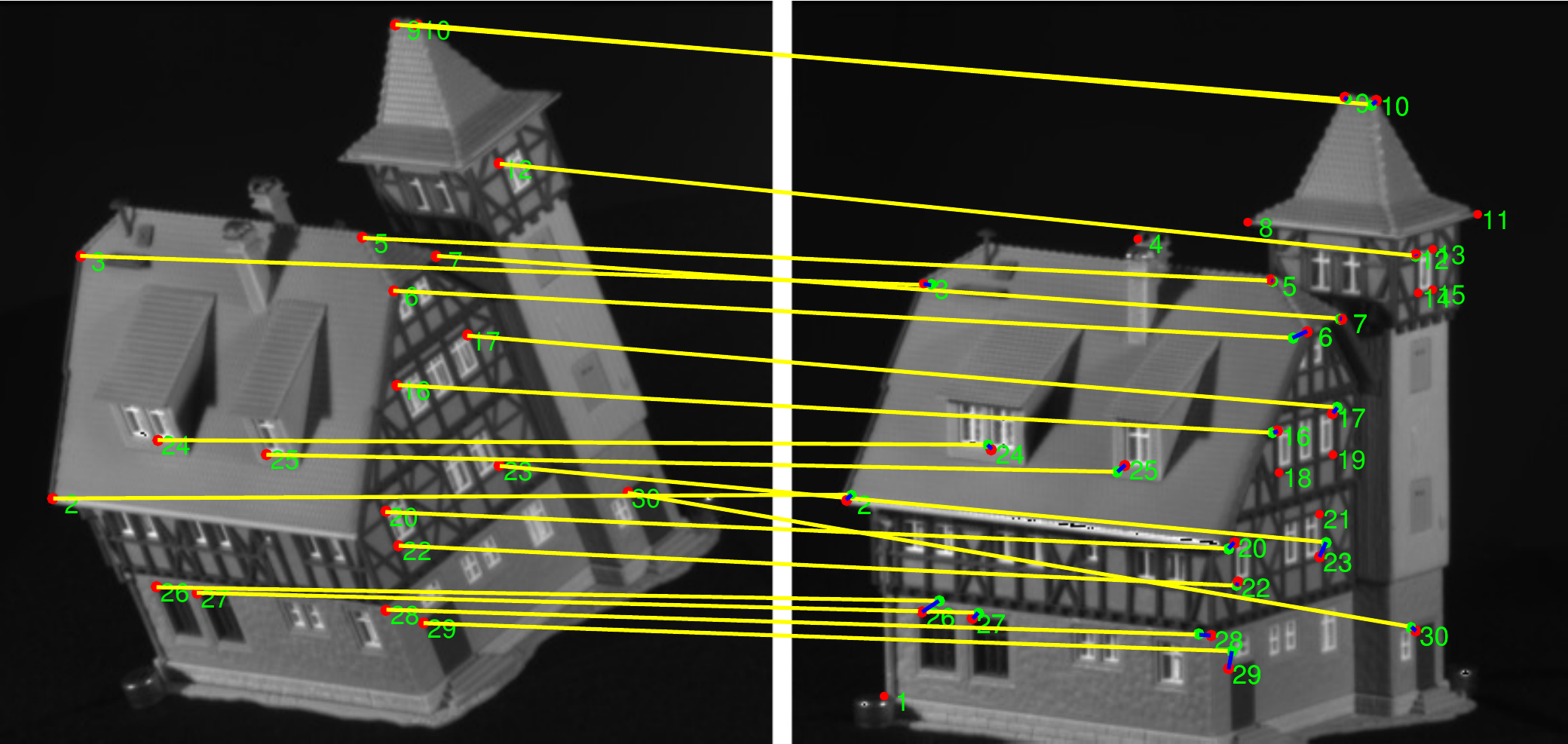}}
		\hspace{1mm}\subfigure[]
		{\includegraphics[height=0.3\linewidth]{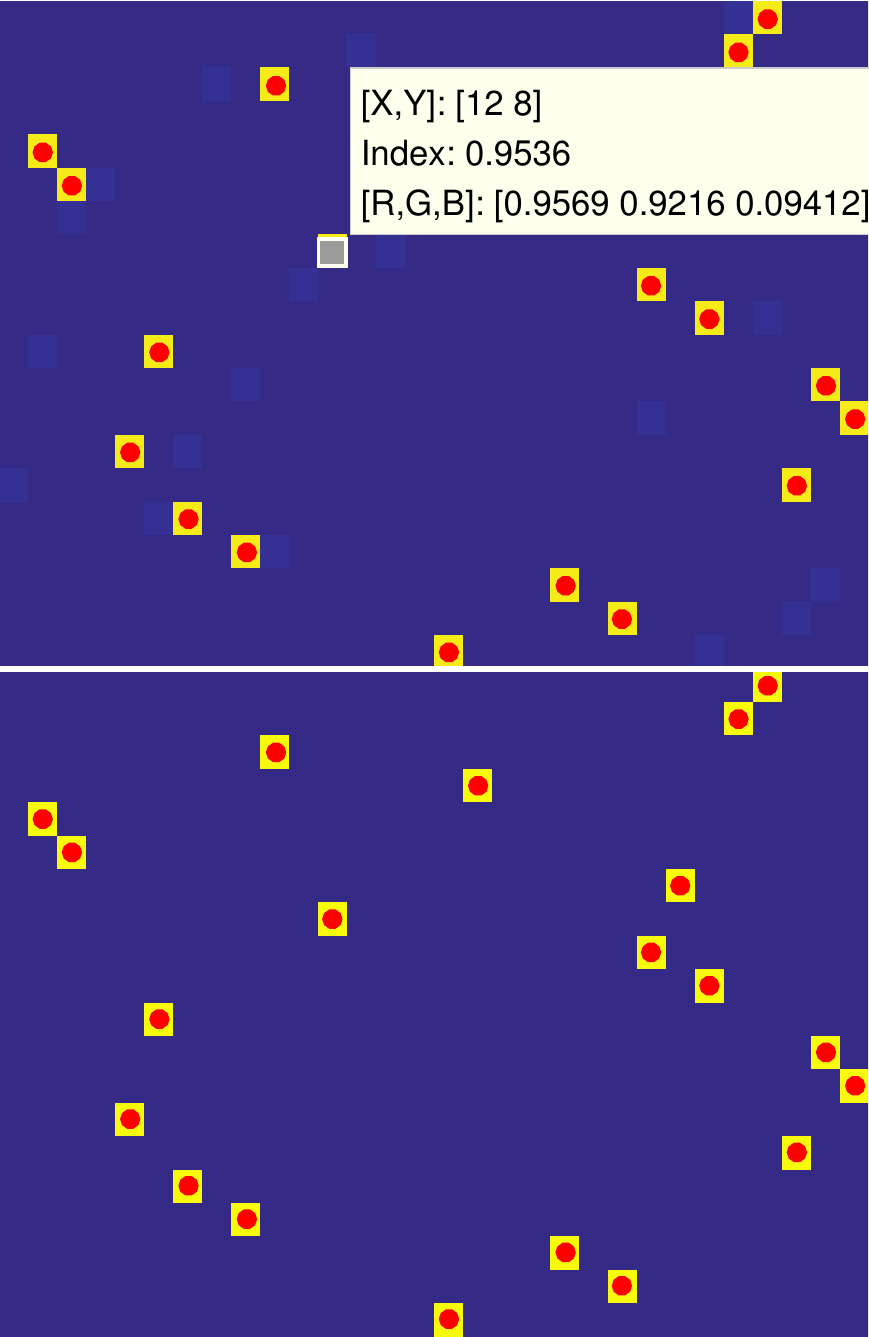}}
	\end{center}
	\vspace{-5mm}
	\caption{(a) Nodes shift after being transformed by minimizing $J_{non}(\mathbf{P})$ in a 20-vs-30 case. The lines in blue are the offset vectors, and the points in green are transformed nodes $\left\{\mathcal{T}(V_i^{(1)})\right\}_{i=1}^m$. 
		(b) Representation map $\mathbf{P}^*_1$ (top) and the post-discretization (bottom) corresponding to (a). (c) Nodes transformed by minimizing $J_{con}(\mathbf{P})$ with almost no offset. 
		(d) Representation map $\mathbf{P}^*_2$ (top) and the post-discretization (bottom) corresponding to (c). In (b) and (d), red points mark the ground-truth correspondence.}
	\label{fig:long}
	\label{fig:longfig1}
	\vspace{-1mm}
\end{figure}

Due to the nonconvexity of $J_{non}(\mathbf{P})$, its solution $\mathbf{P}_1^*\in \hat{\mathcal{P}}$ often reaches a local minimum and is not binary, and the post-discretization procedure will result in low accuracy; see Fig.~\ref{fig:longfig1} (b) for illustration. Consequently, the transformed node $\mathcal{T}(V^{(1)}_i)$ is not exactly equal to a $V^{(2)}_{j}\in \mathcal{V}_2$, and there is often an offset between $\mathcal{T}(V^{(1)}_i)$ and its correct match $V^{(2)}_{\delta_i}$. Fig.~\ref{fig:longfig1} (a) shows this phenomenon, where each $\mathcal{T}(V^{(1)}_i)$ shifts from the correct match $V^{(2)}_{\delta_i}$ to some degree.

\subsection{Reducing node offset}
Benefiting from the property of the solution $\mathbf{P}^*_1$ that preserves the edge length of $\mathcal{G}_1$, the offset vectors of adjacent transformed nodes in $\{(\mathbf{P}^*_1\mathcal{V}_2)_i\}_{i=1}^m$ have similar directions and norms, as shown in Fig.~\ref{fig:longfig1}~(a). To reduce the node offset from $(\mathbf{P}_1^*\mathcal{V}_2)_i$ to the corresponding correct match $V^{(2)}_{\delta_i}$ denoted by
$$
\overrightarrow{(\mathbf{P}_1^*\mathcal{V}_2)_iV^{(2)}_{\delta_i}}=V^{(2)}_{\delta_i}-(\mathbf{P}_1^*\mathcal{V}_2)_i,
$$
we aim to minimize the sum of differences between adjacent offset vectors, {\itshape{i.e.}},
\begin{align}
J_{con}(\mathbf{P})&=\sum_{(i_1,i_2)}\mathbf{S}_{i_1i_2}||\left((\mathbf{P}\mathcal{V}_2)_{i_1}-(\mathbf{P}^*_1\mathcal{V}_2)_{i_1}\right) -\left((\mathbf{P}\mathcal{V}_2)_{i_1}-(\mathbf{P}^*_1\mathcal{V}_2)_{i_1}\right)||^2\nonumber\\
&=\text{Tr}\left((\mathbf{P}\mathcal{V}_2-\mathbf{P}_1^*\mathcal{V}_2)^T\mathbf{L}_{\mathbf{S}}(\mathbf{P}\mathcal{V}_2-\mathbf{P}_1^*\mathcal{V}_2)\right),
\end{align}
where $\mathbf{L}_{\mathbf{S}} = \text{diag}(\mathbf{S}\mathbf{I})-\mathbf{S}$ and $\mathbf{S}\in \mathbb{R}^{m\times m}_+$ is computed to indicate the adjacency relation of node pair $(\mathcal{T}(V^{(1)}_{i_1}),\mathcal{T}(V^{(1)}_{i_2}))$. The undirected graph here will result in a symmetric $\mathbf{\mathbf{S}}$; therefore $\mathbf{L}_{\mathbf{S}}$ is positive-definite and $J_{con}(\mathbf{P})$ is convex.
\begin{figure*}[t!]
	\begin{center}
		{\includegraphics[width=0.99\linewidth]{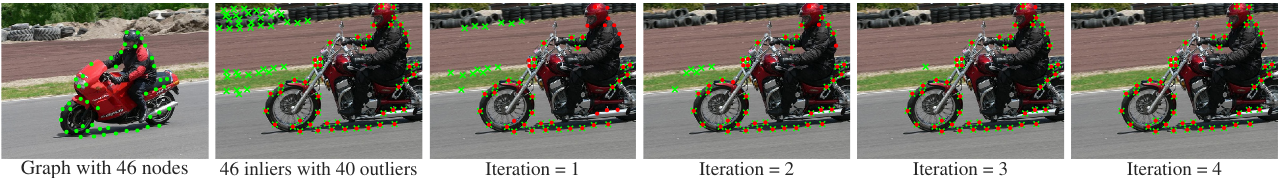}}
	\end{center}
	\vspace{-5mm}
	\caption{Outlier removal with transformation map $\mathbf{P}^*$ obtained by alternately minimizing $J_{non}(\mathbf{P})$ and $J_{con}(\mathbf{P})$. In each iteration, the red dots are inliers, and the green plus signs are the nodes remaining after removal.}
	\label{fig:removal}
\end{figure*}

Compared to the algorithm proposed for general GM in Sec.~\ref{sec:general algo}, the distance matrix $\mathbf{D}$ here can be computed with an explicit geometric interpretation: $\mathbf{D}_{ij}=||\mathcal{T}(V^{(1)}_i)-V^{(2)}_j||$ is the Euclidean distance between the transformed node $\mathcal{T}(V^{(1)}_i)$ and $V^{(2)}_j$. As shown in Fig.~\ref{fig:longfig1}~(c), $||\mathcal{T}(V^{(1)}_i)-V^{(2)}_{\delta_i}||$ is smaller than $||\mathcal{T}(V^{(1)}_i)-V^{(2)}_{j\neq\delta_i}||$, where $V^{(2)}_{\delta_i}$ denotes the correct matching of $V^{(1)}_i$. Therefore, the unary term $\langle \mathbf{P},\mathbf{D}\rangle_F$ can be added as a useful constraint during matching. Finally, $J_{con}(\mathbf{P})$ is summarized as:
\begin{align}
J_{con}(\mathbf{P})&=(1-\lambda_2)\langle\mathbf{P},\mathbf{D}\rangle_F +\lambda_2\text{Tr}\left((\mathbf{P}\mathcal{V}_2-\mathbf{P}_1^*\mathcal{V}_2)^T\mathbf{L}_{\mathbf{S}}(\mathbf{P}\mathcal{V}_2-\mathbf{P}_1^*\mathcal{V}_2)\right).
\end{align}
In general, this objective function $J_{con}(\mathbf{P})$ is solved by a (nearly) binary solution $\mathbf{P}^*_2$ if $\lambda_2 \in [0,1]$ is small. This significantly improves the matching accuracy. See Fig.~\ref{fig:longfig1}~(d) as an example.

\subsection{Explicit outlier-removal  strategy}\label{sec:out-rem}
In practice, outliers generally occur in graphs and affect the matching accuracy. Based on the ability of the optimal representation map $\mathbf{P}^*_1$ and $\mathbf{P}^*_2$ that preserves the geometric structure between $\mathcal{V}_1$ and the transformed graph $\mathbf{P}^*_1\mathcal{V}_2$ or $\mathbf{P}^*_2\mathcal{V}_2$, we can propose an explicit outlier-removal strategy. 

The transformed graph $\mathcal{G}_{\mathcal{T}(\mathcal{V}_1)}$ with nodes $\mathcal{T}(\mathcal{V}_1)=\mathbf{P}^*\mathcal{V}_2$ lies in the convex hull of $\mathcal{V}_2$. In some sense, the operation $\mathcal{T}(\mathcal{V}_1)=\mathbf{P}^*\mathcal{V}_2$ can be viewed as a domain adaptation~\cite{[2017-Courty-pami]} from the source domain $\mathcal{V}_1$ to the target domain $\mathcal{V}_2$. The graph $\mathcal{G}_{\mathcal{T}(\mathcal{V}_1)}$ has a geometric structure similar to the original graph $\mathcal{G}_1$ and lies in the same space of $\mathcal{G}_2$ with a relatively small offset. Then, we can remove outliers adaptively using a ratio test technique. Given two point sets $\mathcal{T}(\mathcal{V}_1)$ and $\mathcal{V}_2$, we compute the Euclidean distance $d_{ij}$ of all the pairs $(\mathcal{T}(V^{(1)}_i),V^{(2)}_{j})$.  For each node $\mathcal{T}(V^{(1)}_i)$, we find the closest node $V^{(2)}_{j^*}$ and remove all the nodes $V^{(2)}_j$ when $d_{ij}>k\cdot d_{ij^*}$ for a given $k>0$. If the number  of remaining nodes $l$ is less than $m$, $m-l$ nodes are selected from the removed ones that are closer to $\mathcal{T}(\mathcal{V}_1)$ and added. See Fig.~\ref{fig:removal} as an example, where after several iterations most outliers are removed.
More experimental results are reported in the experimental section.

\section{FRGM with Geometric Deformation}\label{section:deformal GM}
For Euclidean GM, rigid or nonrigid geometric deformations may exist between graphs. In these cases, we need to estimate both the correspondence and deformation parameters. This section demonstrates that the FRGM can provide a new parameterization of transformation between graphs. Due to the associative law of matrix multiplication, this parameterization is associative with the deformation parameters. Theoretically, this allows us to estimate the correspondence and deformation parameters alternately.

\subsection{Geometric deformation}
Given two point sets $\mathcal{V}_1=\{V^{(1)}_i\}_{i=1}^m, \mathcal{V}_2=\{V^{(2)}_j\}_{j=1}^n \subseteq \mathbb{R}^d$ with geometric transformation $\tau:\mathcal{V}_1\to \mathcal{V}_2$, the task to estimate both the correspondence $\mathbf{P}$ and parameters of $\tau$ is generally formulated as minimizing the sum of residuals:
\begin{equation}
	\min_{\mathbf{P}\in \mathcal{P},\tau\in \chi} J(\mathbf{P},\tau)=\sum_{i,j}\mathbf{P}_{ij}||V^{(1)}_i-\tau(V^{(2)}_j)||^2 + \Upsilon(\tau),
\end{equation}
where $\Upsilon$ is a regularization term of $\tau$. On the one hand, most of the state-of-the-art registration algorithms such as~\cite{[2010-Myronenko],[2011-Jian],[2016-Ma]} do not explicitly recover the correspondence $\mathbf{P}$ as a binary solution. Rather, they estimate $\mathbf{P}$ in a soft way as $\mathbf{P}_{ij}\in [0,1]$ to give $\mathbf{P}_{ij}$ a probability interpretation: $\mathbf{P}_{ij}$ stands for the correspondence probability between $V^{(1)}_i$ and $V^{(2)}_j$. On the other hand, some methods~\cite{[2006-Zheng],[2006-Caetano],[2016-Zhou-pami]} have also been proposed to find the binary correspondence by general GM algorithms. However, these GM-based methods are not consistent with the geometric nature behind the real data. Therefore, they can only handle point sets with simple geometric deformations.

Given a finite point set $\mathcal{V}\subseteq \mathbb{R}^{m\times d}$, the rigid or nonrigid geometric deformation is generally expressed as follows.  
\begin{itemize}
    \item[-]{\em \bf{{Similarity transformation:}}} $\tau(\mathcal{V})=s\mathcal{V}\mathbf{R} + \mathbf{1}_m\mathbf{t}$, where $s\in \mathbb{R}_+$, $\mathbf{R}\in \mathbb{R}^{d\times d}$, and $\mathbf{t}\in \mathbb{R}^{1\times d}$ denote the scaling factor, the rotation matrix and  the translation vector, respectively. Naturally, $\mathbf{R}$ should satisfy the constraint: $\mathbf{R}^T\mathbf{R}=\mathbf{I}_d,\text{det}(\mathbf{R})=1$. 
    \item[-]{\em \bf{{Affine transformation:}}} $\tau(\mathcal{V})=\mathcal{V}\mathbf{A} + \mathbf{1}_m\mathbf{t}$, where $\mathbf{A}\in \mathbb{R}^{d\times d}$ and $\mathbf{t}\in \mathbb{R}^{1\times d}$ denote the affine matrix and the translation vector, respectively.
    \item[-]{\em \bf{{Non-rigid transformation:}}} $\tau(\mathcal{V})=\mathcal{V} + \mathbf{K}\mathbf{W}$, where $\mathbf{K}\in \mathbb{R}^{m\times m}$ is a kernel determined by the basis points $\{V_i\}_i$ and displacement functions $\{\varphi_i\}_i$, and $\mathbf{W}\in \mathbb{R}^{m\times d}$ is a weight matrix that measures the degree of deformation. This definition is based on the radial basis function (RBF) method, which is widely used to parameterize nonrigid transformation. This formulation means that the nonrigid transformation is assumed to be a displacement shifted from its initial position. In this paper, we utilize the Gaussian RBF, {\em i.e.}, $\varphi_i(V)\triangleq \text{exp}({-||V-V_{i}||_2^2/\sigma_w^2})$, where ${\sigma_w}$ is the bandwidth dependent on the degree of deformation. Then, the kernel is computed as $\mathbf{K}_{ij}=\varphi_i(V_j)$. Following some previous works, we set the regularization term as $\Upsilon(\tau)=\text{Tr}(\mathbf{W}^T\mathbf{K}\mathbf{W})$ to penalize the nonsmoothness of nonrigid deformation. 
\end{itemize}

We demonstrate our function-representation-based method for GM with geometric deformations, {\em i.e.} FRGM-D, in the following.

\subsection{Function composition-based method}
With the geometric deformation $\tau:\mathcal{V}_1\to \mathcal{V}_2$, we aim to find a linear representation map $\mathbf{P}$ of matching $\mathcal{T}:\mathcal{V}_1\to \mathcal{V}_2$, which remains consistent with $\tau$ such that the composition of $\mathcal{T}$ and $\tau^{-1}$ is an identity function $\mathbf{I}_{\mathbf{d}}$:
 \begin{equation}
 	\tau^{-1}\circ\mathcal{T}=\mathbf{I}_{\mathbf{d}}:\mathcal{V}_1\to \mathcal{V}_1.
 \end{equation}
According to the associative property of matrix multiplication, the composition $\tau^{-1}\circ\mathcal{T}$ can be rewritten as follows:

\begin{itemize}
    \item[-]{\em For similarity transformation}
    \begin{align}
    \tau^{-1}\circ\mathcal{T}(\mathcal{V}_1) &=\frac{1}{s}(\mathbf{P}\mathcal{V}_2)\mathbf{R}^{-1}-\frac{1}{s}\mathbf{1}_m\mathbf{t}\mathbf{R}^{-1} \nonumber\\
    &=\frac{1}{s}\mathbf{P}(\mathcal{V}_2\mathbf{R^{-1}})-\frac{1}{s}\mathbf{1}_m\mathbf{t}\mathbf{R}^{-1} 
    \end{align}
    
    \item[-]{\em For affine transformation}
    \begin{align}
        \tau^{-1}\circ\mathcal{T}(\mathcal{V}_1)
    &= (\mathbf{P}\mathcal{V}_2)\mathbf{A}^{-1}-\mathbf{1}_m\mathbf{t}\mathbf{A}^{-1} \nonumber\\
    &=\mathbf{P}(\mathcal{V}_2\mathbf{A^{-1}}) -\mathbf{1}_m\mathbf{t}\mathbf{A}^{-1}
    \end{align}
    \item[-]{\em For nonrigid transformation } 
    \begin{align}
    \tau^{-1}\circ\mathcal{T}(\mathcal{V}_1)
    = (\mathbf{P}\mathcal{V}_2)-\mathbf{KW}.
\end{align}
\end{itemize}

Consequently, this associative property allows us to estimate $\mathbf{P}$ and the parameters of $\tau$ alternately.

The alternating estimations of $\mathcal{T}$ and $\tau$ are as follows. First, we  use the identity function $\mathbf{I}_{\mathbf{d}}$ for the initialization of $\tau$. In the alternating steps, once $\tau$ is given, we update the graph as $\mathcal{V}_1\gets\tau(\mathcal{V}_1)$ and then apply our algorithm FRGM-E to find the correspondence between $\mathcal{V}_1$ and $\mathcal{V}_2$. After $\mathbf{P}$ is given, we recover the parameters of $\tau$ by minimizing the objective function as:
{\small
\begin{align}
J(\tau)&=\sum_i||V^{(1)}_i-\tau^{-1}((\mathbf{P}\mathcal{V}_2)_i)||^2 + \lambda \sum_{(i_1,i_2)}\mathcal{E}_{1_{i_1i_2}}||(V^{(1)}_{i_1}-V^{(1)}_{i_2})-\tau^{-1}((\mathbf{P}\mathcal{V}_2)_{i_1}-(\mathbf{P}\mathcal{V}_2)_{i_2})||^2.\nonumber
\end{align}}

Given the correspondence $\mathbf{P}$, the parameters of $\tau$ can be computed in closed form as follows for different transformations:
\begin{itemize}
    \item[-]{\emph{For similarity transformation}}: The optimal translation vector $\mathbf{t}^*$ can be represented as a function of $\mathbf{R}^*$ and $s^*$ as 
\begin{equation}
\mathbf{t}^*=\mathbf{P}{\bar{\mathcal{V}}_2} -{s^*}\bar{\mathcal{V}}_1\mathbf{R}^*, \bar{\mathcal{V}}_1 = \frac{\mathbf{1}_m^T\mathcal{V}_1}{m},\bar{\mathcal{V}}_2 = \frac{\mathbf{1}_n^T\mathcal{V}_2}{n}.
\end{equation}
By the centralization of points, $\mathcal{V}_1\gets \mathcal{V}_1 -\mathbf{1}_m \bar{\mathcal{V}}_1$ and $\mathcal{V}_2\gets \mathcal{V}_2 -\mathbf{1}_n \bar{\mathcal{V}}_2$, we have:
\begin{align}
\mathbf{R}^* 
&= (\mathbf{U}\text{diag}(1,...,|\mathbf{UV}^T|)\mathbf{V}^T)^{-1}.\\
s^* &=\frac{\text{Tr}[(\mathbf{P}\mathcal{V}_2)^T(\mathbf{I}_m+\lambda\mathbf{L}_1)(\mathbf{P}\mathcal{V}_2)]}{\text{Tr}[\mathcal{V}_1^T(\mathbf{I}_m+\lambda\mathbf{L}_1)(\mathbf{P}\mathcal{V}_2)\mathbf{R}^*]},
\end{align}
where $\mathbf{U\Sigma V}^T=\text{svd}(\mathcal{V}_1^T(\mathbf{I}_m+\lambda\mathbf{L}_1)(\mathbf{P}\mathcal{V}_2))$ and $\mathbf{L}_1=\text{diag}(\mathcal{E}_1\mathbf{I})-\mathcal{E}_1$.

\item[-]{\emph{For affine transformation}}: The parameters can be computed as:
\begin{align}
\mathbf{t}^*
&=\mathbf{P}{\bar{\mathcal{V}}_2} -\bar{\mathcal{V}}_1\mathbf{A}^*,\\
\mathbf{A}^*
&=\frac{(\mathbf{P}\mathcal{V}_2)^T(\mathbf{I}_m+\lambda\mathbf{L}_1)(\mathbf{P}\mathcal{V}_2)}{\mathcal{V}_1^T(\mathbf{I}_m+\lambda\mathbf{L}_1)(\mathbf{P}\mathcal{V}_2)},
\end{align}
with centralized points $\mathcal{V}_1\gets \mathcal{V}_1 -\mathbf{1}_m \bar{\mathcal{V}}_1$ and $\mathcal{V}_2\gets \mathcal{V}_2 -\mathbf{1}_n \bar{\mathcal{V}}_2$.

\item[-]{\emph{For nonrigid transformation}}: We choose points $\mathcal{V}_1=\{V^{(1)}_i\}_{i=1}^m$ as the basis points to compute the kernel matrix $\mathbf{K}$. Note that, a regularization term $\sigma^2\text{Tr}(\mathbf{W}^T\mathbf{KW})$ is added to $J(\tau)$. After centralizing the points, the optimal solution $\mathbf{W}^*$ is 
\begin{equation}
	\mathbf{W}^*=-\frac{(\mathcal{V}_1-\mathbf{P}\mathcal{V}_2)^T(\mathbf{I}_m+\lambda\mathbf{L}_1)}{\mathbf{K}(\mathbf{I}_m+\lambda\mathbf{L}_1)+\sigma^2},
	\label{solution:W}
\end{equation}
where $\sigma^2=\frac{1}{mn}\sum_{ij}\mathbf{P}_{ij}||V^{(1)}_i-\mathcal{V}^{(2)}_j||^2$ is used to avoid the singularity of matrix division in Eq.~\eqref{solution:W}.
\end{itemize}

\begin{figure*}[t!]
	\begin{center}		
		{\includegraphics[width=1\linewidth]{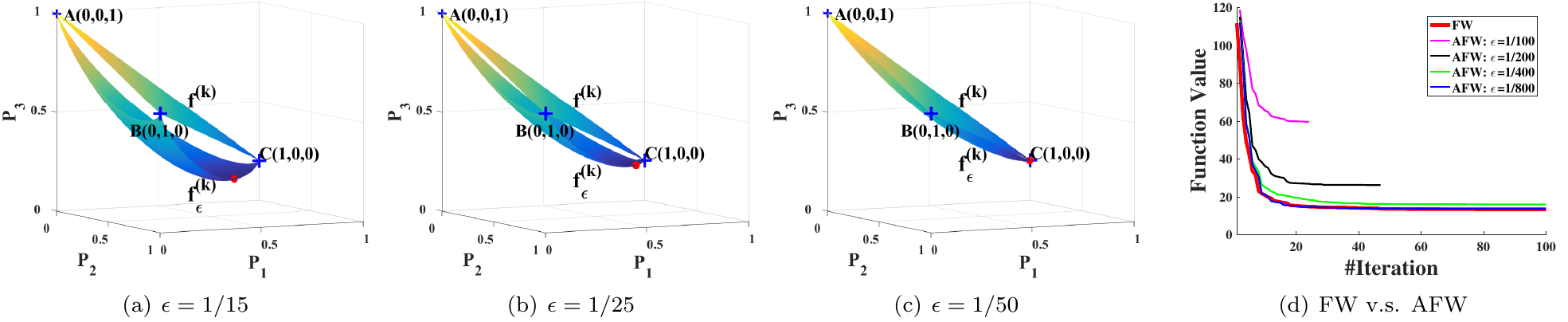}}	
	\end{center}
	\vspace{-5mm}
	\caption{Comparison between FW and AFW. We construct a toy example in (a)--(c), where $\hat{\mathcal{P}}\subseteq \mathbb{R}_+^{1\times 3}$, $\mathbf{P}=(\mathbf{P}_1,\mathbf{P}_2,\mathbf{P}_3)$, and the hyperplane $\nabla f(\mathbf{P}^{k})=(0.3,0.4,0.5)$. There are three extreme points of $\hat{\mathcal{P}}$: ${\bf{A}}=(0,0,1),{\bf{B}}=(0,1,0),{\bf{C}}=(1,0,0)$. From (a) to (b), $\epsilon= \frac{1}{15},\frac{1}{25},\frac{1}{50}$, respectively. $f^{(k)}(\mathbf{P})$ reaches its minimum at ${\bf{C}}=(1,0,0)$, and $\{f^{(k)}_{\epsilon}(\mathbf{P})\}_{\epsilon}$ reach their minima at the red dots. A cool color means a small value. It shows that during the iterations, the solution of $f^{(k)}_{\epsilon}(\mathbf{P})$ gradually approximates the solution of $f^{(k)}(\mathbf{P})$ with $\epsilon$ tending to be smaller. In (d), we show a real example from Sec.~\ref{sec:shape}. The function values of objective functions calculated by AFW tend to be equal to the values obtained by FW when $\epsilon$ becomes smaller.}
	\label{fig:FW-AFW}
\end{figure*}

\section{Numerical Analysis}\label{section:numerical}
In this section, we discuss the optimization strategy for solving the proposed algorithms. We first introduce an efficient optimization algorithm based on the Frank-Wolfe method. Then, we propose an entropy regularization-based approximation of the Frank-Wolfe method to further reduce the time complexity.

\subsection{The Frank-Wolfe method}
The objective functions proposed in the previous sections may be either convex or nonconvex, and the feasible field $\hat{\mathcal{P}}$ is a convex and compact set. The Frank-Wolfe (FW) method (also known as conditional gradient)~\cite{[2015-Simon-nips],[FW2016]} has been well studied for solving constraint convex or nonconvex optimization problems with at least a sublinear convergence rate.

Given that $f(\mathbf{P})$ is differentiable with an
$L$-Lipschitz gradient and $\hat{\mathcal{P}}$ is a convex and compact set, the FW method iterates the following steps until it converges:
\begin{align}
\tilde{\mathbf{P}}^{(k+1)}&\in \mathop{\text{argmin}}\limits_{{\mathbf{P}}\in \hat{\mathcal{P}}}f^{(k)}(\mathbf{P})\triangleq\langle\nabla f(\mathbf{P}^{(k)}),\mathbf{P}\rangle,\label{eq:fw1}\\
\mathbf{P}^{(k+1)}&=\mathbf{P}^{(k)} + \alpha^{(k)}(\tilde{\mathbf{P}}^{(k+1)}-\mathbf{P}^{(k)}),\label{eq:fw2}
\end{align}
where $\alpha^{(k)}$ is the step size obtained by exact or inexact line search~\cite{[Goldstein-1965]}, and $\nabla f(\mathbf{P}^{(k)})$ is the gradient of $f$ at $\mathbf{P}^{(k)}$. 

In Eq.~\eqref{eq:fw1}, the minimizer $\tilde{\mathbf{P}}^{(k+1)} \in \hat{\mathcal{P}}$ is theoretically an extreme point of $\hat{\mathcal{P}}$ (thus, it is binary). This means that $\tilde{\mathbf{P}}^{(k+1)} \in\mathcal{P}$. Therefore, Eq.\eqref{eq:fw1} is a linear assignment problem (LAP) that can be efficiently solved by approaches such as the Hungarian~\cite{[2010-Kuhn]} and LAPJV~\cite{[1987-Jonker]} algorithms. Moreover, since $\tilde{\mathbf{P}}^{(k+1)}$ is binary in each iteration, the final solution $\mathbf{P}^*$ can be (nearly) binary.

\vspace{1mm}
{\bf Time complexity of the FW method.} The time complexity can be roughly calculated as $\mathbf{O}\left(T(\tau_{f} +\tau_{l}) + \tau_s + \tau_{h}\right)$, where $\tau_s=\mathbf{O}(mn+m^2+n^2)$ is the cost of the unary term and edge attributes for graphs, $\tau_h=\mathbf{O}(n^3)$ is the cost of the Hungarian algorithm used as a post-discretization step, $T$ is the number of iterations. In each iteration, $\tau_{f}=\mathbf{O}(m^2n)$ is the cost to compute the gradient, function value and step size at $\mathbf{P}^{(k)}$, and $\tau_{l}=\mathbf{O}(n^3)$ is the cost to compute LAP using the Hungarian or LAPJV algorithm. Note that since $\tau_f=\mathbf{O}(m^2n)$ is computed in closed form, it takes much less time compared to $\tau_l=\mathbf{O}(n^3)$. Since $m\leq n$, the time complexity approximately equals $\mathbf{O}(Tn^3)$ with the maximum number of iterations $T$.

\subsection{A fast approximated FW method}
Similar to the above analysis, the time complexity of applying the FW method to solve FRGM-D is roughly $\mathbf{O}(kTn^3)$, where $k$ is the alternations to estimate the correspondence $\mathbf{P}$ and transformation $\tau$. Note that we can neglect the computation for the parameters of $\tau$ because it is calculated in closed form and is much less than the cost of computing $\mathbf{P}$. Therefore, the time complexity $\mathbf{O}(kTn^3)$ is mainly caused by solving the LAP. To achieve faster execution and proper approximation, we approximate the original Frank-Wolfe method based on the generalized conditional gradient algorithm~\cite{[2009-Bredies]} by adding a convex entropy regularization term in each iteration of solving LAP. The approximated Frank-Wolfe method (AFW) is defined as follows:
\begin{align}
\hat{\mathbf{P}}^{(k+1)}&\in \mathop{\text{argmin}}\limits_{\mathbf{P}\in \hat{\mathcal{P}}} f^{(k)}_{\epsilon}(\mathbf{P})\triangleq \langle\nabla f(\mathbf{P}^{(k)}),\mathbf{P}\rangle -\epsilon H(\mathbf{P}),\label{FW-en}\\
\mathbf{P}^{(k+1)}&=\mathbf{P}^{(k)} + \hat{\alpha}^{(k)}(\hat{\mathbf{P}}^{(k+1)}-\mathbf{P}^{(k)}),
\end{align}
where $H(\mathbf{P})=-\sum_{ij}\mathbf{P}_{ij}\text{log}(\mathbf{P}_{ij})$ is the entropy of $\mathbf{P}$.
To minimize Eq.~\eqref{FW-en}, we can use the Sinkhorn method~\cite{[2013-Cuturi]} as a fast implementation, which has a time complexity of $\mathbf{O}(mn)$.

 With the entropy regularization $H(\mathbf{P})$, we can approximate the original FW method well within a given tolerance:
\begin{prop}
	The solution $\hat{\mathbf{P}}^{(k+1)}$ tends to $\tilde{\mathbf{P}}^{(k+1)}$  as $\epsilon \to 0$.
	\begin{equation}
		||\hat{\mathbf{P}}^{(k+1)}-\tilde{\mathbf{P}}^{(k+1)}||\leq \frac{\sqrt{mn}}{\epsilon}e^{-\frac{c}{\epsilon}},
	\end{equation}
	where $c\in [0,1]$ is a constant dependent on $m,n$ and $\nabla f(\mathbf{P}^{(k)})$.
\end{prop}
The proof can be given based on the primal-dual method for linear programming~\cite{[1994-Cominetti]}. Therefore, with $\epsilon > 0$ small enough, we can obtain a good approximation. We can prove that the AFW method achieves at least a sublinear convergence rate inspired by ~\cite{[2015-Simon-nips]}.

\begin{prop} Assume that $f(\mathbf{P})$ is differentiable with an $L$-Lipschitz gradient; by choosing a series of $\epsilon_k\leq \frac{1}{k+1}$, the AFW method ensures at least a sublinear convergence rate.
 \begin{equation}
 	0\leq f(\mathbf{P}^{(k+1)}) - f(\mathbf{P}^*)\leq \frac{2(LC^2+m\text{log}(m))}{k+2},
 \end{equation}
 where $\mathbf{P}^*$ is the ideal solution of $f(\mathbf{P})$ and $C=\text{diam}(\hat{\mathcal{P}})$ is the diameter of the feasible field $\hat{\mathcal{P}}$.
\end{prop}

Another reason for choosing $H(\mathbf{P})$ is that $H(\mathbf{P})=0$ when $\mathbf{P}$ is an extreme point of $\hat{\mathcal{P}}$, {\em i.e.}, a binary correspondence between graphs. This means that $f^{(k)}_{\epsilon}(\mathbf{P})$ has the same function value as $f^{(k)}(\mathbf{P})$ at any  extreme point. See Fig.~\ref{fig:FW-AFW} for an illustration.

{\bf{Time complexity of the AFW method}} Similar to the FW method, the time complexity of the AFW method can be roughly calculated as $\mathbf{O}\left(T(\tau_{f} +\tau'_{l}) + \tau_s + \tau_{h}\right)$, where $\tau'_{l}=\mathbf{O}(mn)$ is the cost to compute Eq.~\eqref{FW-en} using the Sinkhorn method.

\section{Experimental analysis}\label{sec:experiment}
In this section, we evaluate our functional-representation-based GM methods, {\em i.e.} the general GM (FRGM-G), Euclidean GM (FRGM-E) and deformable GM (FRGM-D) algorithms. 

In Sec.~\ref{section:face}--Sec.~\ref{section:real}, we compare FRGM-G and FRGM-E to several state-of-the-art GM algorithms, including GA~\cite{[1996-Gold]}, PM~\cite{[2008-Zass-cvpr]}, SM~\cite{[2005-Leordeanu]}, SMAC~\cite{[2006-Cour-nips]}, IPFP-S~\cite{[2009-Leordeanu-nips]}, RRWM~\cite{[2010-Cho-eccv]}, FGM-D~\cite{[2016-Zhou-pami]} and MPM~\cite{[2014-Cho-cvpr]}. In Sec.~\ref{sec:shape}, we compare FRGM-D with several state-of-the-art point registration algorithms, including GLS~\cite{[2016-Ma]}, GMM~\cite{[2011-Jian]}, CPD~\cite{[2010-Myronenko]}. Note that in Sec.~\ref{sec:shape} where we evaluate our FRGM-D on geometrically deformed graphs, we do not compare the GM algorithms used in Sec.~\ref{section:face}--Sec.~\ref{section:real} because those methods can neither be directly used for this task nor handle graphs with significant geometric deformations. All comparisons are conducted on both synthetic and real-world datasets that are commonly used to evaluate GM or point registration algorithms. We obtained the codes of the compared methods from the author's websites and implemented all the experiments on a desktop with a 3.5GHz Intel Xeon CPU E3-1240 and 16 GB memory.

\subsection{Results on 3D face}\label{section:face}
This section aims to evaluate our algorithm FRGM-G. We conducted two experiments on graphs in a low-dimensional manifold, {\em i.e.}, 3D face. In part one, we implemented FRGM-G on graphs with varying edge densities. In part two, we compared FRGM-G with other state-of-the-art GM methods on complete graphs. 

\begin{figure}[t!]
	\centering
	\subfigure
	{\includegraphics[width=0.95\linewidth]{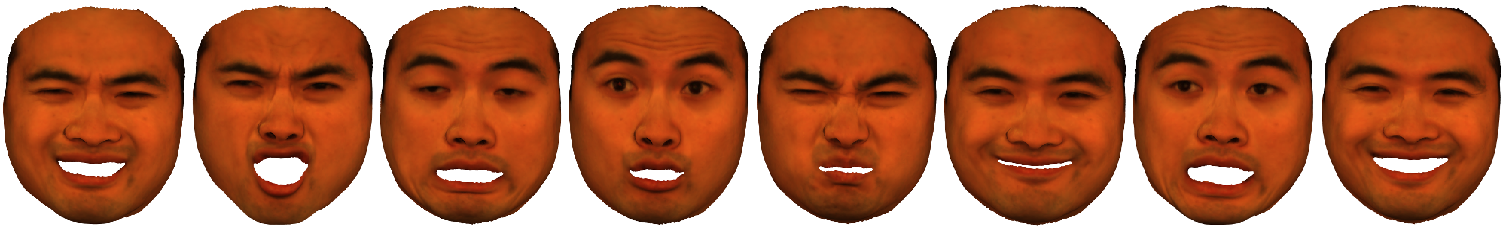}}
	\vspace{-2mm}
	\caption{Some instances of 3D faces with  gradually changing expressions used for the evaluation.}
	\label{fig:face_em}
\end{figure}

This experiment was performed on $383$ continuous frames of 3D faces~\cite{[2004-Zhang]} with gradually changing expressions. We selected $38$ frames whose expressions were more dissimilar to each other, and each frame was marked with 50 landmarks as the ground-truth. Some examples are shown in Fig.~\ref{fig:face_em}. For each pair of faces, we construct two graphs $\mathcal{G}_1$ and $\mathcal{G}_2$ with node attributes $\{\mathbf{v}^{(1)}_i\}_{i=1}^m$ and $\{\mathbf{v}^{(2)}_j\}_{j=1}^n$ consisting of the HKS~\cite{[2009-Sun]} feature descriptors. The edge attribute matrices $\mathbf{E}_1$ and $\mathbf{E}_2$ were computed as the geodesic distance between graph nodes on the faces. For the implementation of FRGM-G, the unary term measuring node dissimilarity is computed as $\mathbf{U}_{ij}\triangleq||\mathbf{v}^{(1)}_i-\mathbf{v}^{(2)}_j||$. We updated the raw matrices $\mathbf{E}_1$ and $\mathbf{E}_2$ into $\hat{\mathbf{E}}_{1}=\text{exp}(-{\mathbf{E}_1^2}/{0.5^2})$
and $\hat{\mathbf{E}}_{2}=\text{exp}(-{\mathbf{E}_2^2}/{0.5^2})$ to honor the inner product illustrated in {\bf Proposition  }\ref{prop:inner}. Then, we used $\hat{\mathbf{E}}_{1}$ and $\hat{\mathbf{E}}_{2}$ to compute the functionals, {\em i.e.}, inner products $\mathbb{F}_{\mathcal{V}_1}(\cdot,\cdot)$ and $\mathbb{F}_{\mathcal{V}_2}(\cdot,\cdot)$ defined in Eq.\eqref{eq:inner}. We used the metric defined in \eqref{eq:metric} to compute $\mathbf{D}$, and we chose the parameters $\alpha_1=0.99,\alpha_2 = 0.5$.

\begin{figure}[t!]
	\centering
	\begin{minipage}{0.42\linewidth}
		\centering
	{\includegraphics[width=0.87\linewidth]{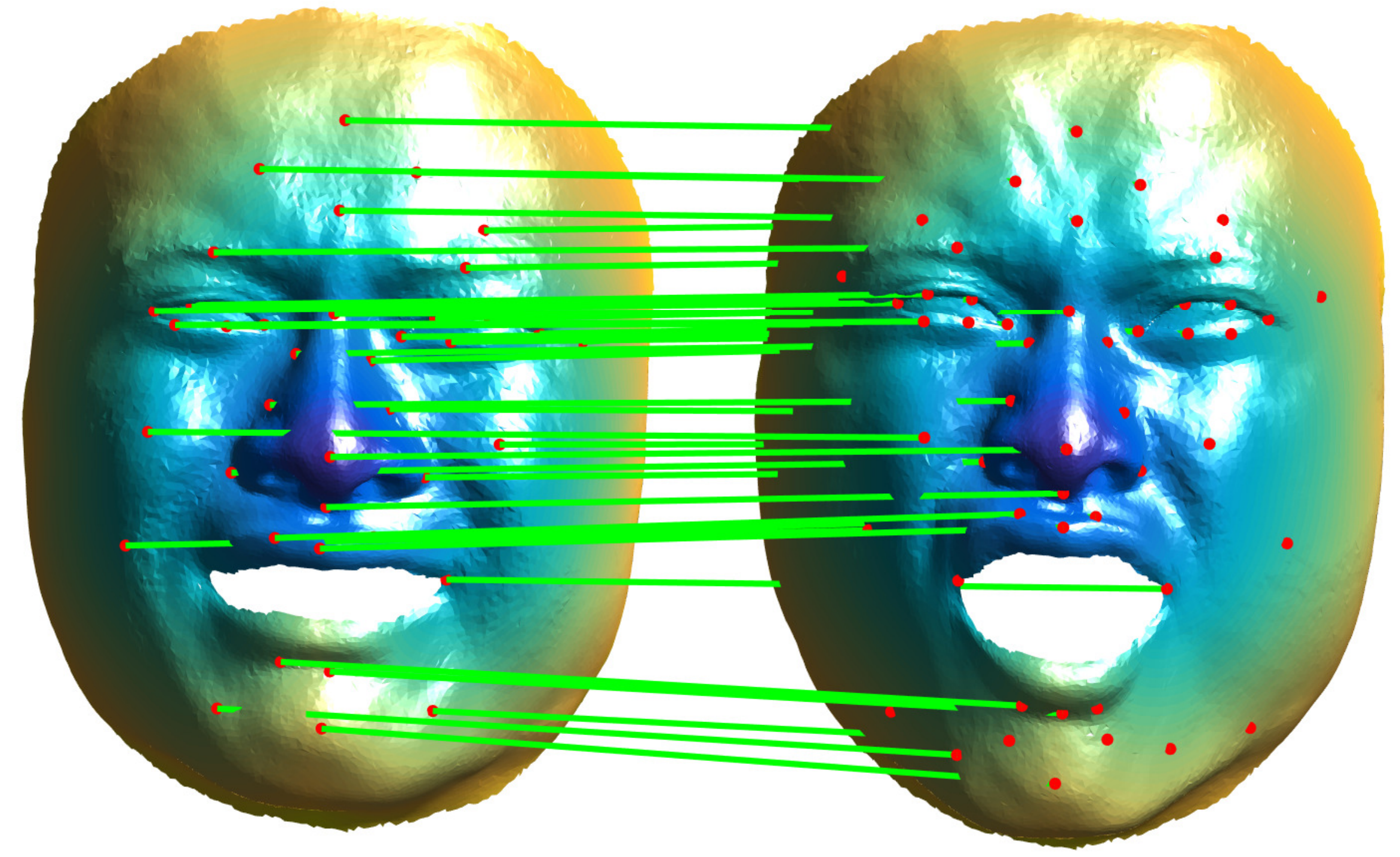}}
	\end{minipage}							
	\begin{minipage}{0.55\linewidth}
		\centering
	{\includegraphics[width=0.67\linewidth]{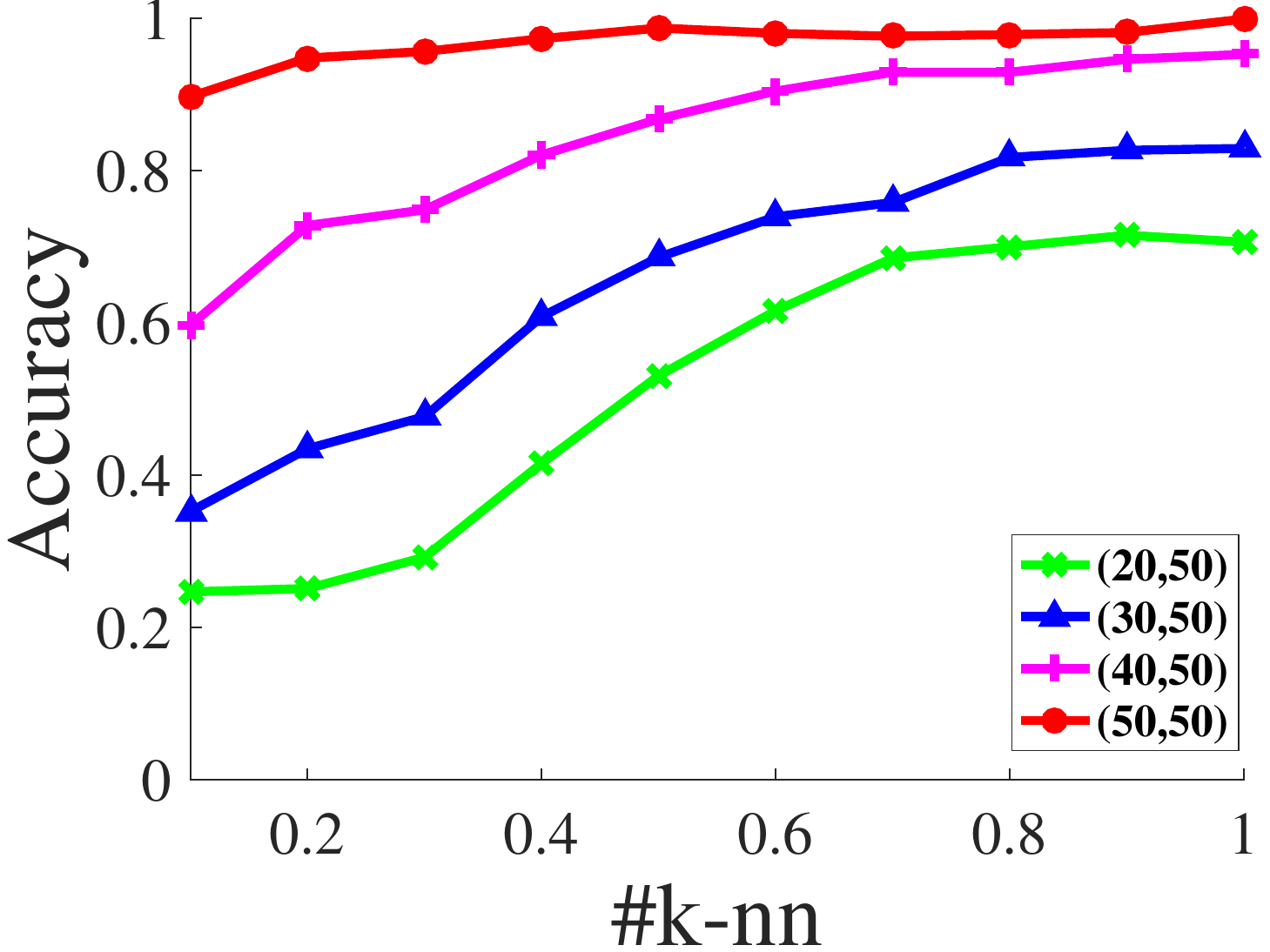}}
	\end{minipage}
\caption{Left: an example of unequal-sized graph pairs with sizes $(40,50)$. Right: results on $k$-nn connected graph pairs. For each node in a graph pair with sizes $(m,n)$, there are $10\%,20\%,...,100\% \times (m,n)$ nodes connected to generate edges.}
\label{fig:face_knn}
\vspace{-3mm}
\end{figure}

In the first experiment, to evaluate FRGM-G on graphs with varying edge densities, we used $k$-nn graphs, {\em i.e.}, each node was connected by the $k$-nearest neighborhood nodes to generate adjacency matrices $\mathcal{E}_1$ and $\mathcal{E}_2$. The edge density of the graph was determined by $k$, which was set to $10\%,20\%,...,100\%$ of the number of nodes. The results are shown in Fig.\ref{fig:face_knn}. For both equal-sized and unequal-sized graph pairs, FRGM-G achieves higher accuracy with more edges because the geometric structures such as inner product or metric will be more complete with more edges.

\begin{table}[t!]
	\centering
	\small
		\caption{Comparison results of average accuracy $(\%)$ on the 3D face dataset.}
	\begin{tabular}{c|cccccc}
		\toprule[1.0pt]
		\diagbox[width=15mm,trim=l]{Method}{Size} 
		& (25,50) & (30,50) & (35,50) & (40,50) & (45,50) & (50,50) \\
		\hline
		{GA ~\cite{[1996-Gold]}}
		&  12.00   & 15.41   & 21.39  & 29.26  &  39.94  &  50.54  \\
		\hline
		{PM ~\cite{[2008-Zass-cvpr]}}
		&  6.70    & 7.03    & 11.27  & 16.01  &  22.16  &  33.30  \\
		\hline	
		{SM ~\cite{[2005-Leordeanu]}}
		& 11.24    & 11.62   & 16.76  & 23.18  &  37.96  &  54.59  \\
		\hline	
		{SMAC ~\cite{[2006-Cour-nips]}}
		& 26.49    & 33.69   & 45.87  & 58.11  &  68.41  &  82.43  \\
		\hline
       {IPFP-S ~\cite{[2009-Leordeanu-nips]}}
		& 11.35    & 7.12    & 5.95   & 10.07  &  7.63   &  69.19  \\
		\hline
	    {RRWM ~\cite{[2010-Cho-eccv]}}
		& 19.14    & 28.56   & 44.71  & 55.47  &  65.59  &  90.05  \\
		\hline
		{FGM-D ~\cite{[2016-Zhou-pami]}}
		& 46.81    & 59.91   & 75.06  & 84.66  &  92.01  &  99.78  \\
		\hline
		{MPM \cite{[2014-Cho-cvpr]}}
		& 8.32     & 11.17   & 16.76  & 29.39  &  40.66  &  52.76  \\
		\hline
		{\bf FRGM-G }
		&\bf 76.14   & \bf 79.88 &\bf 89.16  &\bf 91.58  &\bf 98.04 &\bf 100.00\\
		\bottomrule[1.0pt]
	\end{tabular}
	\label{tab:face_com}
\end{table}

In the second experiment, we compared FRGM-G and the other GM algorithms with complete graphs of sizes $(m,n)$. For the compared methods, the node affinity was computed as $\mathbf{K}_{ij;ij}=\text{exp}({-{||\mathbf{v}^{(1)}_i-\mathbf{v}^{(2)}_j||}/{500}})$, and the edge affinity was computed as $\mathbf{K}_{i_ij_i;i_2j_2}=\text{exp}({-(\mathbf{E_{1_{i_1i_2}}}-\mathbf{E_{2_{j_1j_2}}})^2/2500})$, as used in~\cite{[2016-Zhou-pami]}. The comparison results are shown in Tab.~\ref{tab:face_com}. Because the geometric structures defined on function spaces of graphs are more efficient for representing the distinguishing feature of graphs, our proposed algorithm FRGM-G achieves much higher average accuracy in both equal-sized and unequal-sized cases.

\begin{figure*}[htb!]
	\begin{center}
		{\includegraphics[width=0.7\linewidth]{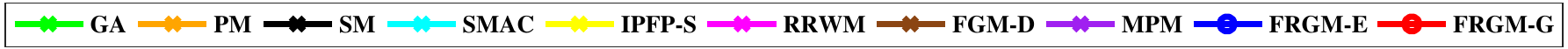}}
		
		\subfigure[]
		{\includegraphics[width=0.22\linewidth]{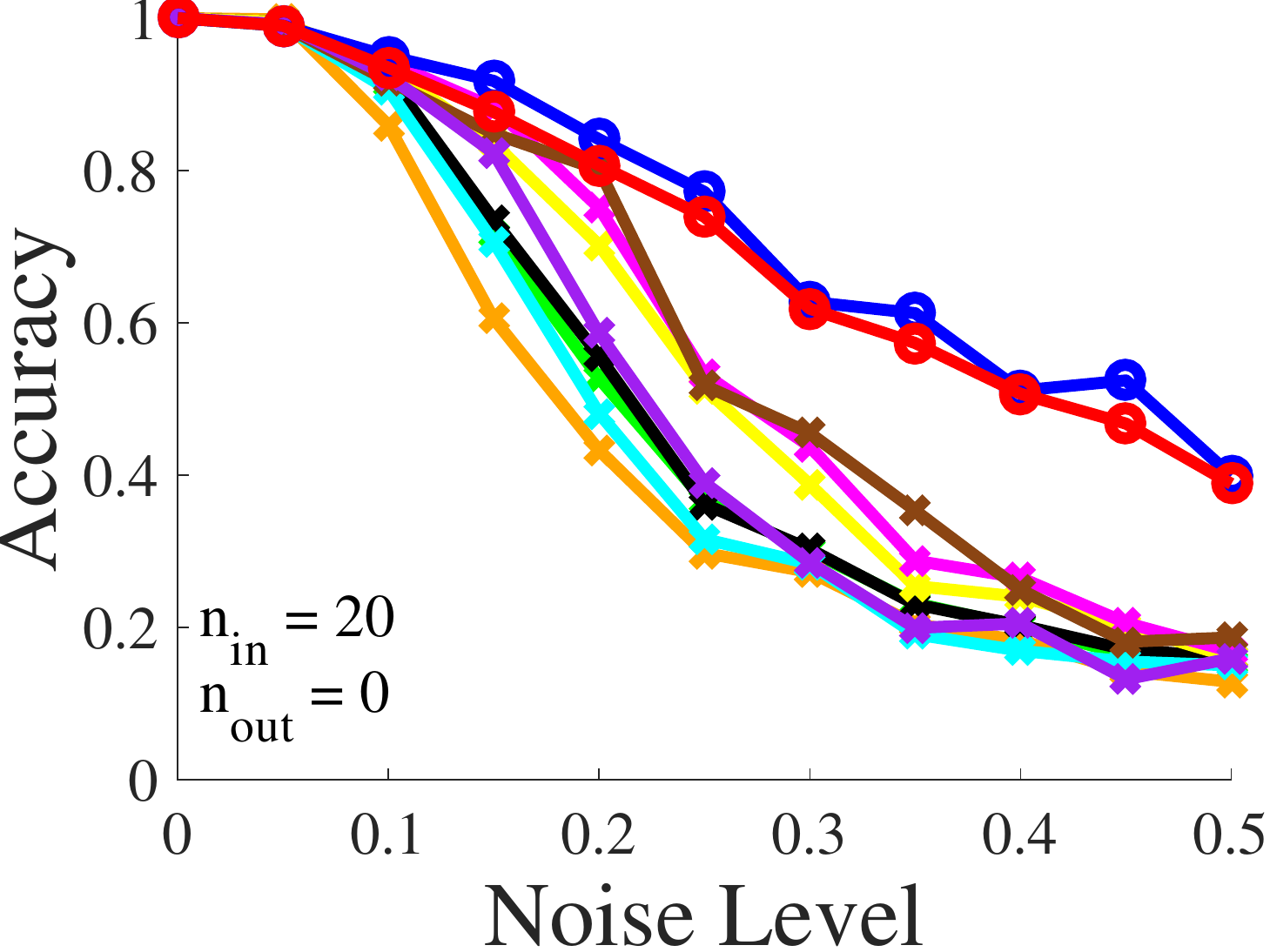}}
		\hspace{3mm}\subfigure[]
		{\includegraphics[width=0.22\linewidth]{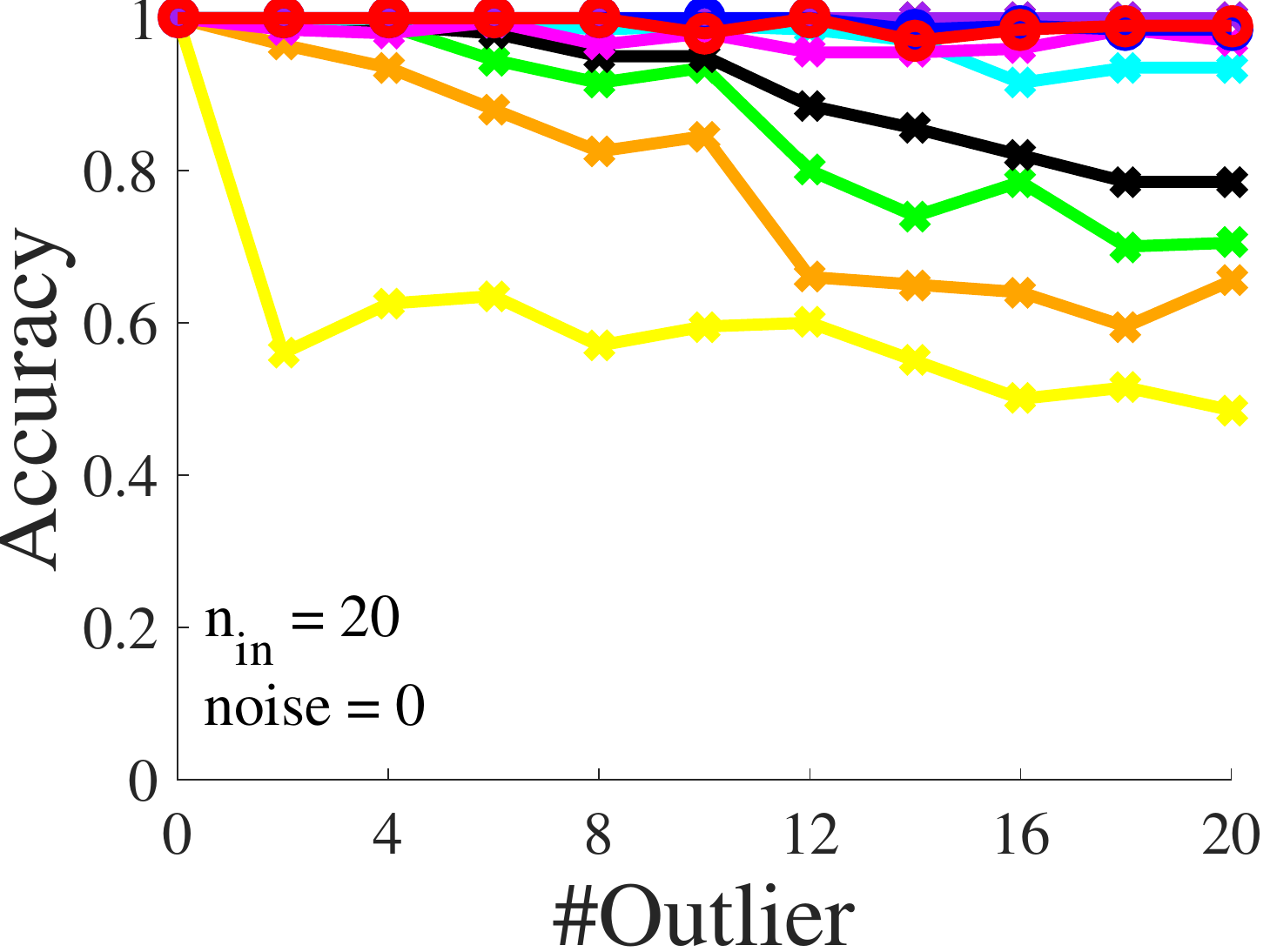}}
		\hspace{3mm}\subfigure[]
		{\includegraphics[width=0.22\linewidth]{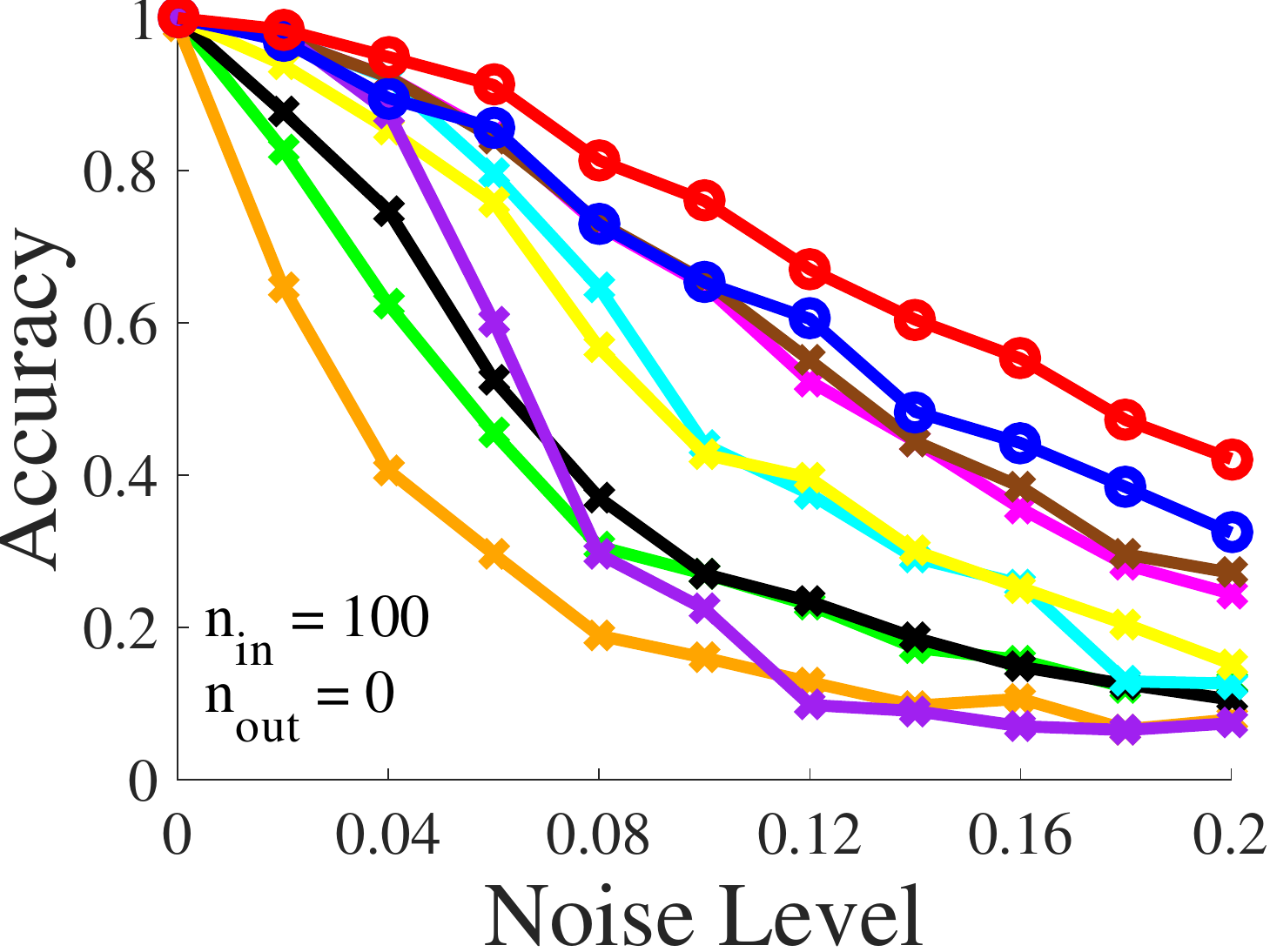}}
		\hspace{3mm}\subfigure[]
		{\includegraphics[width=0.22\linewidth]{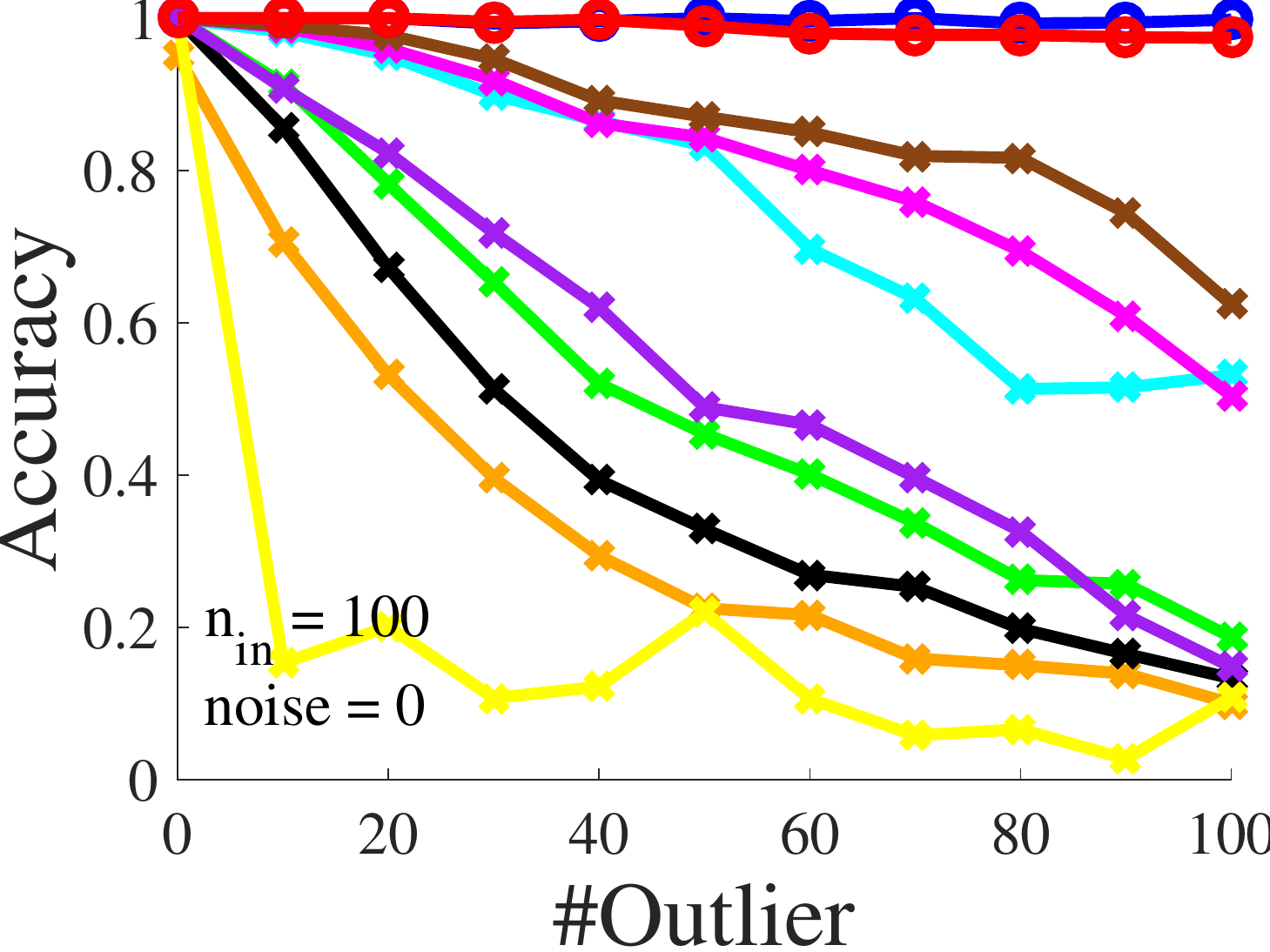}}
	\end{center}
	\vspace{-5mm}
	\caption{Comparisons of the robustness to noise and outliers. For complete graphs, the accuracies with respect to the noise and number of outliers are shown in (a) and (b), respectively. The results for graphs connected by Delaunay triangulation are shown in (c) and (d). FRGM-G and FRGM-E outperform all the others for graphs with noise and outliers.}
	\label{fig:syn_menory}
\end{figure*}
\begin{figure*}[htb!]
	\begin{center}
		{\includegraphics[width=0.7\linewidth]{./Images/cmu_legend.pdf}}
		
		\subfigure[]
		{\includegraphics[width=0.22\linewidth]{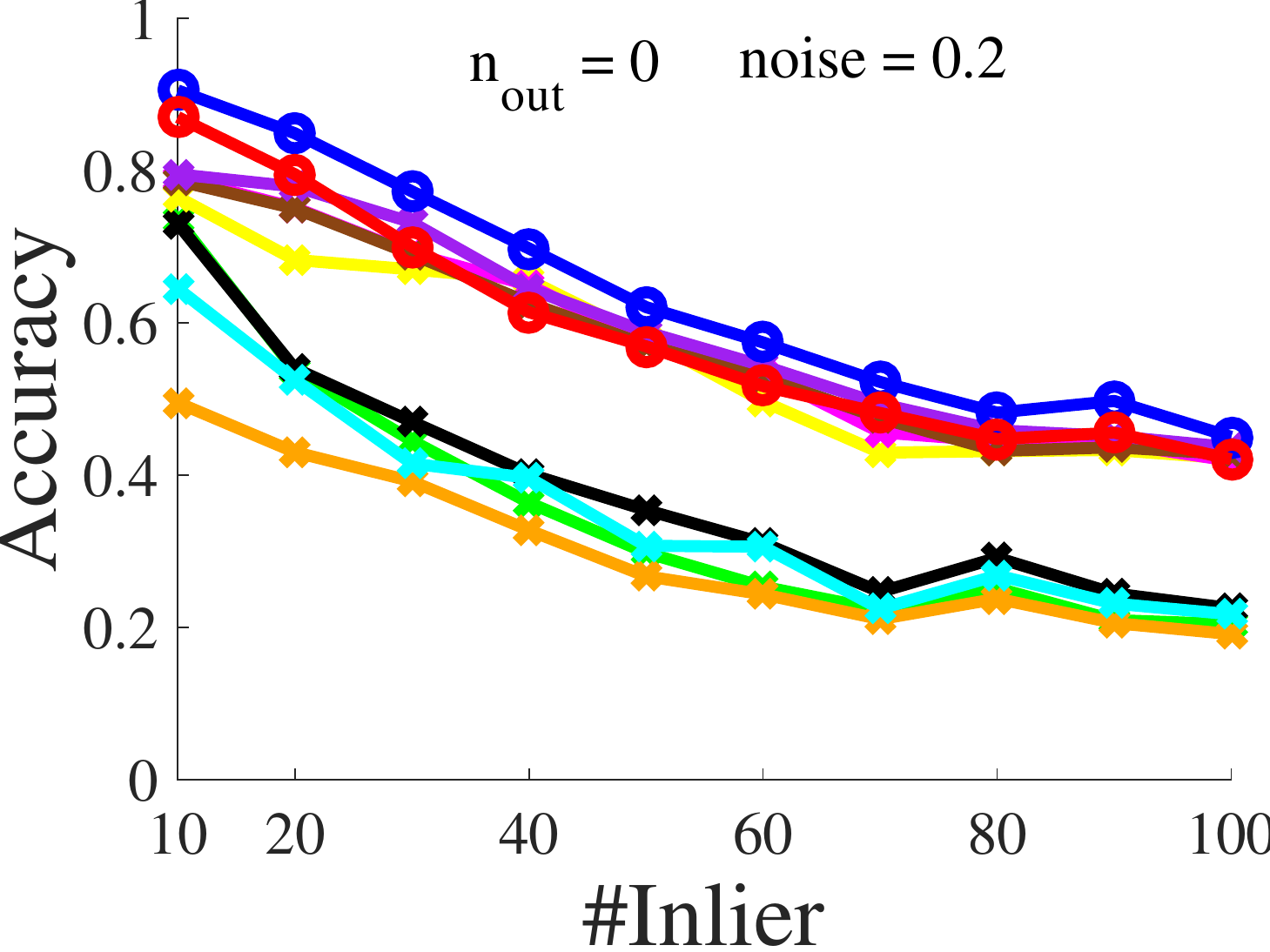}}
		\hspace{3mm}\subfigure[]
		{\includegraphics[width=0.22\linewidth]{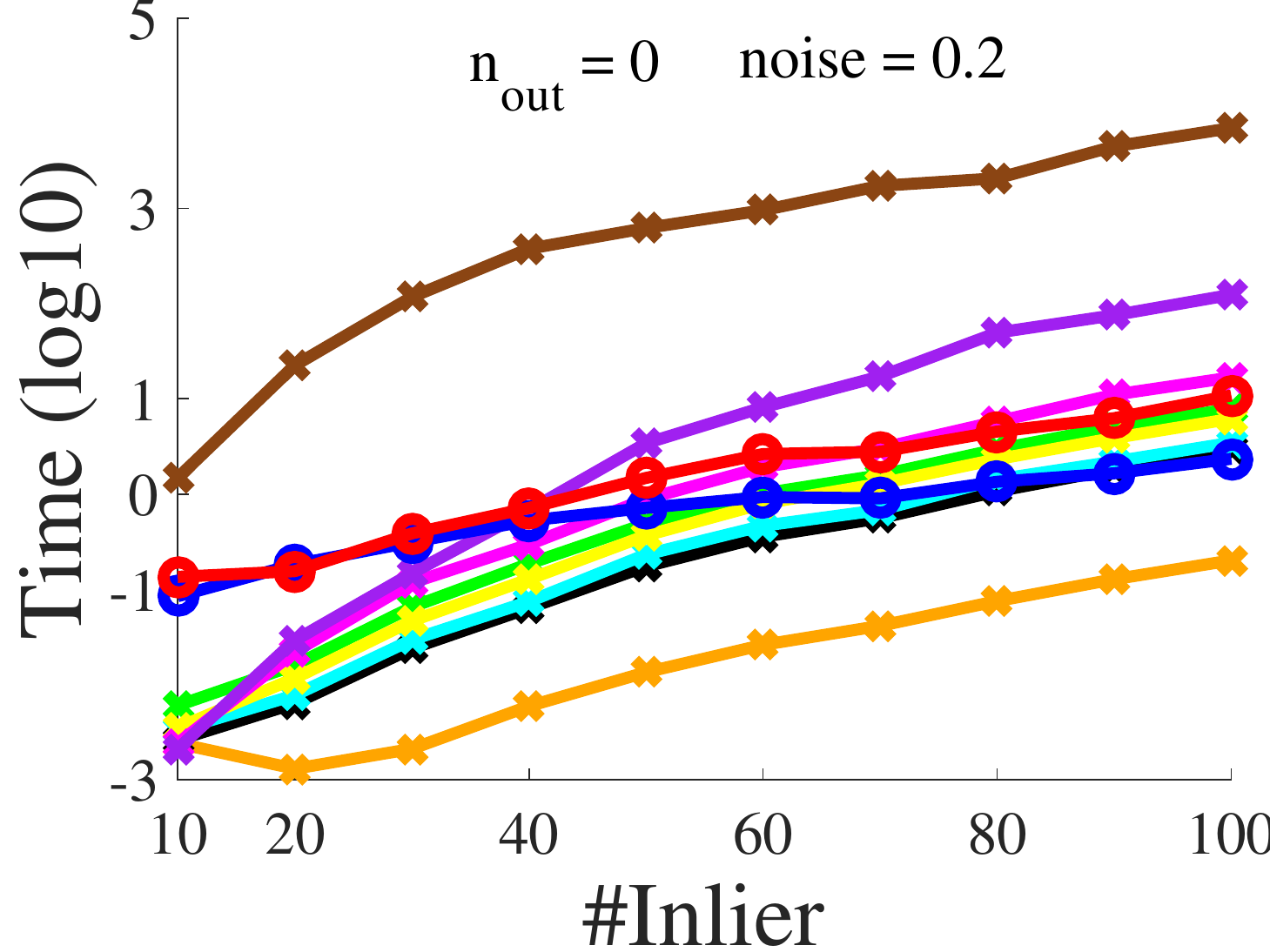}}	
		\hspace{3mm}\subfigure[]
		{\includegraphics[width=0.22\linewidth]{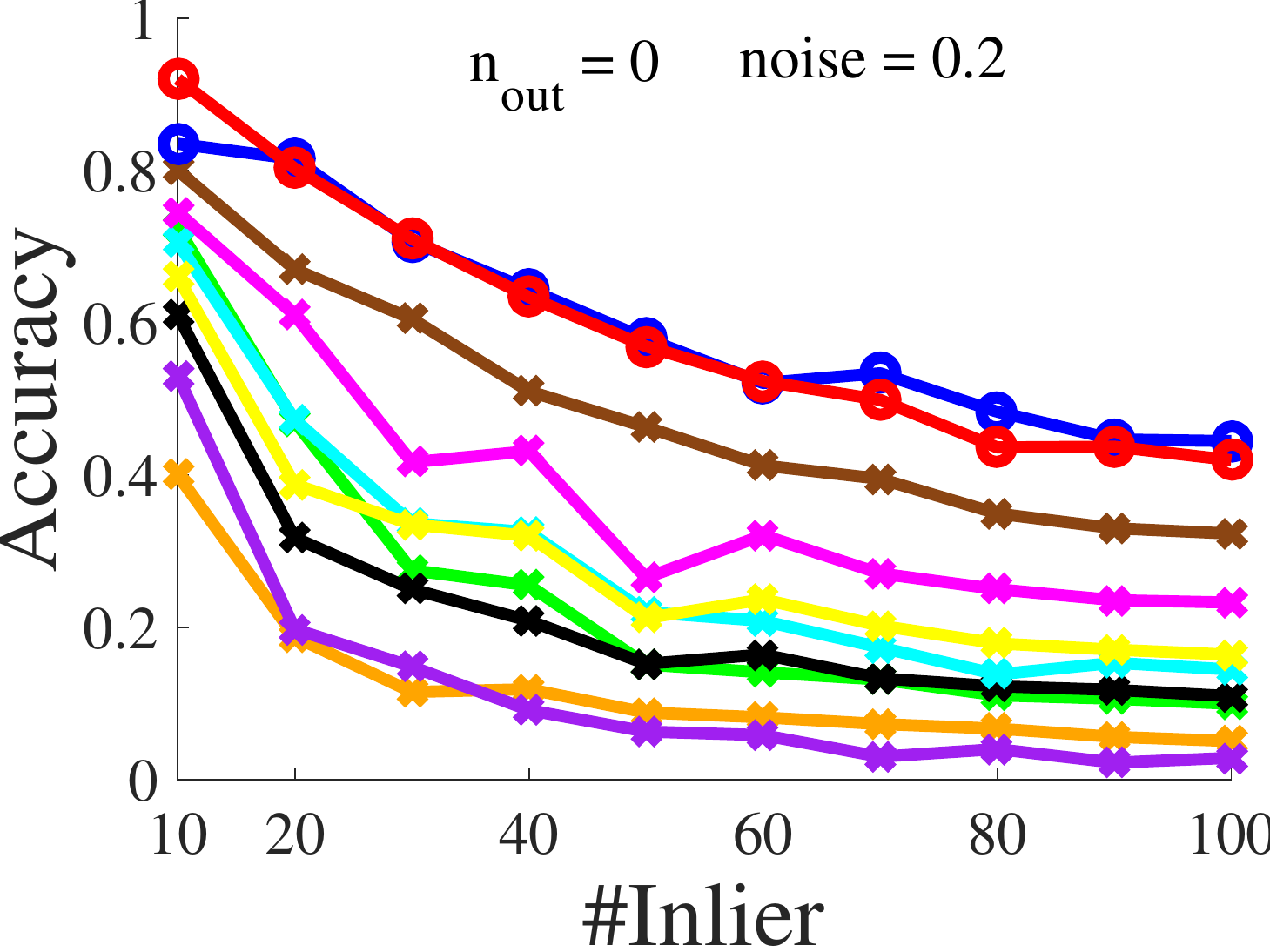}}	
		\hspace{3mm}\subfigure[]
		{\includegraphics[width=0.22\linewidth]{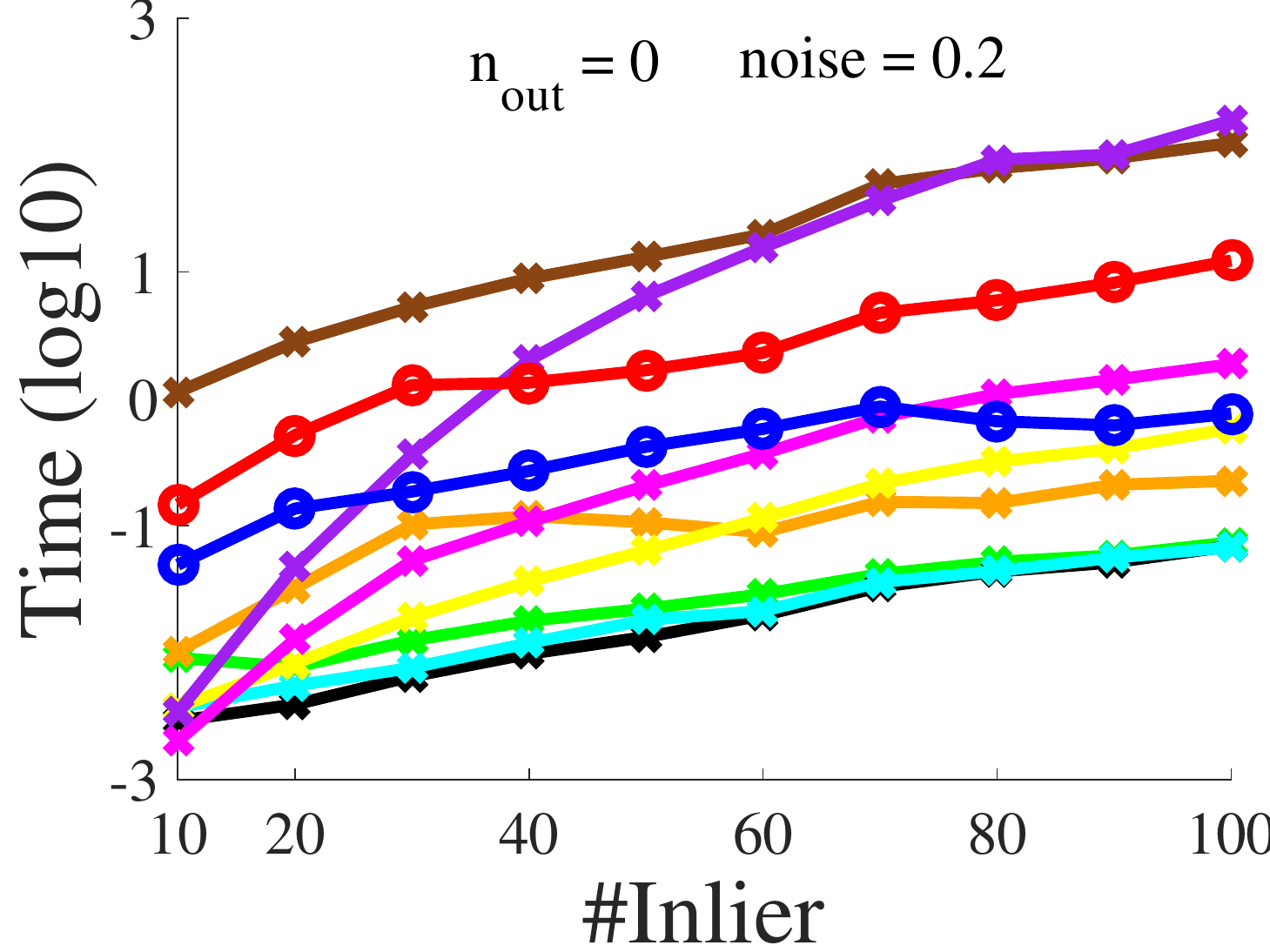}}
	\end{center}
	\vspace{-5mm}
	\caption{Comparisons of running time and average accuracy. The graphs in (a) and (b) are complete, and the graphs in (c) and (d) are connected through Delaunay triangulation. FRGM-G and FRGM-E outperform all the others in terms of matching accuracy with modest running time.}
	\label{fig_syn_time}
\end{figure*}

\subsection{Results on synthetic data}\label{section:synthetic}

In this section, we performed a comparative evaluation of both FRGM-G and FRGM-E on synthesized graphs following ~\cite{[2016-Zhou-pami],[2017-Jiang-cvpr],[2010-Cho-eccv]}. The synthetic nodes of $\mathcal{G}_1$ and $\mathcal{G}_2$ were generated as follows: for graph $\mathcal{G}_1$, $n_{in}$  inlier points were randomly generated on $\mathbb{R}^2$ with Gaussian distribution $\mathcal{N}(0,1)$. 
Graph $\mathcal{G}_2$ with noise was generated by adding Gaussian noise $\mathcal{N}(0,\sigma^2)$ to each $V^{(1)}_i$ to evaluate the robustness to noise. 
Graph $\mathcal{G}_2$ with outliers was generated by adding $n_{out}$ additional points 
on $\mathbb{R}^2$ with a Gaussian distribution $\mathcal{N}(0,1)$ to evaluate the robustness to outliers.

For the  compared methods, we computed the node affinity as $\mathbf{K}_{ij;ij}=\text{exp}({-{||\mathbf{v}^{(1)}_i-\mathbf{v}^{(2)}_j||}})$ with $\mathbf{v}^{(1)}_i, \mathbf{v}^{(2)}_j$ computed by shape context~\cite{[2002-Belongie-pami]}, and we computed the edge affinity as $$\mathbf{K}_{i_1j_1;i_2j_2}=\text{exp}(-{(||V^{(1)}_{i_1}-V^{(1)}_{i_2}||-||V^{(2)}_{j_1}-V^{(2)}_{j_2}||)^2}/{0.15}),$$ 
as used in~\cite{[2016-Zhou-pami]}. For FRGM-G, we computed the unary term $\mathbf{U}$ by shape context and set $\alpha_1=0.99, \alpha_2=0.5$. For FRGM-E, we set $\lambda_1=0.99, \lambda_2=0.5$. To compute the matrix $\mathbf{S}\in \mathbb{R}^{m\times m}$ that indicated the adjacent nodes in $\{\mathcal{T}(V^{(1)}_i)\}_{i=1}^m$ , we performed a Delaunay triangulation on $\mathcal{V}_1$ to connect the edges. Then these edges were divided into two parts using k-means by considering the edge length. Edges with longer lengths were then abandoned.

\vspace{1mm}
{\bf Average accuracy.} Since the compared methods have a considerable space complexity of $\mathbf{O}(m^2n^2)$ when the graphs are fully connected, they can hardly handle complete graphs with more than a hundred nodes. Therefore, for fairness, we performed the experiment on two types of graphs: smaller graphs that were fully connected and lager graphs that were connected by Delaunay triangulation. We first applied all methods to complete graphs with a small size $n_{in}=20$ with either the noise level $\sigma$ varying from $0$ to $0.5$ (by intervals of 0.05) or number of outliers $n_{out}$ varying from $0$ to $20$ (by intervals of 2). Then, we enlarged the size of the graphs to $n_{in}=100$ and connected the edges by Delaunay triangulation. Similarly, noise and outliers were added.

As shown in Fig.~\ref{fig:syn_menory} (a) and (b), under the complete graph setting, our algorithms FRGM-G and FRGM-E achieve higher average accuracy than the other algorithms in the case with noise and achieve competitive results in the case with outliers, respectively. 
As shown in Fig.~\ref{fig:syn_menory} (c) and (d), with larger graphs connected by Delaunay triangulation, both FRGM-G and FRGM-E outperform all the other methods. Moreover, we can observe that all algorithms achieve higher accuracy on complete graphs than graphs connected by Delaunay triangulation.

\vspace{1mm}
{\bf Running time.} To compare the time consumptions of all methods, we tested all methods on graphs with inliers varying as $n_{in}=10,20,...,100$ and noise $\sigma = 0.2$. Considering the effect of the number of edges on time consumption, we used both complete and Delaunay-triangulation-connected graphs. 

As shown in Fig.~\ref{fig_syn_time}, under the same conditions in which graphs are either complete or connected through Delaunay triangulation, our algorithms FRGM-G and FRGM-E achieve higher average accuracy within an intermediate running time. For all methods, matching complete graphs comsumes more time than Delaunay-triangulation-connected graps. Compared with GA, SM, PM, SMAC, and IPFP-S, which run faster, our method can achieve higher average accuracy. 
The methods RRWM, FGM and MPM can achieve competitive accuracy when the graphs are fully connected. However, the time consumptions of these methods rapidly increase and become larger than that of ours method, and they will take an unacceptable amount time to match complete graphs that have more than a hundred nodes. 

\vspace{1mm}
{\bf Large-scale graph matching.}
To test the efficiency of our algorithms FRGM-G and FRGM-E on large-scale graphs, we used more challenging settings for evaluation. 
We set the number of inliers as $n_{in}=100,300,500,1000$ with Gaussian noise or outliers. The number of outliers was set to $20\%,40\%,...,100\%$ of the number of inliers.

Tab.~\ref{table_largescale1} reports the results of FRGM-G and FRGM-E on complete graphs with hundreds and thousands of nodes. Both FRGM-G and FRGM-E are very robust to outliers and less robust to strong noise with larger graphs. FRGM-E is much faster than FRGM-G because FRGM-E searches for the optimal solution upon a simpler Euclidean space $\mathbb{R}^d$, while FRGM-G searches for the optimal solution upon a more complex function space $\mathcal{F}(\mathcal{V}_2,\mathbb{R})$. Since the compared methods need to store affinity matrices with a size of approximately $n_{in}^2(n_{in}+n_{out})^2$, applying these methods to large-scale graphs with hundreds or thousands of nodes is infeasible. 

\begin{table*}[htb!]
	\centering
\caption{Average accuracy and running time of FRGM-G (left) v.s. FRGM-E (right) on synthetic data with varying inliers, noise and outliers.}
\vspace{2mm}
	\begin{minipage}{0.95\linewidth}
	\centering
		\footnotesize
		\begin{tabular}{cc|ccccc}				
		\toprule[1.0pt]	 
		\multicolumn{1}{c}{\#Inlier}&{Noise ($\sigma$)}&0.02 & 0.04  & 0.06 & 0.08 & 0.10   \\
		\hline
		\multirow{2}{20pt}{100}
		&time (s) &{\bf 0.07} / { 0.11} &{\bf 0.11} / 0.32 &{\bf 0.14} / { 0.54} &{\bf 0.24} / {0.73}&{\bf 0.25} / { 0.88} \\
		&acc. (\%) &98.38 / {\bf 98.86} &94.28 / {\bf 95.70} &88.74 / {\bf 90.70} &81.10 / {\bf 85.20} &72.16 / {\bf 75.76} \\
		\hline		
		\multirow{2}{20pt}{300}
		&time (s)  &{\bf 2.07} / 4.43  &{\bf 3.78} / { 6.80}  &7.88 / {\bf 7.06}  &15.20 / {\bf 7.29}  &19.18 / {\bf 7.34} \\
		&acc. (\%) &95.33 / {\bf 96.34} &85.08 / {\bf 87.66} &71.32 / {\bf 76.11} &60.17 / {\bf 63.73} &48.60 / {\bf 50.79} \\
		\hline			
		\multirow{2}{20pt}{500}
		&time (s) &{\bf 18.10} / 27.18 &34.43 / {\bf 27.83} &86.30 / {\bf 28.94} &161.14 / {\bf 29.61} &214.29 / {\bf 30.33} \\
		&acc. (\%)&93.14 / {\bf 94.35} &77.62 / {\bf 80.27} &60.32 / {\bf 63.80} &46.74 / {\bf 49.70} &37.44 / {\bf 39.14} \\
		\hline 
		\multirow{2}{20pt}{1000}
		&time (s) &{\bf 120.85} / 124.66&233.14 / {\bf 179.87}&426.40 / {\bf 184.53}&711.78 / {\bf 187.67}&1810.42 / {\bf 191.76} \\
		&acc. (\%)&88.16 / {\bf 89.43} &63.29 / {\bf 66.34} &43.79 / {\bf 45.23} &32.20 / {\bf 33.47} &24.23 / {\bf 25.27} \\
		\bottomrule[1.0pt]	 			
	\end{tabular}	
	\setlength{\tabcolsep}{2.93mm}			
	\begin{tabular}{cc|ccccc}				
		\toprule[1.0pt]	 
		\multicolumn{1}{c}{\#Inlier}&{\#Outlier} &0.2 &0.4  & 0.6 & 0.8 & 1.0   \\
		\hline	
		\multirow{2}{20pt}{100}
		&time (s) &0.06 / {\bf 0.03}  &0.08 / {\bf 0.04}  &0.10 / {\bf 0.07}  &0.15 / {\bf 0.14}  &0.22 / {\bf 0.17}\\
		&acc. (\%)&99.86 / {\bf 99.98} &99.74 / {\bf 99.88} &99.22 / {\bf 99.85} &98.77 / {\bf 99.76} &98.08 / {\bf 99.68} \\
		\hline				
		\multirow{2}{20pt}{300}
		&time (s) &1.26 / {\bf 0.18}&2.04 / {\bf 0.34} &5.53 / {\bf 0.74} &7.87 / {\bf 1.17} &10.33 / {\bf 2.02}\\
		&acc. (\%)&99.90 / {\bf 99.98} &99.66 / {\bf 99.92} &99.33 / {\bf 99.93} &98.66 / {\bf 99.83} &98.19 / {\bf 99.83} \\
		\hline
		\multirow{2}{20pt}{500}
		&time (s) &27.72 / {\bf 0.97}  &48.34 / {\bf 2.01}  &74.65 / {\bf 3.43}  &136.82 / {\bf 5.89}  &222.28 / {\bf 10.91}\\
		&acc. (\%)&99.94 / {\bf 99.95} &99.86 / {\bf 99.95} &99.44 / {\bf 99.89} &98.18 / {\bf 99.87} &97.54 / {\bf 99.96}\\
		\hline
		\multirow{2}{20pt}{1000}
		&time (s) &181.63 / {\bf 4.48} &337.41 / {\bf 10.99} &461.22 / {\bf 30.94} &502.51 / {\bf 54.95} &626.03 / {\bf 77.72}\\
		&acc. (\%)&99.92 / {\bf 99.99} &99.68 / {\bf 99.97} &99.16 / {\bf 99.97} &97.95 / {\bf 99.97} &97.05 / {\bf 99.96} \\
		\bottomrule[1.0pt]	 
	\end{tabular}	
	\end{minipage}
	\label{table_largescale1}
\end{table*}

\subsection{Results on real-world image datasets}\label{section:real}
We also performed comparative evaluations on real-world datasets, including the CMU House and Hotel sequences\footnote{\url{http://vasc.ri.cmu.edu//idb/html/motion/house/index.html}}, the PASCAL Cars and Motorbikes pairs~\cite{[2012-Leordeanu-ijcv]}, which are commonly used to evaluate GM algorithms. All methods are applied to complete graphs.

The CMU House and Hotel sequences consist of 111 and 101 frames of a synthetic house and hotel, respectively. Each image contains 30 points that are manually marked with known correspondence. In this experiment, we matched all the image pairs separated by 10, 20,.., 90 frames. The unequal-sized cases are set as 20-vs-30 and 25-vs-30. For the compared methods, we computed the node affinity $\mathbf{K}_{ij;ij}=\text{exp}({-{||\mathbf{v}^{(1)}_i-\mathbf{v}^{(2)}_j||}})$ with shape context, and we computed the edge affinity as $\mathbf{K}_{i_1j_1;i_2j_2}=\text{exp}(-{(||V^{(1)}_{i_1}-V^{(1)}_{i_2}||-||V^{(2)}_{j_1}-V^{(2)}_{j_2}||)^2}/{2500})$ as used in ~\cite{[2016-Zhou-pami]}.

\begin{figure*}[t!]
\centering
		\includegraphics[width=0.7\linewidth]{./Images/cmu_legend.pdf}\\
		{\includegraphics[width=1\linewidth]{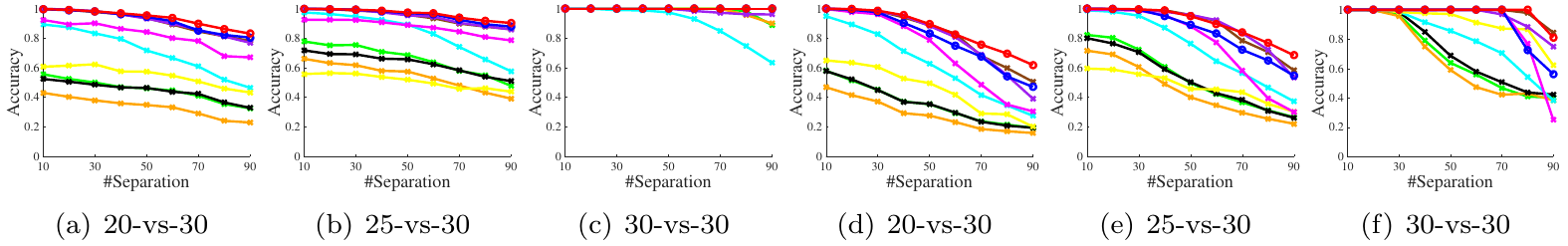}}
    \vspace{-8mm}
	\caption{Comparison of average accuracy on the {\em house} (a)-(c) and {\em hotel} (d)-(f) sequences in both equal-sized and unequal-sized cases. FRGM-G achieves higher average accuracies than the other methods.}
	\label{fig:CMU-house}
\end{figure*}

\vspace{1mm}
{\bf Average accuracy.} 
As shown in Fig.~\ref{fig:CMU-house}, for the house sequence, our algorithms FRGM-G and FRGM-E achieve higher accuracy in both equal-sized and unequal-sized cases. For the hotel sequence, FRGM-G outperforms all the other methods. 

\begin{figure*}[t!]
	\centering
	\subfigure[20-vs-30 (Acc:~20/20)] 
    {\includegraphics[height=0.1\linewidth,width = 0.24\linewidth]{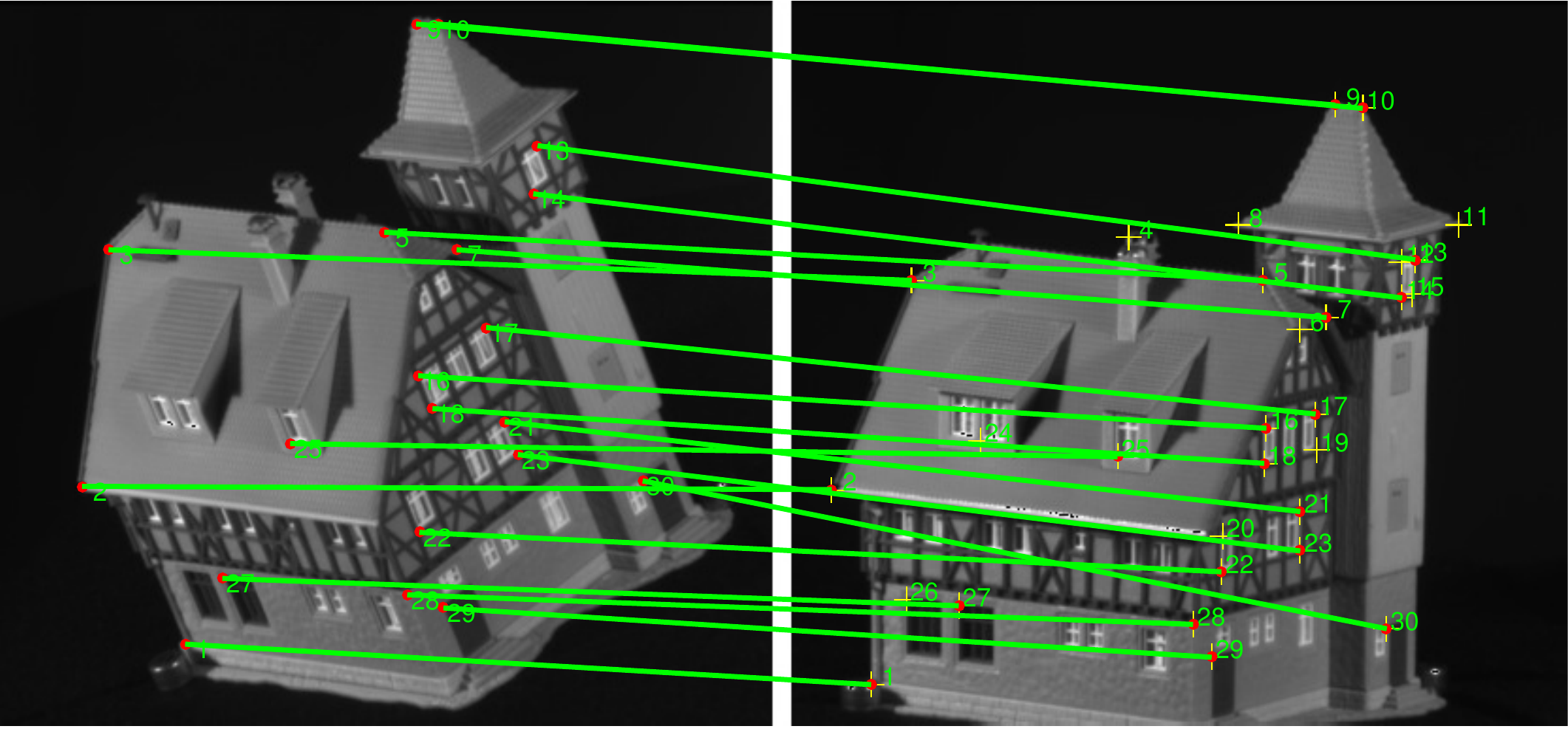}}
    \subfigure[20-vs-30 (Acc:~20/20)] 
	{\includegraphics[height=0.1\linewidth,width = 0.24\linewidth]{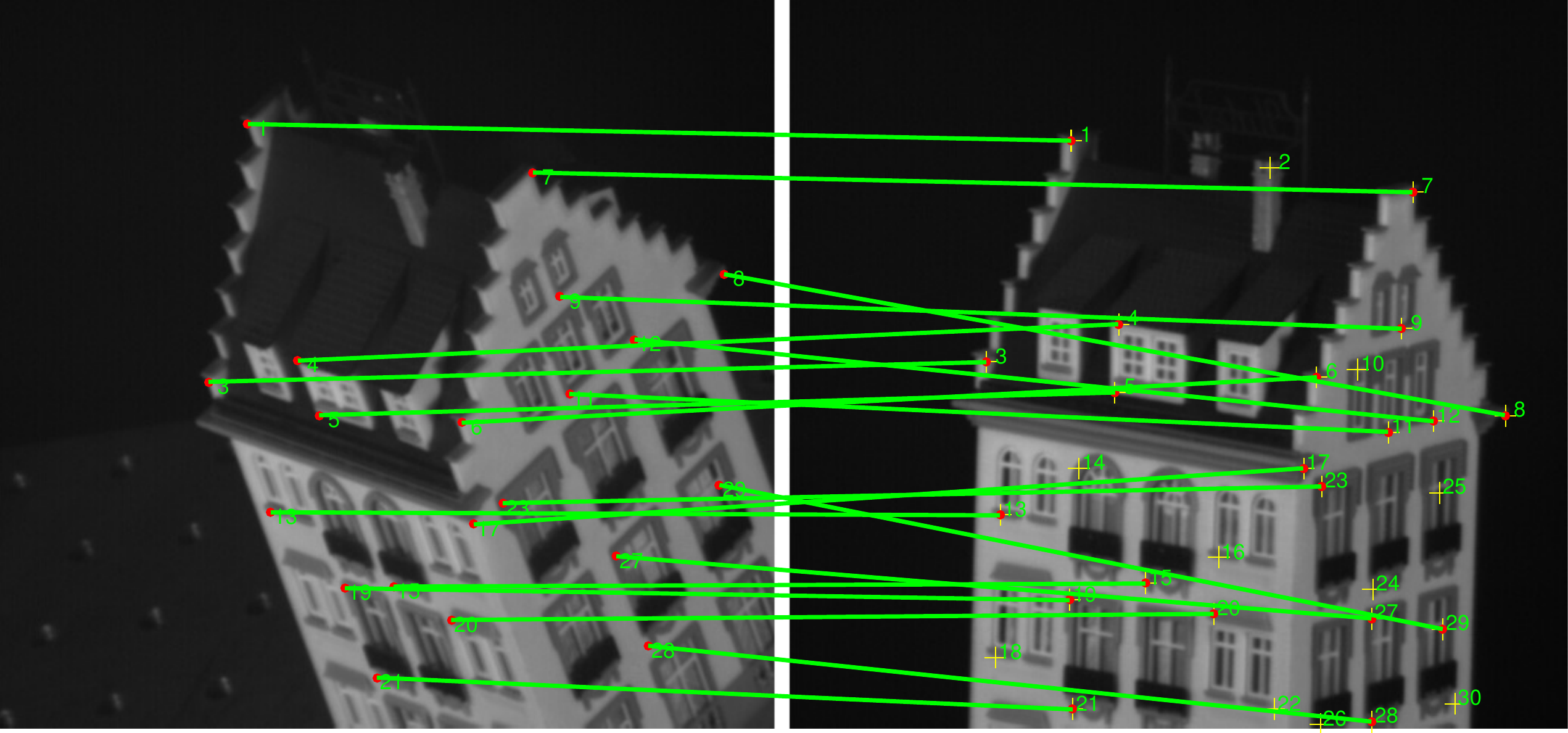}} 
	\hspace{1mm}\subfigure[28-vs-48 (Acc:~28/28)]  	    
	{\includegraphics[height=0.1\linewidth,width = 0.24\linewidth]{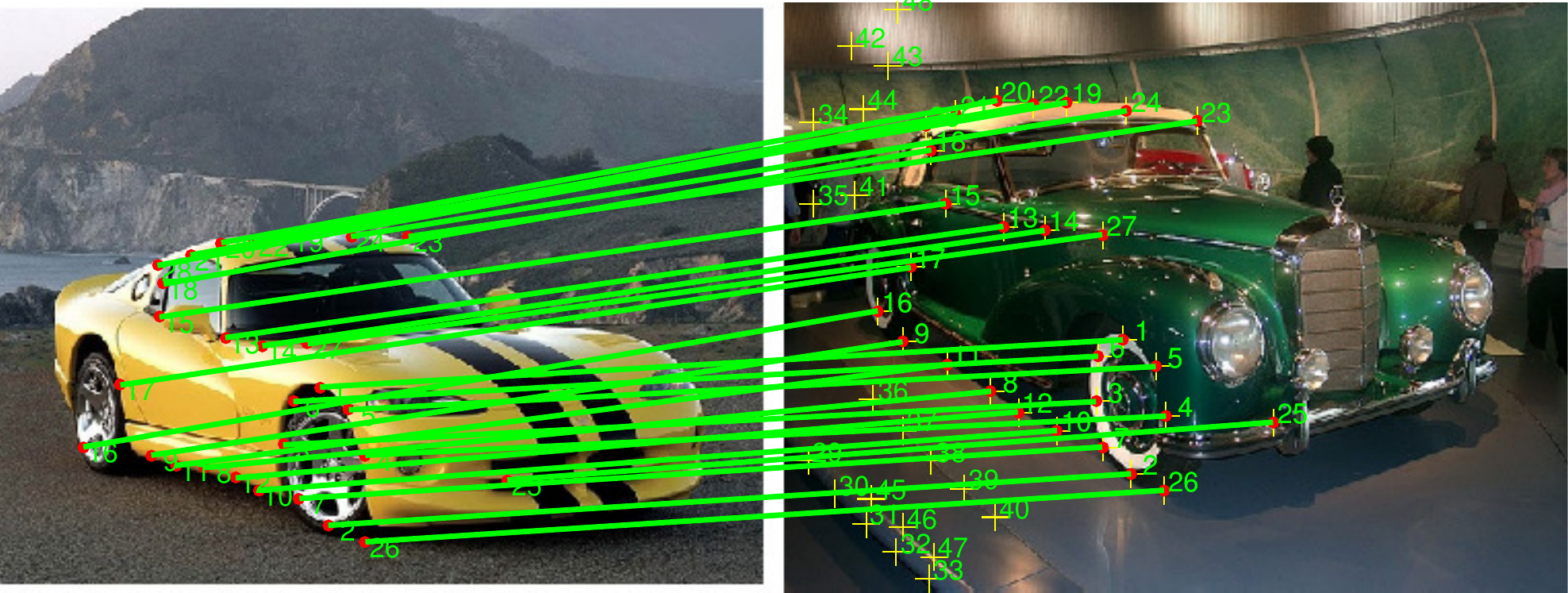}}
    \subfigure[46-vs-86 (Acc:~46/46)]  	    
    {\includegraphics[height=0.1\linewidth,width = 0.24\linewidth]{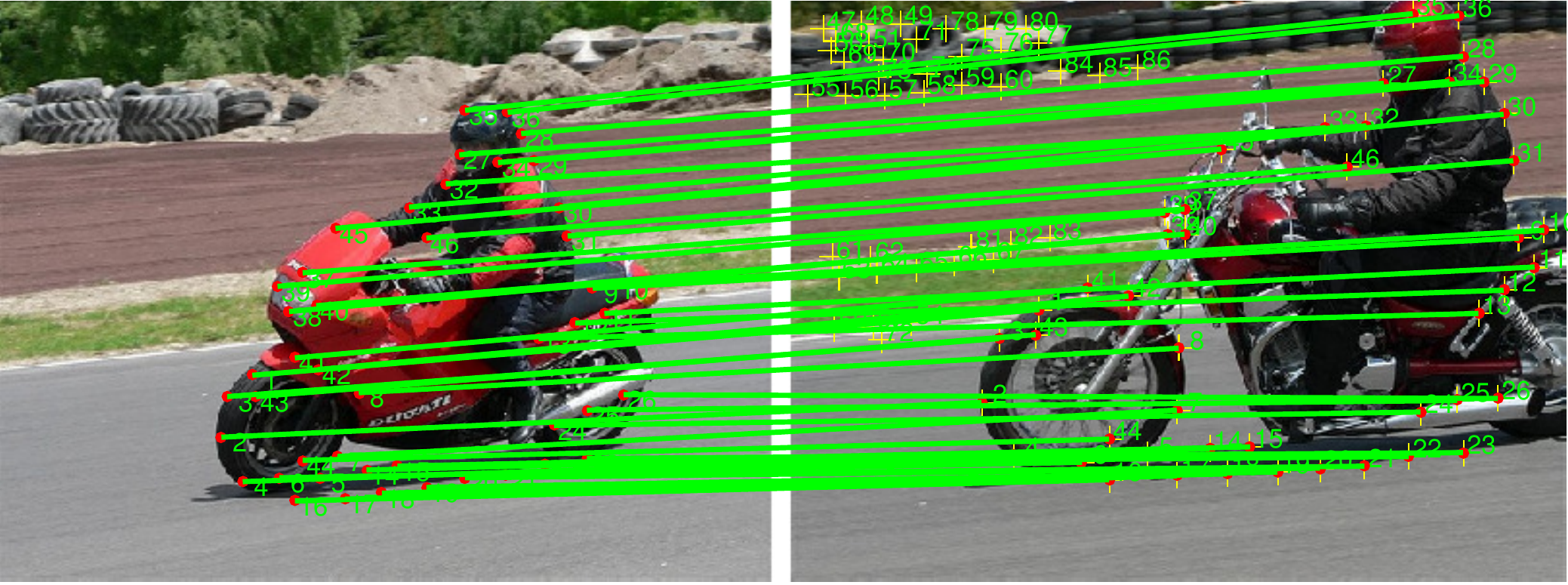}}
	\vspace{-2mm}
	\caption{Examples of matching unequal-sized graphs using FFRGM-G (in (a) and (b)) and FRGM-E (in (c) and (d)). The red dots are inliers in $\mathcal{G}_1$, and the yellow plus signs are inliers with outliers in $\mathcal{G}_2$. The lines in green are correct matches.}
\end{figure*}
\begin{figure}[htb!]
	\centering
    {\includegraphics[width=0.78\linewidth]{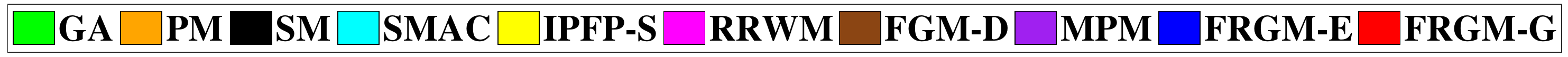}}\\	
	\subfigure[Cars]
	{\includegraphics[width=0.42\linewidth]{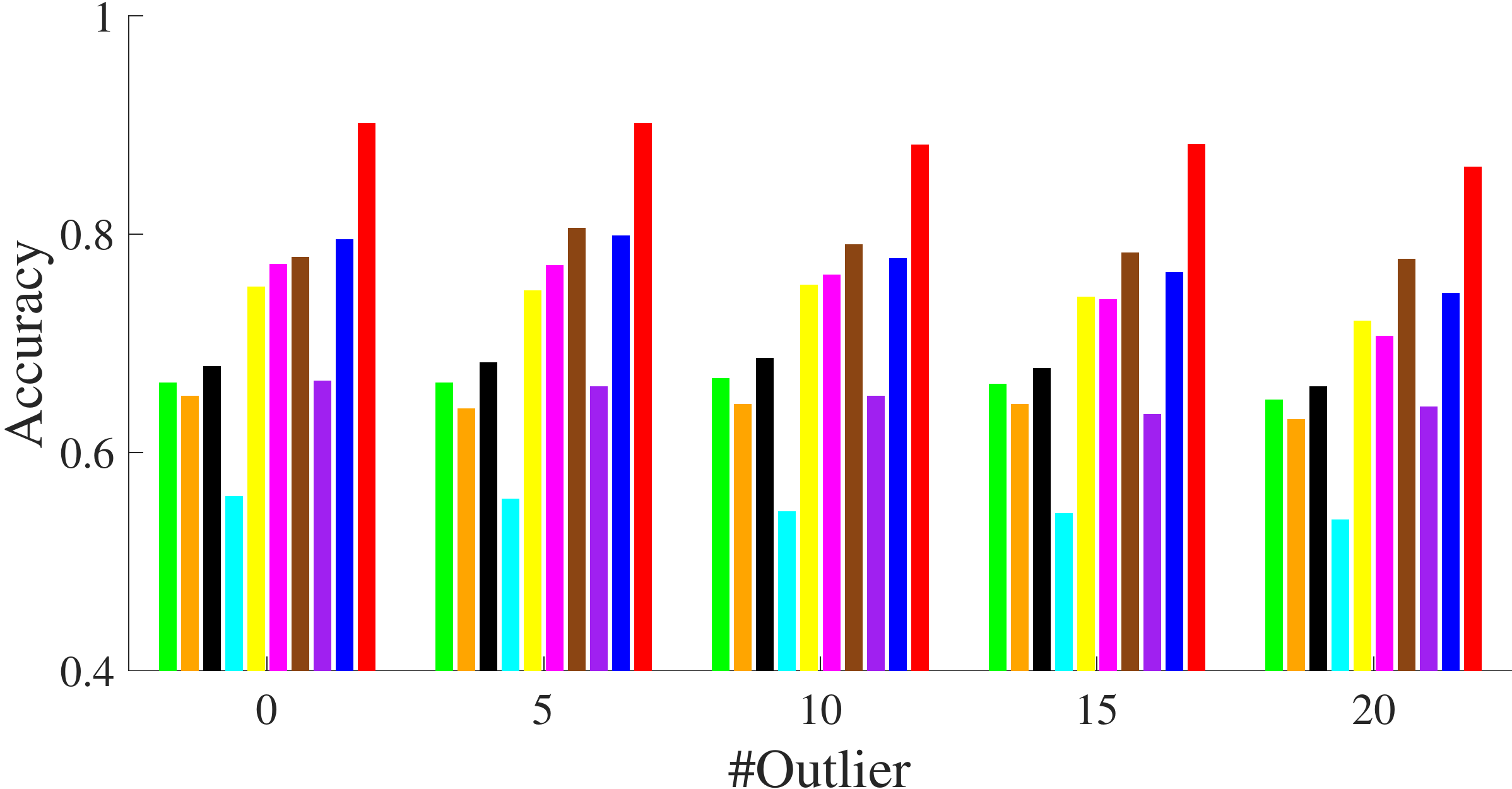}}
	\subfigure[Motorbikes]
	{\includegraphics[width=0.42\linewidth]{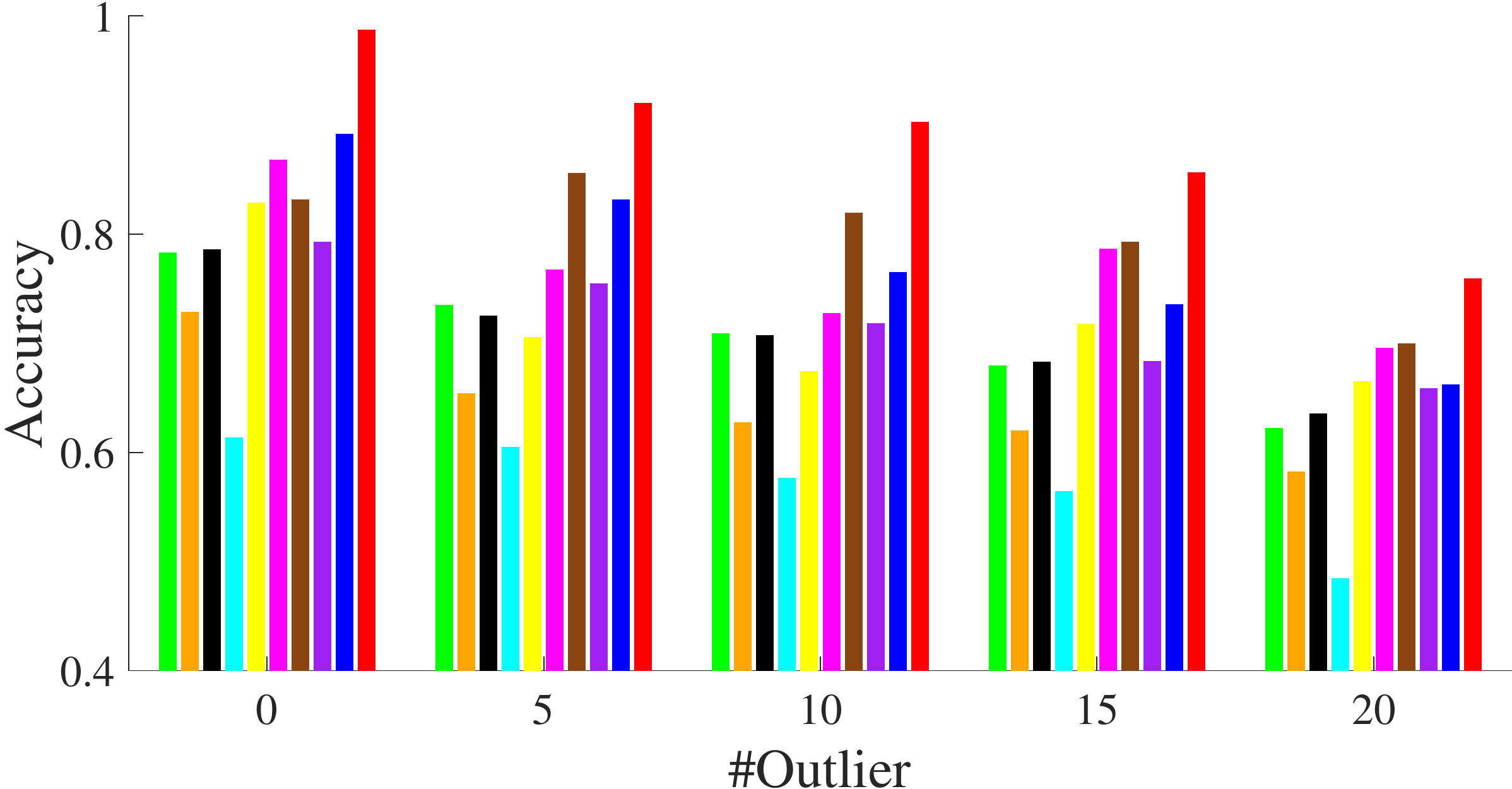}}
	\vspace{-3mm}
	\caption{Comparison on car and motorbike image pairs with outliers. FRGM-G achieves the highest average accuracy.}
	\label{fig:car-motor}
\end{figure}

The PASCAL dataset consists of 30 pairs of car images and 20 pairs of motorbike images. Each pair contains both inliers with known correspondence and randomly selected outliers. In the unequal-sized case, we added 5, 10, 15, and 20 outliers to $\mathcal{G}_2$. For the compared methods, we computed the same node affinity used on the CMU dataset, and we computed the edge affinity with the edge length and edge angle, which were used in ~\cite{[2016-Zhou-pami]}. 

\vspace{1mm}{\bf Outlier removal effectiveness.} The outliers occurring in the PASCAL dataset often seriously affect the matching accuracy. Therefore, we tested our proposed outlier-removal strategy on this dataset. We first applied it to graph pairs with outliers as a preprocessing step, and then we executed all the algorithms on the preprocessed graphs. As shown in Tab.~\ref{table_re}, the average accuracy of all the methods is greatly improvement, and almost all the methods improve their performance by more than $10\%$. Moreover, as shown in Fig.~\ref{fig:car-motor}, FRGM-G achieves the highest average accuracy, and FRGM-E achieves a competitive result. 
\begin{table}[htb!]
	\centering
	\small
	\caption{Effectiveness of outlier-removal strategy. It improves the average matching accuracy by more than $10\%$ for almost all the methods.}
	\vspace{1mm}
	\begin{tabular}{cc|cc}
		\toprule[1.0pt]
		Method & Out. Re.& Cars & Motorbikes\\
		\hline
		\multirow{2}{40pt}{GA ~\cite{[1996-Gold]}}
		& w/o  & 34.50 & 45.97\\
		& w/  & \textbf{66.14}& \textbf{70.60}\\
		\hline
		\multirow{2}{40pt}{PM ~\cite{[2008-Zass-cvpr]}}
		& w/o  & 37.04 & 43.56\\
		& w/  &\textbf{ 64.21}& \textbf{64.26}\\
		\hline	
		\multirow{2}{40pt}{SM ~\cite{[2005-Leordeanu]}}
		& w/o  & 38.04 & 47.13\\
		& w/  &\textbf{ 67.72}& \textbf{70.75}\\
		\hline	
		\multirow{2}{40pt}{SMAC ~\cite{[2006-Cour-nips]}}
		& w/o  & 38.53 & 43.84\\
		& w/  & \textbf{54.91}& \textbf{56.89}\\
		\hline
		\multirow{2}{40pt}{IPFP-S ~\cite{[2009-Leordeanu-nips]}}
		& w/o  & 38.53 & 43.84\\
		& w/  & \textbf{74.36}& \textbf{71.84}\\
		\hline
		\multirow{2}{40pt}{RRWM ~\cite{[2010-Cho-eccv]}}
		& w/o  & 53.84 & 65.64\\
		& w/  & \textbf{75.09}& \textbf{76.92}\\
		\hline
		\multirow{2}{40pt}{FGM-D ~\cite{[2016-Zhou-pami]}}
		& w/o  & 49.05 & 67.31\\
		& w/  & \textbf{78.72}& \textbf{80.01}\\
		\hline
		\multirow{2}{40pt}{MPM \cite{[2014-Cho-cvpr]}}
		& w/o  & 58.02 & 65.73\\
		& w/  &\textbf{ 65.11}& \textbf{72.17}\\
		\hline
		\multirow{2}{40pt}{FRGM-E}
		& w/o  & 32.74 & 43.19\\
		& w/  & \textbf{77.67}& \textbf{77.74}\\
		\hline
		\multirow{2}{40pt}{FRGM-G}
		& w/o  & 55.02 & 68.19\\
		& w/  & \textbf{88.59}& \textbf{88.51}\\
		\bottomrule[1.0pt]
	\end{tabular}
	\label{table_re}
\end{table}

\begin{figure*}[htb!]
	\centering
	\subfigure
	{\includegraphics[width=0.98\linewidth]{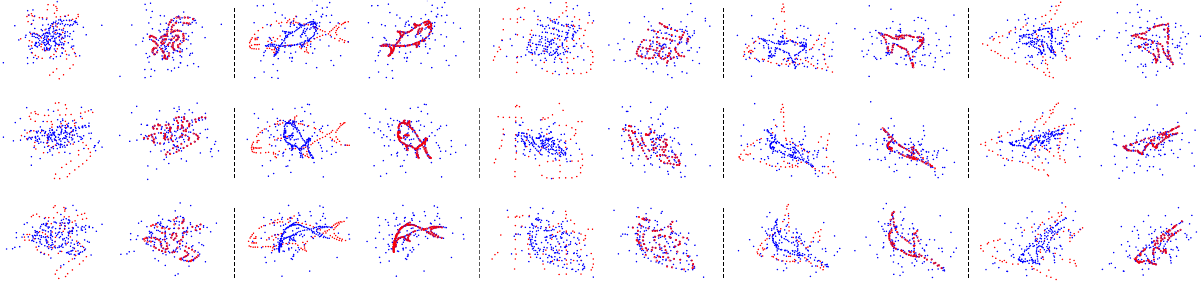}}
	\vspace{-4mm}
	\caption{Examples of matching graphs with geometric deformations by our algorithm FRGM-D. From left to right:  the Olympic logo, whale, Chinese character, tropical fish and UCF fish. From top to bottom: the graphs $\mathcal{G}_1$ (red dots) are deformed with similarity (the first row), affine (the second row) and nonrigid (the third row) transformations. Graphs $\mathcal{G}_2$ are disturbed by geometric deformations, noises, missing points and outliers.}
	\label{fig:5shape_example_all}
\end{figure*}

\begin{figure*}[htb!]
	\centering
	\subfigure[]
	{\includegraphics[width=0.32\linewidth]{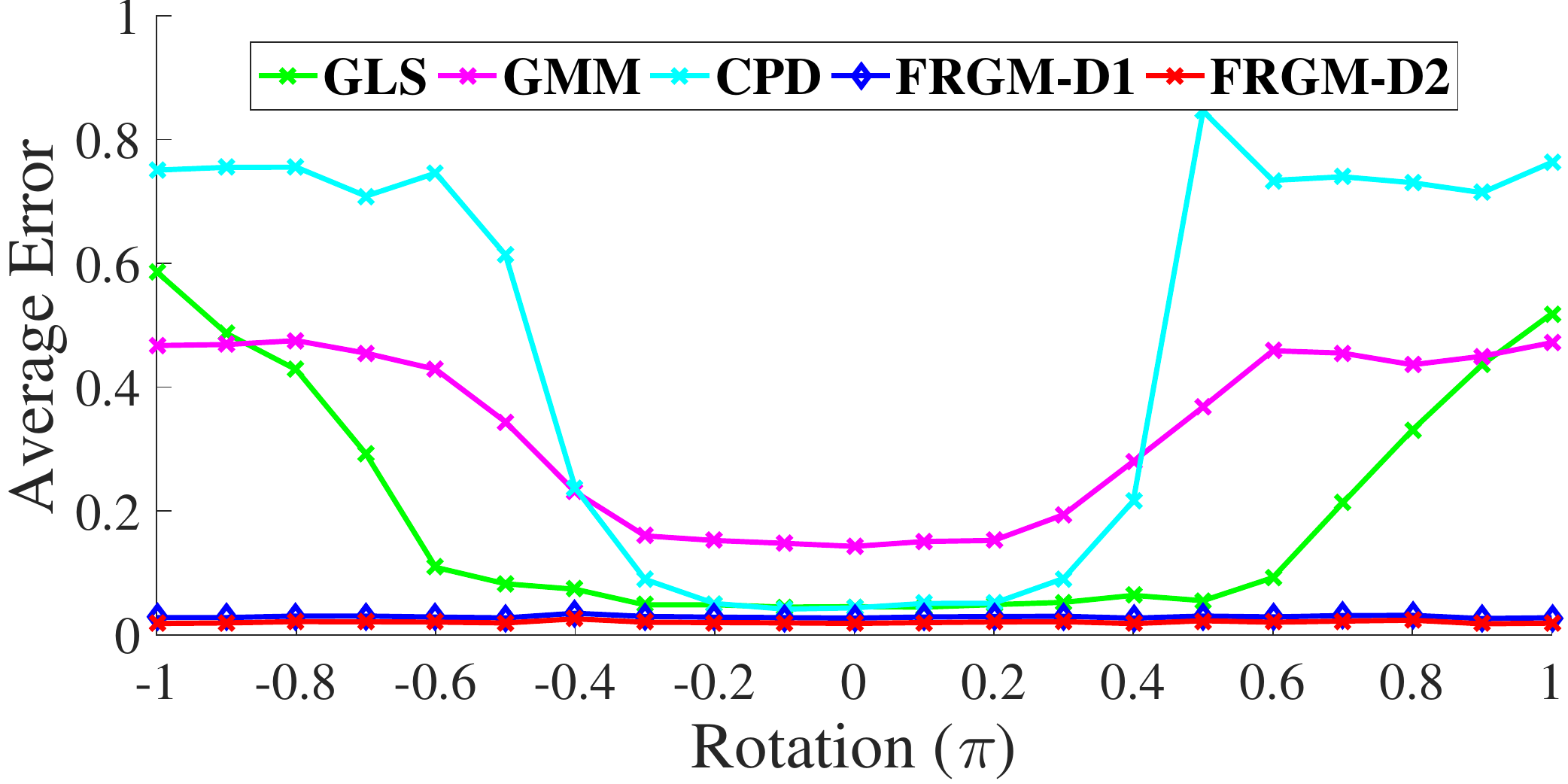}}
	\hspace{2mm}\subfigure[]
	{\includegraphics[width=0.32\linewidth]{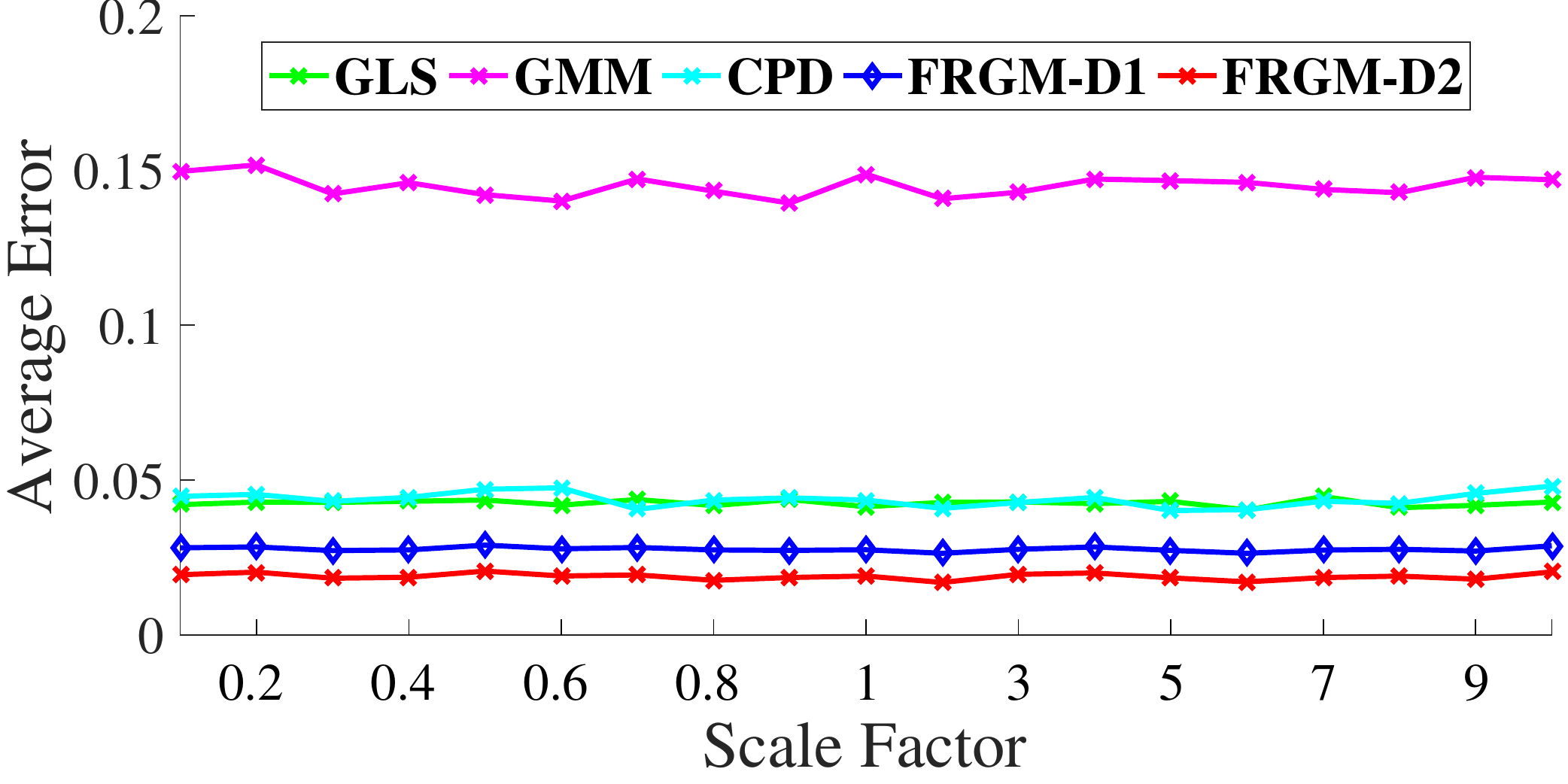}}
	\hspace{2mm}\subfigure[]
	{\includegraphics[width=0.32\linewidth]{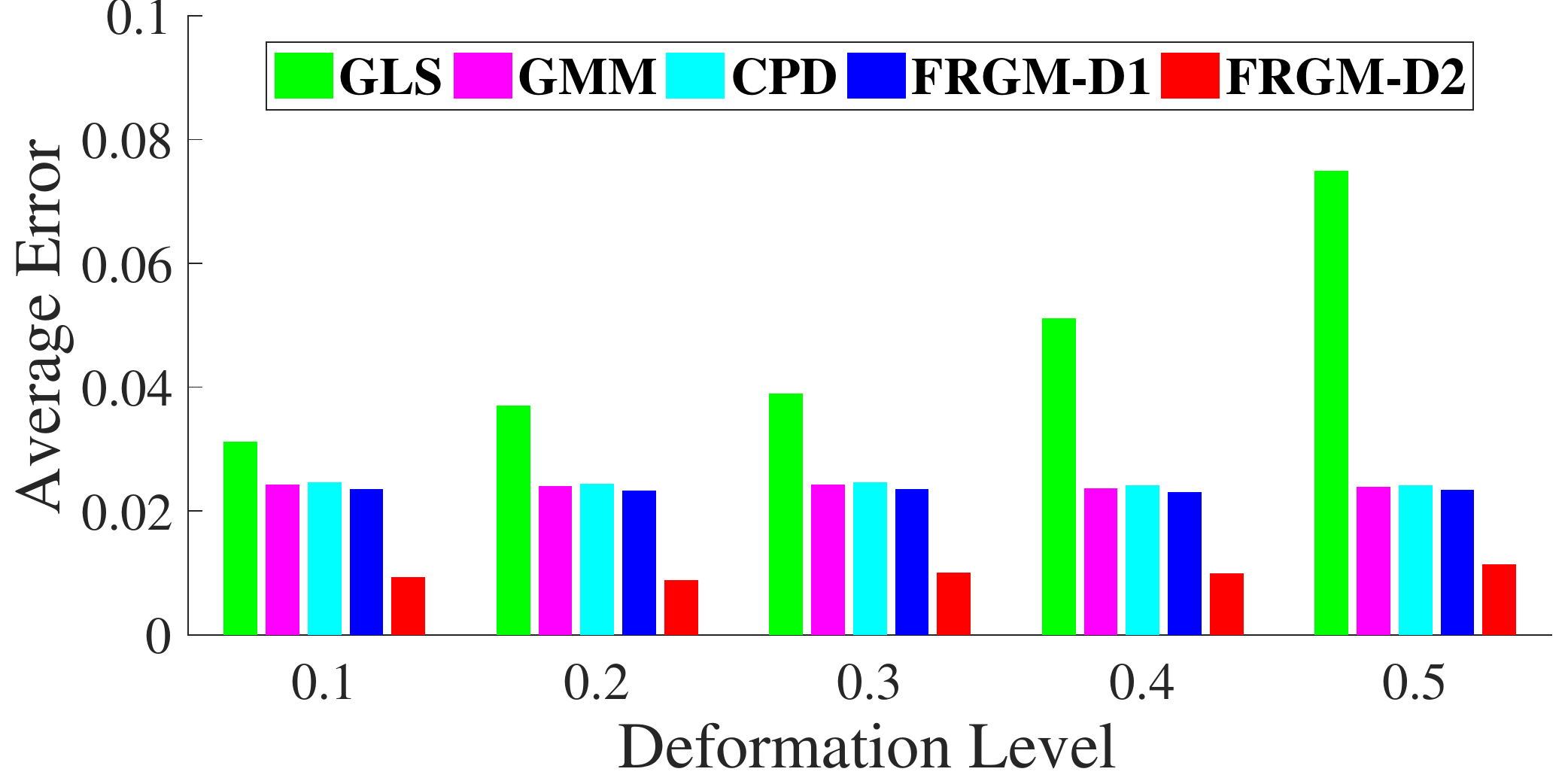}}
	\vspace{-4mm}
	\caption{Comparisons on rotation (a), scaling factor (b) and deformation level (c). Both FRGM-D1 and FRGM-D2 have less average errors than the other algorithms.}
	\label{fig:5shape_rotation}
\end{figure*}
\begin{figure}[t!]
	\centering
	{\includegraphics[width = 0.6\linewidth,height = 0.03\linewidth]{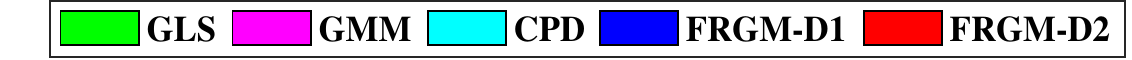}}	
	{\includegraphics[width = 0.5\linewidth]{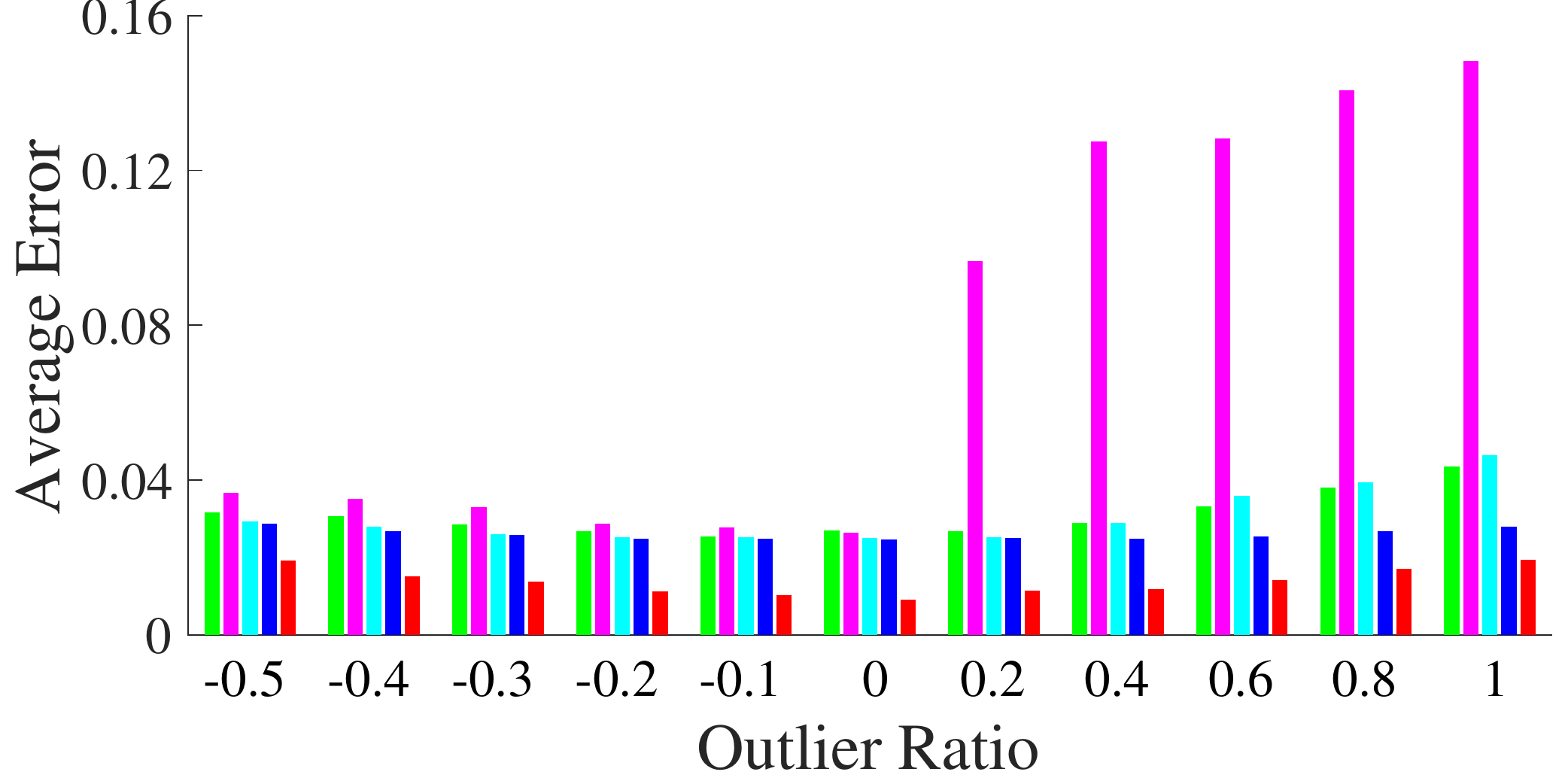}}
	{\includegraphics[width = 0.27\linewidth]{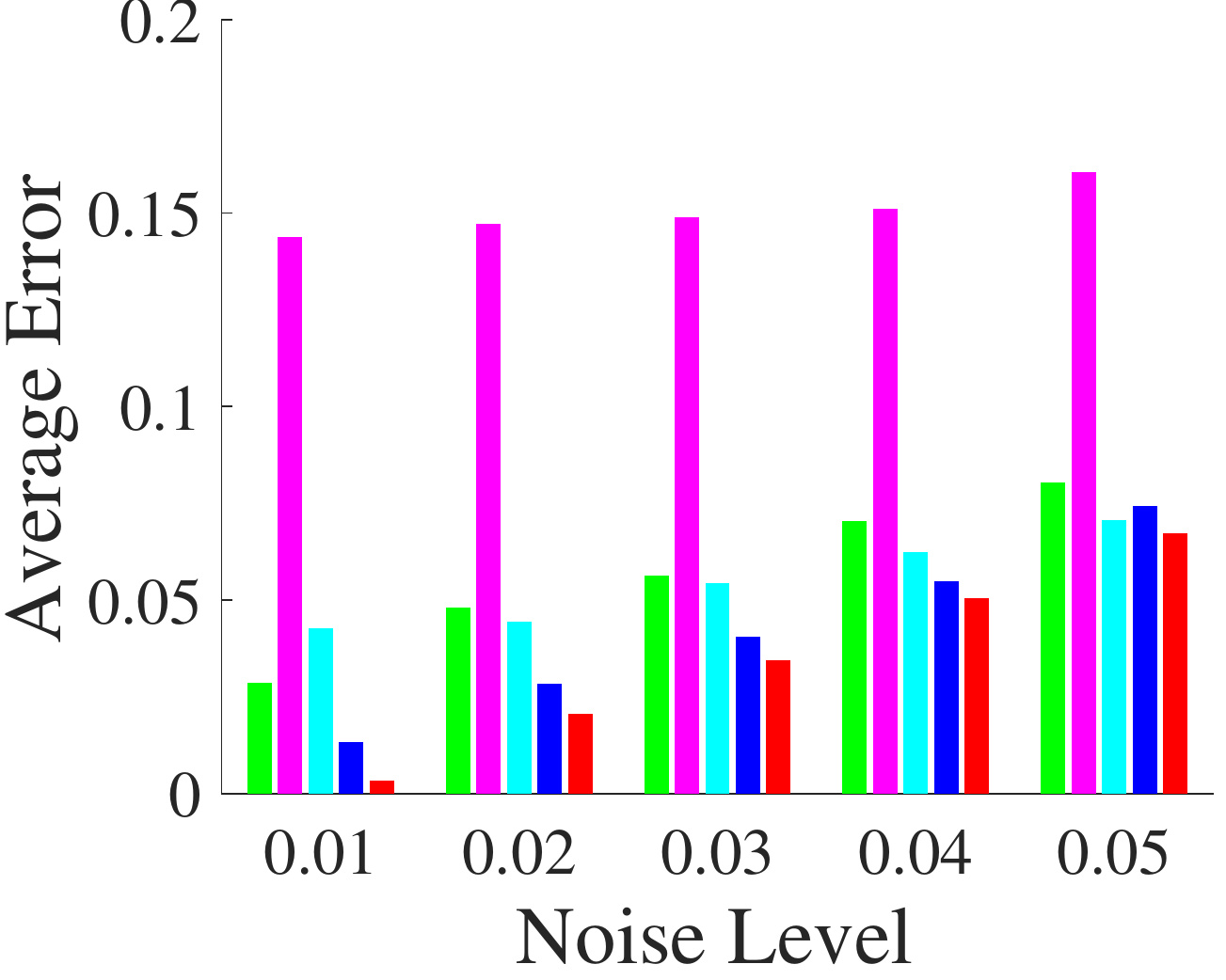}}
	{\includegraphics[width = 0.5\linewidth]{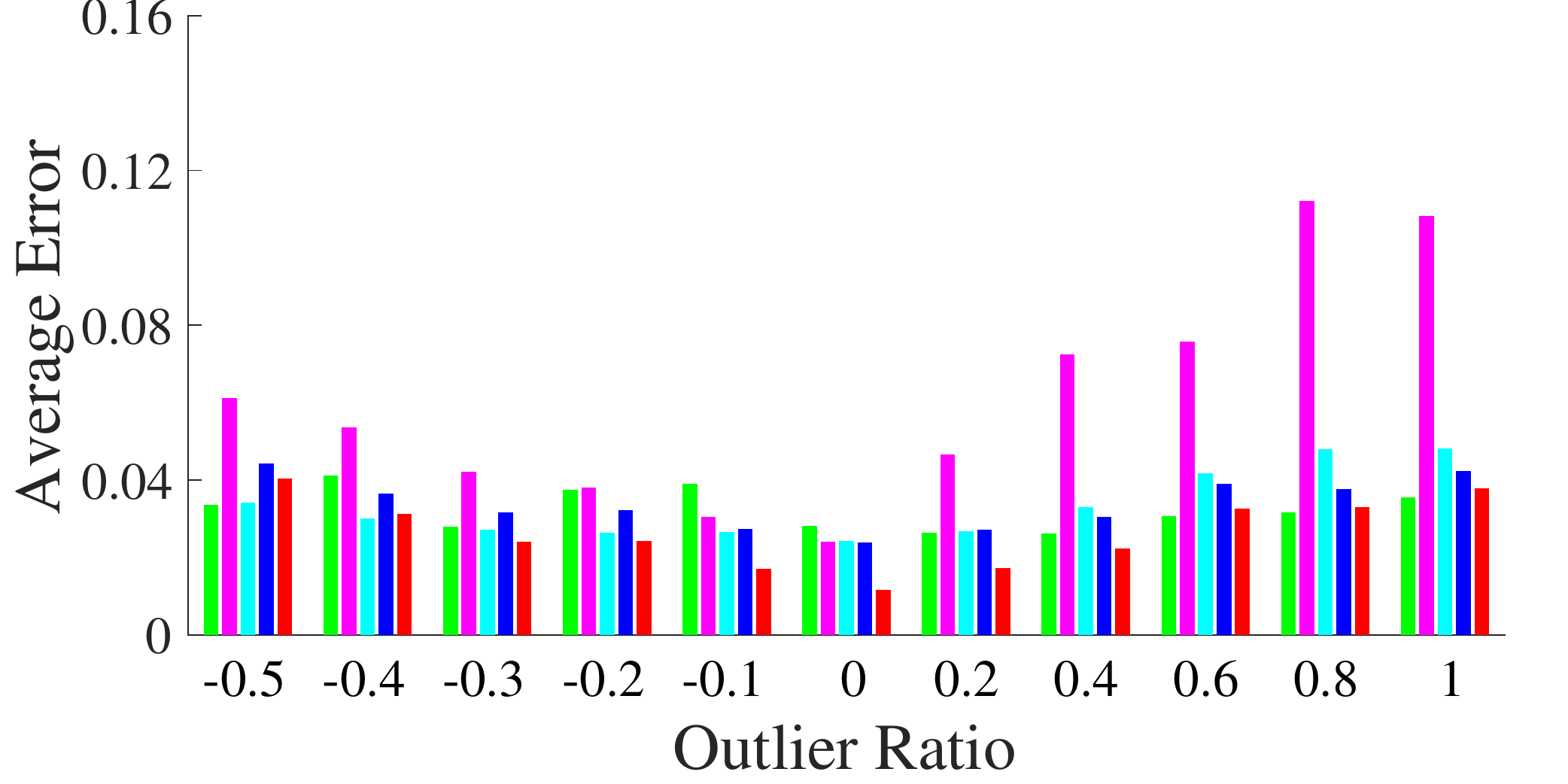}}
	{\includegraphics[width = 0.27\linewidth]{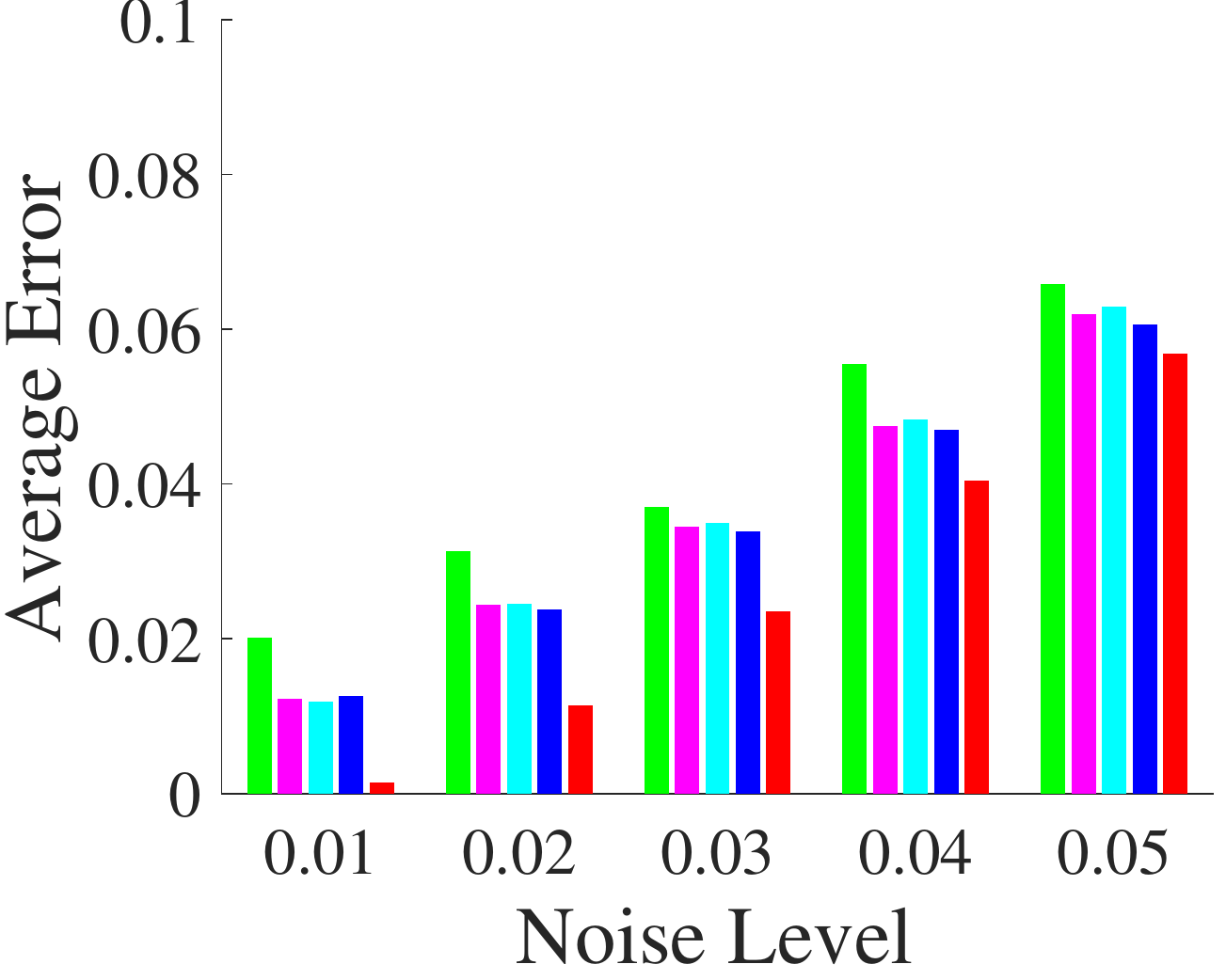}}
	\vspace{-3mm}
	\caption{Comparisons on outliers and noises for similarity deformation (the first row) and nonrigid deformation (the second row).}
	\label{fig:5shape_noise_out}
\end{figure}

\subsection{Results on geometrically deformed graphs}\label{sec:shape}
In this section, we evaluated our algorithm FRGM-D for matching graphs with geometric deformations. We chose 5 templates: Olympic logo (113 nodes), whale (150 nodes), Chinese character (105 nodes), tropical fish (91 nodes) and UCF fish (98 nodes), which have been widely used by registration methods~\cite{[2011-Jian],[2010-Myronenko],[2010-Chen]}. Fig.~\ref{fig:5shape_example_all} shows some results obtained by FRGM-D, in which the graphs are disturbed by geometric deformations, noises, missing points and outliers.

\vspace{1mm}
{\bf{Robustness to deformations}}.
For each template denoted as graph $\mathcal{G}_1$, the graph $\mathcal{G}_2$ was generated by adding geometric deformations to $\mathcal{G}_1$. To evaluate the robustness to rotation and scaling, we rotated the template $\mathcal{G}_1$ by varying degrees in $[-\pi,\pi]$ and scaled $\mathcal{G}_1$ with varying scaling factors in $[0.1,1]$ and $[2,20]$. In addition, the graph $\mathcal{G}_2$ was also disturbed with noise with distribution $\mathcal{N}(0,0.02)$ and outliers $n_{out}=100$ with distribution $\mathcal{N}(0,0.25)$.
To evaluate the robustness to nonrigid deformations, we deformed $\mathcal{G}_1$ by weight matrix $\mathbf{W}$ with distribution $\mathcal{N}(0,\sigma^2)$ with varying $\sigma\in [0,0.5]$ and abandoned the extremely deformed $\mathcal{G}_2$. Then, some slight noises with distribution $\mathcal{N}(0,0.02)$ were also added. For our algorithm FRGM-D, in the alternation that estimated the correspondence $\mathbf{P}$, the unary term $\mathbf{U}$ was computed using the rotation-invariant shape context that was also used in some other works~\cite{[2016-Ma],[2006-Zheng]}. Moreover, we solved FRGM-D by the proposed AFW method.
 
For all methods, we computed the average error between each point $\tau(V^{(1)}_i)$ and the corresponding point $V^{(2)}_{\delta_i}$, {\em i.e.}, $\frac{1}{m}\sum_i||\tau(V^{(1)}_i)-V^{(2)}_{\delta_i}||$. Moreover, to evaluate the parameterization $\mathcal{T}$ ({\em i.e.}, binary correspondence $\mathbf{P}$) obtained by FRGM-D, we also reported the average error between $\mathcal{T}(V^{(1)}_i)$ and the correspondence $V^{(2)}_{\delta_i}$, {\em i.e.}, $\frac{1}{m}\sum_i||\mathcal{T}(V^{(1)}_i)-V^{(2)}_{\delta_i}||$. We denoted these two types of average errors as FRGM-D1 and FRGM-D2 for our algorithm. As shown in Fig.~\ref{fig:5shape_rotation} (a) and (b), our algorithm is more robust to rotation and scaling factor. As shown in Fig.~\ref{fig:5shape_rotation} (c), FRGM-D1 is competitive with GMM and CPD, and FRGM-D2 has less average errors due to the well-estimated $\mathbf{P}$. 

\vspace{1mm}
{\bf{Robustness to noise and outliers.}} In this experiment, each template $\mathcal{G}_1$ was first deformed with similarity and nonrigid deformations to obtain $\mathcal{G}_2$. Then, $\mathcal{G}_2$ was disturbed by noises $\mathcal{N}(0,\sigma^2)$ with $\sigma\in[0,0.05]$ and ratios of outliers varying in $[0,1]$. In addition, we also randomly neglected inliers in $\mathcal{G}_1$ with missing point ratios in $[-0.5,0]$.

As shown in Fig.~\ref{fig:5shape_noise_out} (a) and (b), under similarity deformation, both FRGM-D1 and FRGM-D2 have less average errors for graphs with missing inliers, outliers and noises. As shown in Fig.~\ref{fig:5shape_noise_out} (c) and (d), under nonrigid deformation,  FRGM-D2 has less average error. FRGM-D1 is competitive with GMM and CPD in the cases with noise and results in higher average error when there are too many missing points or outliers.

\vspace{1mm}{\bf{Running time of FW and AFW.}}
Finally, we evaluated the average execution time on the graphs with rigid or nonrigid geometric deformations when the algorithm FRGM-D was solved by FW or AFW, respectively. As shown in Tab.\ref{tab:FW_AFW}, the AFW-based implementation is nearly 10 times faster.
\begin{table}[t!]
	\centering
	\small
	\renewcommand{\arraystretch}{1.4}
		\caption{The execution times for FW and AFW implementations.}
	\vspace{3mm}
	\begin{tabular}{c|ccccc}
		\toprule[1.0pt]
		\diagbox[width=18mm,trim=l]{Method}{Template} 
		& Temp-1 & Temp-2 & Temp-3 & Temp-4 & Temp-5\\
		\hline
		{FW}
		&  35.3s   & 69.4s   & 25.8s  & 24.9s  &  30.0s  \\
		\hline
		{AFW}
		&  4.4s    & 7.8s    & 3.8s  & 3.3s  &  3.6s \\
		\bottomrule[1.0pt]
	\end{tabular}
	\label{tab:FW_AFW}
\end{table}

\section{Conclusion and Discussion}\label{conclusion}
In this paper, we introduce a functional representation for the GM problem. The main idea is to represent both the graphs and node-to-node correspondence by linear function spaces and linear functional representation map and represent the pairwise information of graphs by geometric-aware functionals defined on the function spaces. There are three main contributions resulting from the functional representation. First, the representation provides geometric insights for the general GM, by which we can construct more appropriate objective functions and algorithms. Second, the linear representation map is a new parameterization approach for the Euclidean GM and helps to handle both conventional and geometrically deformed graphs. Third, the representation of graph attributes can be used as a replacement for the costly affinity matrix and reduce the space complexity. Finally, both efficient algorithms and optimization strategy have been proposed to solve the proposed GM algorithms with better performance.

Beyond the scope explored in this paper, there are some other problems that may benefit from our work. There are three inspirations. (1) For the basis functions used to construct the function space of a graph, some flexible choice is available for different real applications, {\em e.g.}, the eigenfunctions of the Laplace-Beltrami operator for 3D surface analysis. Therefore, more proper geometric structures can be defined to fix the GM problem between surfaces. (2) For the hypergraph or multigraph matching problem, which results in much higher computational complexity, the functional representation may be used to reduce the space complexity by representing the higher-order edge attributes and the matching configuration among the product space of function spaces or background Euclidean spaces. (3) The proposed approximated Frank-Wolfe method can be used to improve the other GM algorithms or some problems that need to optimize the objective functions upon the feasible field $\hat{\mathcal{P}}$ by some modifications. 

There are also some limitations of FRGM. For example, the current version of FRGM-G can only handle undirected graphs due to the construction of the inner product or metric, which requires symmetric edge attributes. In future work, we may address this issue by extending the inner product or metric with more general functionals, such as bilinear functional on the product space of function spaces of graphs.


%



%

\bibliographystyle{IEEEtran}
\bibliography{egbib}

%
%

%

\end{document}